\documentclass[10pt,twocolumn]{article}

\usepackage{cvpr}
\usepackage{times}
\usepackage{epsfig}
\usepackage{graphicx}
\usepackage[export]{adjustbox}
\usepackage{amsmath}
\usepackage{amssymb}
\usepackage{bbm}
\usepackage{placeins}
\usepackage{mathtools}
\usepackage{booktabs}
\usepackage{xfrac}
\usepackage{float}
\usepackage{caption}
\usepackage{subcaption}
\usepackage[dvipsnames]{xcolor}

\usepackage{adjustbox}

\usepackage{amsmath}

\newcommand\INN{f_\theta}

\usepackage{multirow}
\usepackage{xcolor,colortbl}
\usepackage{rotating}

\makeatletter
\@namedef{ver@everyshi.sty}{}
\makeatother
\usepackage{tikz}
\usetikzlibrary{arrows, arrows.meta, shapes, shapes.misc, calc, positioning, decorations.pathreplacing, fit, backgrounds}
\usepackage{keyval}
\usepackage{ifthen}

\cvprfinalcopy 

\usepackage[pagebackref=true,breaklinks=true,letterpaper=true,colorlinks,bookmarks=false]{hyperref}

\DeclareMathOperator*{\logsumexp}{logsumexp}
\DeclareMathOperator*{\logsoftmax}{log\,softmax}

\begin{document}

\addtocontents{toc}{\protect\setcounter{tocdepth}{0}}

\title{Generative Classifiers as a Basis for Trustworthy Image Classification}
\author{
Radek Mackowiak* \\
\small{Visual Learning Lab, Heidelberg University}
\and
Lynton Ardizzone*\\
\small{Visual Learning Lab, Heidelberg University}
\and
Ullrich Köthe\\
\small{Visual Learning Lab, Heidelberg University}
\and
Carsten Rother\\
\small{Visual Learning Lab, Heidelberg University}
}
\maketitle
\begin{abstract}
\noindent
With the maturing of deep learning systems, 
trustworthiness is becoming increasingly important for model assessment.
We understand trustworthiness as the combination of explainability and robustness.
Generative classifiers (GCs) are a promising class of models that are said to naturally accomplish these qualities.
However, this has mostly been demonstrated on simple datasets such as MNIST and CIFAR in the past.
In this work, we firstly develop an architecture and training scheme 
that allows GCs to operate on a more relevant level of complexity for practical computer vision, namely the ImageNet challenge.
Secondly, we demonstrate the immense potential of GCs for trustworthy image classification.
Explainability and some aspects of robustness are vastly improved compared to feed-forward models,
even when the GCs are just applied naively.
While not all trustworthiness problems are solved completely,
we observe that GCs are a highly promising basis for further algorithms and modifications.
We release our trained model for download in the hope that it serves as a starting point for other generative classification tasks, 
in much the same way as pretrained ResNet architectures do for discriminative classification. \\ \noindent
Code: \href{https://www.github.com/VLL-HD/trustworthy_GCs}{\texttt{github.com/VLL-HD/trustworthy\_GCs}}
\end{abstract}

\vspace{-3mm}
\section{Introduction}

\begin{figure}[htp]
\centering
\includegraphics[width=\linewidth]{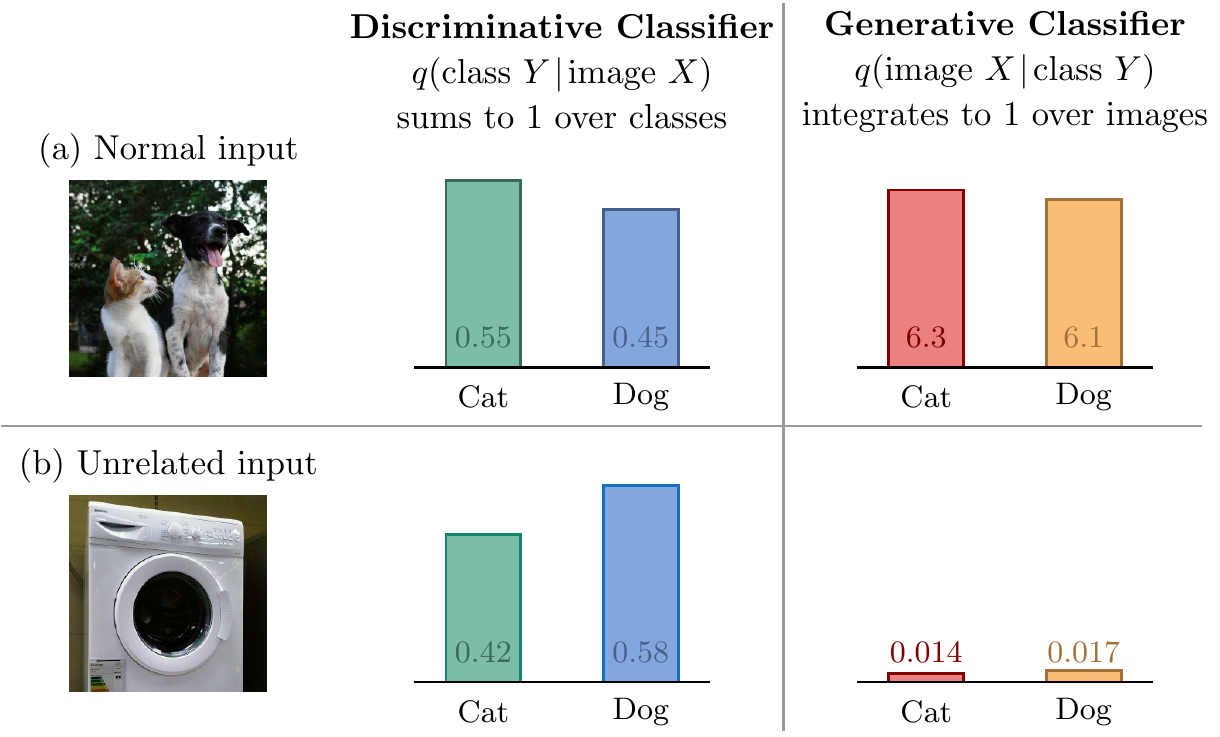}
\vspace{-6mm}
\caption{
Example of one advantage of generative classifiers:
The class posterior of a DC always sums up to 1,
while the likelihoods of the GC do not have this restriction,
constituting inherently more informative outputs.
E.g. the GC can show if
a prediction is uncertain because the input agrees with both classes, or with neither.
}
\label{fig:dc_gc_example}
\vspace{-3mm}
\end{figure}

Generative classifiers (GCs) and discriminative classifiers (DCs) represent two contrasting ways of solving classification tasks.
In short, while standard DCs model the class probability given an input directly, $p(\text{class}\mid\text{image})$ (e.g. softmax classification), generative classifiers (GCs) take the opposite approach:
They model the likelihood of the input image, conditioned on each class, $p(\text{image}\mid\text{class})$.
The actual classification is then performed by finding the class under which the image has the highest likelihood. 

The application of GCs has so far been limited to very simple datasets such as MNIST, SVHN and CIFAR-10/100.
For any practical image classification tasks, DCs are used exclusively, due to their excellent discriminative performance.
In principle, GCs are said to have various advantages over DCs, which align with the term \emph{trustworthiness}. 
In general agreement with \cite{huang2018survey}, we understand trustworthiness as the combination of explainability and robustness.

\noindent
\textbf{Explainability:}
DCs based on deep neural networks are notorious for being `black boxes', prompting many developments in the field of explainable AI.
In the taxonomy laid out in \cite{gilpin2018explaining}, most commonly used algorithms fall into categories I or II:
post-hoc methods that visualize how a network processes information (I), or that show its internal representations (II).
The explanations can vary depending on the chosen method, and there is no guarantee that the results faithfully reflect what the DC is doing internally.

In contrast, GCs bring to mind Feynman's mantra ``What I cannot create, I do not understand''.
As GCs are able to model the input data itself, not just the class posteriors, they have fundamentally more informative outputs. 
For instance, GCs allow us to tell if a decision between two classes is uncertain because the input agrees well with \emph{both} classes, or with \emph{neither} 
(see Fig.~\ref{fig:dc_gc_example}).
In addition, most GCs have interpretable latent spaces with meaningful features,
allowing for the actual decision process to be directly visualized without post-hoc techniques.
Therefore, it could be argued that GCs belong to category III of the explainability  taxonomy \cite{gilpin2018explaining}, 
i.e. methods that intrinsically work in an explainable way, without relying on additional algorithms.

\noindent
\textbf{Robustness:}
A second large concern about the practical use of deep learning based classification systems is their robustness,
which can have different meanings, depending on the context. 
In particular, GCs have been assumed to be superior to DCs in terms of 
generalization under dataset shifts \cite{ulusoy2006comparison,raymond2007generative} and accurately calibrated posteriors \cite{ardizzone2020exact}.
In addition, a big advantage of GCs is their capability to explicitly identify abnormal inputs in a natural way, thus indicating when a decision should not be trusted.
Furthermore, GCs were found to be more robust towards adversarial attacks \cite{li2018generative} and allow for their explicit detection \cite{ghosh2019resisting}.

It is still unclear if GCs can also manifest these advantages in more complex tasks while remaining competitive to DCs in task performance. For example, the authors of~\cite{fetaya2020} find while GCs can successfully detect adversarially attacked MNIST images, this already fails for the CIFAR-10 dataset. 
The authors of \cite{nalisnick2018deep,kirichenko2020normalizing} observe that detection of other forms of OoD data also fails in various ways for natural images.
In \cite{fetaya2019conditional}, the authors cast doubt on whether GCs can be used for high-dimensional input data at all.

In light of this background, our work makes the following contributions:
\emph{(i)} We design and train a GC that performs at a level relevant to practical image classification, demonstrated on the ImageNet dataset.
\emph{(ii)} We show various native explainability techniques unique to GCs.
\emph{(iii)} We examine the model in terms of robustness.

Overall, we find our GC to work better than a comparable DC in terms of trustworthiness.
However, we do observe that previous findings on superior generalization under dataset shift \cite{ulusoy2006comparison}
and immunity to adversarial attacks \cite{schott2019towards} do not hold for the ImageNet dataset.
For other aspects of robustness, our GC shows some great benefits, such as naturally detecting OoD inputs and adversarial attacks.

\vspace{-1mm}
\section{Related Work}
\vspace{-1mm}
Years before the deep learning revolution, works such as \cite{ng2002discriminative,ulusoy2006comparison,raymond2007generative} 
already compared the properties of GCs vs DCs, theoretically and experimentally, with agreement that GCs are more robust and more explainable.
Works like \cite{bouchard2004tradeoff,Lasserre2007GenerativeOD,xue2010generative} presented models that combine the aspects of GCs and DCs,
to reach a more favourable trade-off compared to each extreme.
However, all these works consider simple problems,
and with the unmatched task performance later delivered by deep-learning based DCs in the 2010s, GCs became rarely used.

As one example of more recent work, \cite{fetaya2019conditional} investigates normalizing-flow based GCs trained on natural images.
The authors find that naively trained GC models achieve very poor classification performance, 
and argue that this is due to some model properties that are not properly penalized by maximum likelihood training.
Later, \cite{ardizzone2020exact} propose that this problem can be avoided by training with the Information Bottleneck loss function instead.
The authors of \cite{lee2019robust} modify the problem, and train a GC on features previously extracted from a standard feed-forward network.
For all these works, the most complex dataset used is CIFAR-100, at a resolution of $32\times 32$ pixels.

So-called hybrid models \cite{raina2004classification} have been more successful in practice. 
Here, a likelihood estimation method is involved, commonly for the marginal $p(\mathrm{image})$, while the actual classification is still performed in a discriminative way,
using shared features between the two tasks, the main motivation being semi-supervised learning.
Notable examples are \cite{kingma2014semi, dumoulin2017learned, chongxuan2017triple, nalisnick2019hybrid, grathwohl2019your}.
They have some fundamental differences to GCs, 
e.g.~that the conditional likelihoods are not directly modeled and the latent space has no explicit class structure.

Concerning OoD detection with generative models,
the authors of \cite{nalisnick2018deep} and later \cite{kirichenko2020normalizing} observed that likelihood models trained on natural images fail to detect certain OoD inputs,
and may perform significantly worse than random.
This problem is addressed e.g.~by \cite{nalisnick2019detecting,choi2018waic,serra2019input,song2019unsupervised,zhang2020out}, where different OoD scores are introduced that correct for these shortcomings.
These works only consider unconditional likelihood models for OoD detection, while a separate classifier is still needed to perform the actual task.
GCs combine both these steps into a single model, simplifying the process and potentially improving OoD detection at the same time.

GCs have also been examined for adversarial defense recently \cite{schott2019towards, ghosh2019resisting, li2018generative}.
While these works highlight the potential of GCs, they are limited to simple datasets such as MNIST and SVHN, and do not scale to problems with more than approx.~$10$ classes,
or to natural images \cite{fetaya2020}.

\vspace{-1mm}
\section{Methods}
\vspace{-1mm}
\newcommand{\qt}{q_\theta}
\subsection{Invertible Neural Networks}
\noindent
While VAEs have been used as generative classifiers with some success~\cite{schott2019towards, ghosh2019resisting, li2018generative}, perhaps the most natural choice are normalizing flows, due to their exact likelihood estimation capabilities~\cite{dinh2016density}.
The networks used in normalizing flows are so-called invertible neural networks (INNs),
a class of neural network architectures that meet the following conditions:
(i) the network represents a diffeomorphism  by construction (essentially, a smooth and invertible function),
(ii) the inversion can be computed efficiently, and
(iii) the network has a tractable Jacobian determinant.
These conditions place some restrictions on the architecture, e.g. that the number of input and output dimensions have to be equal,
and that non-invertible operations such as max-pooling can not be used.
In recent years, various different invertible architectures have been developed that fulfill these conditions \cite{dinh2014nice, dinh2016density, behrmann2018invertible, grathwohl2018ffjord}.
In this work, we employ the affine coupling block architecture proposed in \cite{dinh2016density}, 
with additional modifications, as described in Appendix \ref{app:architecture_and_training}.

In any generative setting, there are training images $X$, that follow some unknown image distribution $p(X)$.
The goal is then to approximate $p(X)$ as closely as possible with a distribution given by the network, which we denote as $\qt(X)$.
In the case of normalizing flows, $\qt(X)$ is represented by transforming possible inputs $X$ to a latent space $Z$ using an INN $\INN$ (`flow'), 
with a prescribed standard normal latent distribution $p(Z) = \mathcal{N}(0,1)$ (`normalizing').
Then, the change-of-variables formula can be used to compute $\qt(X)$ at any point $x$ through
\vspace{-2mm}
\begin{equation}
    \qt(x) = p\Big(Z\!=\! \INN(x) \Big) \left| \det J(x) \right|
    \label{eq:change_of_vars}
\end{equation}\\[-6mm]
with $J\equiv \partial \INN/ \partial X$ being the Jacobian.
It can be shown that the network will learn the true distribution ($\qt(X) = p(X)$) by maximizing the expected log-likelihood $\log \qt(X)$, as given through Eq.~\ref{eq:change_of_vars} above \cite{tabak2013family}. 
After training is complete, the model can not only be used to estimate likelihoods $\qt(X)$, but also to generate new samples 
by inverting the network, in order to map sampled instances of $Z$ back to image space.

In our case, this approach is not sufficient, as we want to use the INN as a generative {classifier},
meaning we need to model conditional likelihoods $\qt(X \mid Y)$.
While different approaches for this exist \cite{winkler2019learning, ardizzone2019guided}, we adopt the form introduced in \cite{izmailov2019semi}.
Here, the latent distribution is a conditional density $p(Z\mid Y)$:
The standard normal distribution $p(Z)$ is replaced with a Gaussian Mixture Model (GMM) containing a unit-variance mixture component per class
\vspace{-2.5mm}
\begin{align}
    p(Z\,|\,Y) &= \mathcal{N}(Z; \mu_Y,\mathbbm{1}) \\
    p(Z) = \sum_y p(y)\, p(Z\,|\,y) &= \sum_y p(y) \mathcal{N}(Z; \mu_y,\mathbbm{1}) 
\end{align} \\[-6mm]
where $\mu_y$ is the mean of class $y$ in latent space; and the mixture weights are the class priors $p(y)$, 
i.e. the frequency of occurrence of each class in the dataset.
The conditional likelihood $\qt(X|Y)$ can be evaluated with the change-of-variables formula (Eq.~\ref{eq:change_of_vars}) as before by replacing the full distribution $p(Z)$ with the appropriate mixture component:
\vspace{-1.5mm}
\begin{equation}
    \qt(X\,|\, Y) = p\Big(Z\!=\! \INN(X)\,\Big|\, Y \Big) \left| \det J \right| .
    \label{eq:change_of_vars_cond}
\end{equation}

\subsection{Training INNs with Information Bottleneck}
\noindent
An INN naively trained with a class-conditional log-likelihood loss will perform very poorly as GC, even on mildly challenging tasks \cite{fetaya2019conditional}.
Instead, we require a loss function where the focus on the generative and class-separating capabilities can be explicitly controlled.
For this, we utilize the IB objective \cite{tishby2000information}, the ideal loss function for robust classification from an information theoretic point of view.
Given some features $Z$ of a network, inputs $X$, and ground-truth outputs $Y$, the IB loss consists of two terms using the mutual information $I$ (MI):
\vspace{-2mm}
\begin{equation}
    \mathcal{L}_\mathrm{IB} = I(X, Z) - \hat \beta I(Y, Z) .
\end{equation}\\[-6mm]
The MI quantifies the degree of shared information between variables and can be written as 
$ I(V, W) = D_{\mathrm{KL}}( p(V,W) \| p(V) p(W) )$.
Minimizing the IB loss means maximizing the information about the desired output $Y$ contained in the features, $I(Y,Z)$.
Simultaneously, it minimizes the information about the original image contained in the features, $I(X,Z)$, resulting in robust and efficient representations $Z$.
The trade-off between these two aspects is explicitly adjusted by choosing $\hat \beta$.

How to apply this objective to INNs is not immediately obvious, as INNs preserve information, and the loss becomes ill-defined. 
The authors of \cite{ardizzone2020exact} show that this can be avoided by adding very low noise to the inputs.
This is already an established practice in the context of normalizing flows for the purpose of dequantization.
From this, the authors go on to
derive two loss terms representing the IB objective, $\mathcal{L}_\mathrm{IB} = \mathcal{L}_X  + \beta \mathcal{L}_Y$. 
In practice, the two terms amount to the following:
\vspace{-2.5mm}
\begin{align}
\mathcal{L}_X(x) = - \! \log|\det J_x|  \! + \!\frac{1}{2} \logsumexp_{y'} \Big( v_{y'}^2 \!-\! 2w_{y'}\Big) \\
\mathcal{L}_Y(x,y) = \mathrm{onehot}(y) \cdot \logsoftmax_{y'} \left( \frac{v_{y'}^2}{2} - w_{y'}\right)
\end{align} \\[-5mm]
Hereby, we use $v_y \coloneqq f(x) \!-\! \mu_y$, and $w_y \coloneqq \log p(y)$ ($\log(1/ \text{(\# classes)})$ for uniform class priors in our case).
$J_x$ is the Jacobian $\partial f(x) / \partial x$.
$y'$ denotes the summation over all classes in the $\logsumexp$ and $\logsoftmax$ operations.
The difference between $\hat \beta$ in the original IB and $\beta$ in the loss is a constant weighting factor for convenience \cite{ardizzone2020exact}, 
producing a sensible objective for manageable values of $\beta$ in the rough range $(1,100)$.

Intuitively, we find the following: 
The $\mathcal{L}_X$-loss forces the data to follow the GMM in latent space, making the INN a generative model.
However, it has no effect on the class-conditional aspect, as the class $y$ is summed out. 
This loss can be rearranged to look similar to the maximum-likelihood-loss used for normalizing flows, but with a GMM as a latent distribution.
On the other hand, the $\mathcal{L}_Y$-loss bears resemblance to the categorical cross entropy loss, except that the usual logits are replaced by $\log p(z|y)p(y) = \log p(z,y)$.
Therefore, $\mathcal{L}_Y$ is responsible for making the likelihood model conditional on the class, but otherwise ignores the generative performance.

\subsection{Detecting OoD Inputs}
\label{subsec:ood_test_method}
\newcommand{\logq}{\log \qt(x)}

For likelihood-based generative models, detecting OoD inputs is straight forward, by directly utilizing the estimated probability density $\qt$:
in principle, if an input is outside the support of the training data, and the model has learned the true distribution,
the OoD sample should be assigned $\logq = - \infty$.
In practice, it is only required that OoD samples have lower likelihood scores than the training data.
From here, any input with an inferred likelihood below a threshold can be treated as OoD.
However, in \cite{nalisnick2018deep}, the authors identified various special cases where OoD inputs have an unnaturally high log-likelihood score.
This prompted the development of a \emph{typicality-test} in \cite{nalisnick2019detecting}, that uses both an upper and a lower threshold.
Even better performing extensions to this exist \cite{choi2018waic,serra2019input,song2019unsupervised,zhang2020out}, but we choose the typicality-test as the simplest option, 
to examine the natural capabilities of the model.
We slightly modify the typicality-test to make it a traditional hypothesis test, with the null hypothesis being that the input is in-distribution,
more details in Appendix \ref{app:ood_detectionn_method}.
The $p$-value for the hypothesis test is the fraction of training samples with scores in the OoD-zone, which also equals the false positive rate.
To evaluate the OoD detection capabilities independent of the threshold,
we vary the $p$-value of the test and produce a receiver operating characteristic (ROC) curve.
The area under this curve (ROC-AUC), in percent, serves as a scalar measurement of the OoD detection capabilities,
with ROC-AUC of 100\% meaning that the OoD samples and in-distribution samples are perfectly separated,
and a value at 50\% or below indicates a random performance or worse.

\vspace{-1mm}
\section{Experiments}
\vspace{-1mm}

A detailed description of the network architecture is found in Appendix \ref{app:architecture_and_training},
we summarize the main points in the following.
We construct the invertible network (INN) from affine coupling blocks, as introduced in \cite{dinh2016density},
with various modifications from other recent works 
\cite{ardizzone2018analyzing,ardizzone2019guided,jacobsen2018excessive,kingma2018glow}.
As invertible alternatives to $2\times 2$ max-pooling and global mean-pooling, we use
a Haar wavelet transform \cite{ardizzone2019guided}
and a DCT transform \cite{jacobsen2018excessive} respectively.

Because of the similarities between affine coupling blocks and residual blocks as used in a ResNet,
we match the design of the INN to that of a standard ResNet-50 wherever possible.
The overall layout is summarized in Table \ref{tab:dims_and_features}, c.f.~\cite[Table 1]{he2016deep}.
Some differences arise due to the constraint of invertibility:
the number of feature channels and the available receptive field vary between the two networks.
Regarding the effective rather than maximum receptive field, see Appendix \ref{app:receptive_field}.
The invertibility is also associated with an extra cost of parameters and computation,
summarized in Appendix Table \ref{app:tab:computational_cost}:
Both in terms of network parameters, as well as FLOPs for one forward pass,
the cost of the INN is about twice as high as a standard ResNet-50.
We are optimistic that this overhead can be reduced in the future with more efficient INN architectures.

\begin{table}
    \centering
    \resizebox{\linewidth}{!}{

\begin{tabular}{l|r|r|r|r|r|r}
\textbf{Layer} & \textbf{Blocks} & \textbf{Im. size} & \multicolumn{2}{c|}{\textbf{Channels}} & \multicolumn{2}{c}{\textbf{R.F.}} \\
      &        &          & INN  & ResNet                & INN & ResNet   \\
    \hline

Input & & 224 & 3 & 3 & & \\
Entry flow & 1 & 112 & 12 & 64 & 8 & 6\\
Pool (Haar/max) & & 56 & 48 & 64 & 10 & 10 \\
    \hline 
Conv\_2\_x & 3 & 56 & 48 & 256 &  34 &  34 \\
Conv\_3\_x & 4 & 28 & 192 & 512 & 106 &  90 \\
Conv\_4\_x & 6 & 14 & 768 & 1024 & 314 &  266 \\
Conv\_5\_x & 3 & 7 & 3072 & 2048 & 538 &  426 \\
    \hline 
Pool (DCT/avg.) & & 1 & 150\,528 & 2048 & $\infty$  & $\infty$ \\
\end{tabular}}
    \vspace{-2mm}
    \caption{For each of the resolution levels in the INN and ResNet-50,
    the number of coupling/residual blocks and spatial size is given,
    along with the number of feature channels and the maximum possible receptive field (R.F.).}
    \label{tab:dims_and_features}
    \vspace{-4mm}
\end{table}

\label{sec:experiments}
\subsection{General Performance}

\begin{figure}
    \centering
    \includegraphics[width=0.49\linewidth]{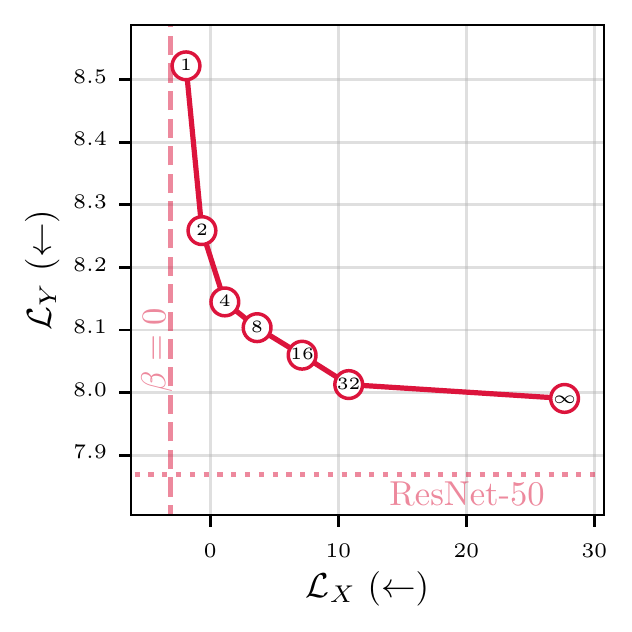} \hfill %
    \includegraphics[width=0.49\linewidth]{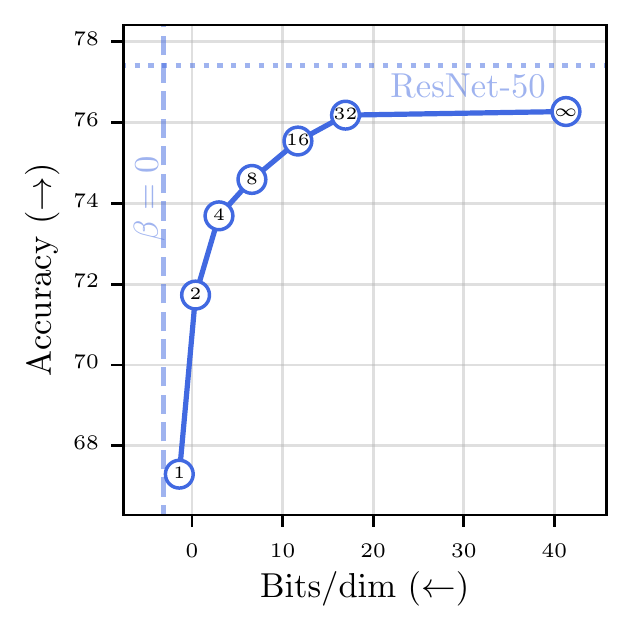}
    \vspace{-4mm}
    \caption{Trade-off between the two losses $\mathcal{L}_X$ and $\mathcal{L}_Y$ (left), 
             and between generative modeling accuracy in bits/dim, and top-1 accuracy (right).
             Each point represents one model, trained with a different beta.
             A standard ResNet has no $\mathcal{L}_X$ loss and is shown as a horizontal line.
             The model with $\beta=0$ (standard normalizing flow) is missing the $\mathcal{L}_Y$ loss 
             and is shown as a vertical line.
             The small numbers inside the markers give the value of $\beta$ of that particular model.
             }
    \label{fig:performance_trade_off_curves}
    \vspace{-4mm}
\end{figure}

We train several generative classifiers, with the following values for the hyperparameter $\beta \in \{ 1, 2, 4, 8, 16, 32, \infty \}$.
Again, $\beta$ controls how much the model focuses on the generative likelihood estimation aspect (low $\beta$), 
vs.~prioritizing good classification performance (high $\beta$).
In addition, we include a model trained with $\beta=0$, i.e.~no classification at all, analogous to a standard normalizing flow,
as well as a standard feed-forward ResNet-50 \cite{he2016deep}, i.e.~a pure DC.

The primary performance metrics used in Table \ref{tab:performance_table} and Fig.~\ref{fig:performance_trade_off_curves} are
firstly, the top-1 accuracy on the test set (in our case, the ILSVCR 2012 validation set \cite{russakovsky15imagenet}). 
We use 10-crop testing, which is most commonly used for performance evaluation in this setting.
Secondly, for the generative likelihood estimation performance, we use the bits per dimension (`\emph{bits/dim}') metric, as this is the prevalent evaluation metric for 
likelihood-based generative models such as normalizing flows.
It quantitatively measures the accuracy of the density estimation (i.e.~generative performance), explained e.g.~in \cite{theis2015note}, where a lower bits/dim corresponds to a more accurate generative model.

In Table \ref{tab:performance_table}, we report the test losses and the two discussed performance metrics for the different models.
Further shown in Fig.~\ref{fig:performance_trade_off_curves}, changing $\beta$ moves smoothly between the limit cases of a feed-forward network,
and a pure density estimation model:
the classification accuracy increases continuously with $\beta$, but a minor gap remains to the feed-forward ResNet-50, in line with works such as \cite{jacobsen2018irevnet}.
Simultaneously and as expected, the bits/dim get worse as we move away from a purely generative model ($\beta=0$).

Lastly, we examine the uncertainty calibration, a quantitative measure of the quality of the predictive posteriors.
The full analysis is provided in the Appendix Table \ref{app:tab:calibration_error}.
Here we only report the overconfidence error `OCE',
which measures the normalized classification error of predictions with a high confidence $C \geq C_\mathrm{crit}=99.7\%$.
For instance, if the error rate in these cases is $1.1\%$, although it should only be $0.3\%$ according to the confidence,
this gives an OCE of $1.1/0.3 \approx 3.7$.
Our findings are in line with previous works, in that the uncertainty calibration improves with lower $\beta$ and better generative capabilities \cite{ardizzone2020exact}.

\begin{table}
\begin{center}
\resizebox{\linewidth}{!}{\begin{tabular}{r|r|r|r|r|r}
$\beta$ & $\mathcal{L}_X^{\text{(test)}}$ ($\downarrow$) & $\mathcal{L}_Y^{\text{(test)}}$ ($\downarrow$) 
    & Bits/dim ($\downarrow$) &  Acc. (\%) ($\uparrow$) & OCE ($\downarrow$) \\ \hline
    
       $1$ &              $-1.90$ &               $8.52$ &               $4.34$ &              $67.30$  & $3.87$\\
       $2$ &              $-0.65$ &               $8.26$ &               $6.14$ &              $71.73$  & $4.13$\\
       $4$ &               $1.14$ &               $8.14$ &               $8.72$ &              $73.69$  & $4.31$\\
       $8$ &               $3.66$ &               $8.10$ &              $12.35$ &              $74.59$  & $4.73$\\
      $16$ &               $7.17$ &               $8.06$ &              $17.43$ &              $75.54$  & $4.15$\\
      $32$ &              $10.81$ &               $8.01$ &              $22.68$ &              $76.18$  & $4.94$\\
  $\infty$ &              $27.68$ &               $7.99$ &              $47.01$ &              $76.27$  & $5.12$\\ \hline
       $0$ &              $-3.11$ &                   -- &               $2.59$ &                   --  &    -- \\
    ResNet &                   -- &               $7.87$ &                   -- &              $77.40$  & $6.75$\\
\end{tabular}}
\end{center}
\vspace{-4mm}
\caption{
Test losses and metrics for models trained with different $\beta$.
Bits/dimension quantifies the performance of density estimation models
(see text, smaller is better, i.e.~more accurate generative model).
As with the original ResNet, the classification accuracy uses 10-crop testing.
OCE is the overconfidence error, i.e.~how often confident predictions are wrong
(see text, smaller is better).
}
\vspace{-4mm}
\label{tab:performance_table}
\end{table}

\begin{figure}
    \centering
    \includegraphics[width=0.49\linewidth]{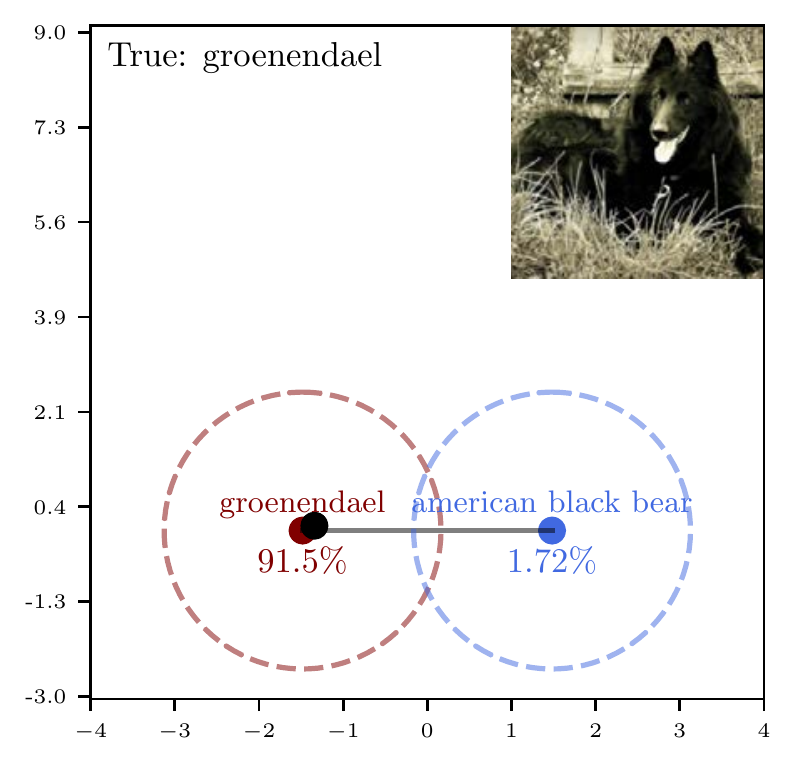} %
    \includegraphics[width=0.49\linewidth]{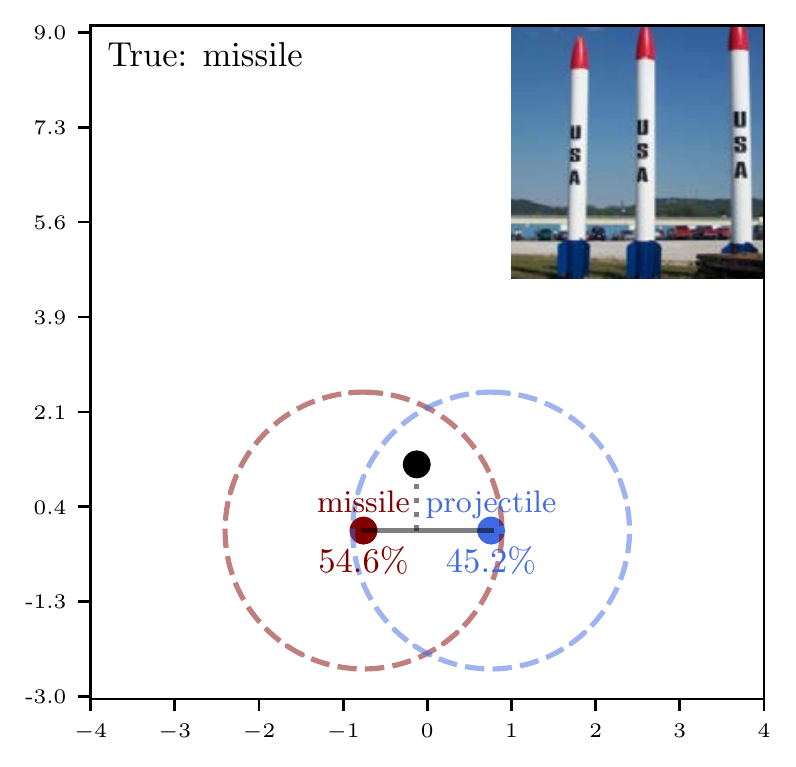} 
    \includegraphics[width=0.49\linewidth]{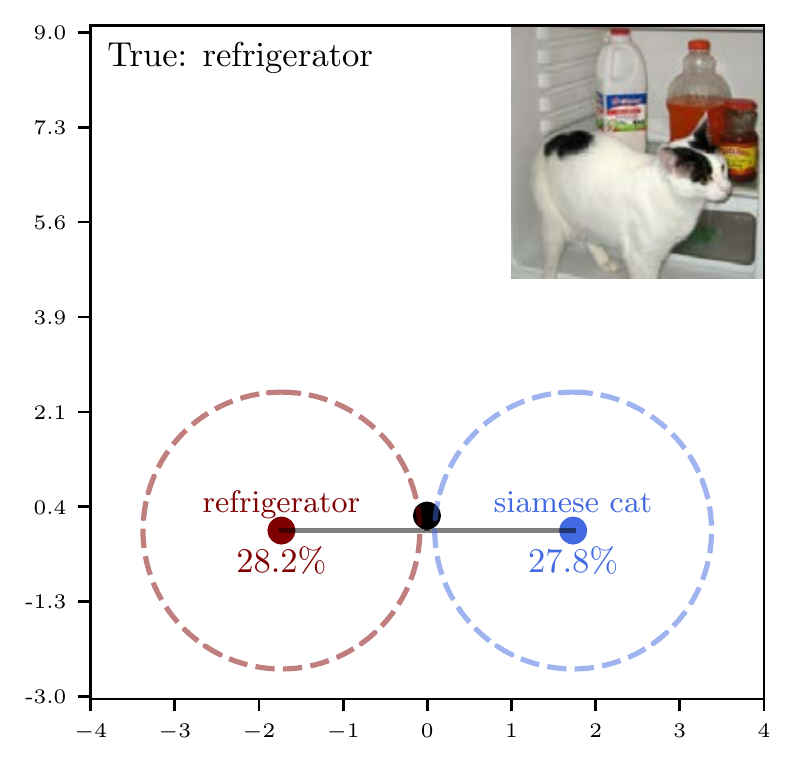} %
    \includegraphics[width=0.49\linewidth]{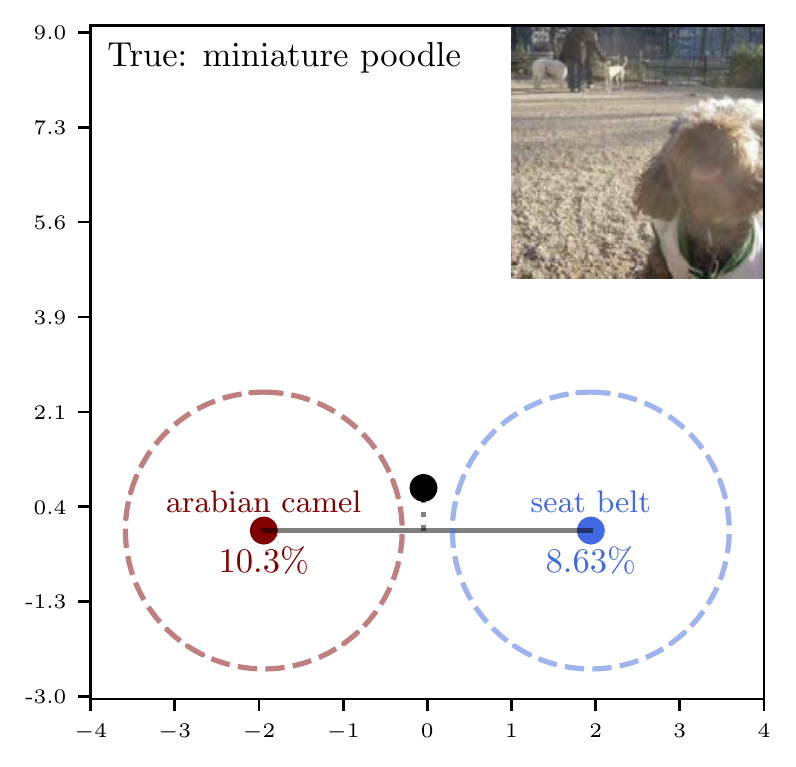}
    \vspace{-3mm}
    \caption{
    Latent space location of input images (\emph{black point}) in the decision space spanned by the $\mu_y$ of the top 5 predicted classes.
    The horizontal axis of the plot is the axis connecting the top 2 predicted classes (\emph{red and blue points}).
    The vertical axis of the plot shows the radial distance from the horizontal axis in the 5D space.
    The illustrative circles are chosen such that in both the vertical and horizontal directions, 
    90\% of the mass of the Gaussian mixture component lies inside.
    Note that the axes in the plot are scaled differently to make it appear as a circle.
    Test examples from left to right: 
    a confident in-distribution decision, 
    an uncertain in-distribution decision due to ambiguous classes, an uncertain decision due to multiple plausible image interpretations, 
    an uncertain out-of-distribution decision.
    }
    \vspace{-4mm}
    \label{fig:decision_process_latent_space}
\end{figure}

\subsection{Explainability}
In the following, we demonstrate several examples on how GCs can be used for native and intuitive explanations of the data and the prediction outputs.
Certainly, algorithms and approaches exist that can generate similar results for DCs.
The point of the following examples is to show that in GCs a range of explanations is available using only the structure of latent space and the learned likelihoods,
without requiring additional modifications or algorithms applied in a post-hoc manner.
\\[1.5mm]
\noindent \textbf{Visualizing decision-space:}
The properties of a classification decision are fully determined by the latent code of an input image in relation to the surrounding classes.
The only difficulty consists in reducing the high-dimensional latent space to a 2D plot.
Fig.~\ref{fig:decision_process_latent_space} shows one possibility:
latent codes are visualized in a plane through the centers of the two most probable classes, such that relative distances to the centers and to their connecting axis are preserved.
A second approach is given in Appendix \ref{app:explainable_2d_space}, where the classification among a subset of classes can be fully visualized.
\begin{figure*}
    \vspace{-6mm}
    \resizebox{\textwidth}{!}{
        \parbox{6.2cm}{
        \begin{flushright}
        \includegraphics[]{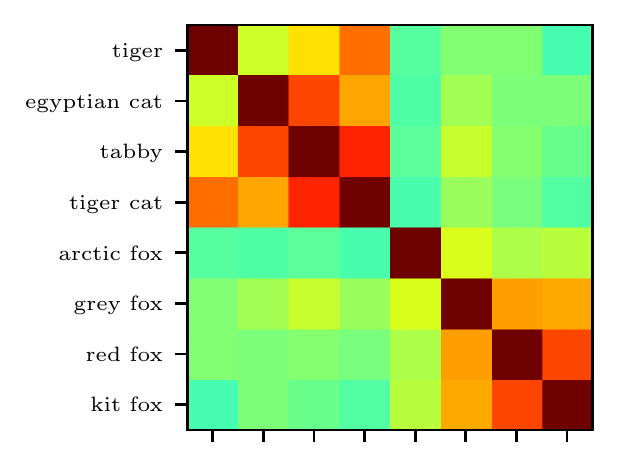}
        \end{flushright}
        }
        \parbox{6.2cm}{
        \begin{flushright}
        \includegraphics[]{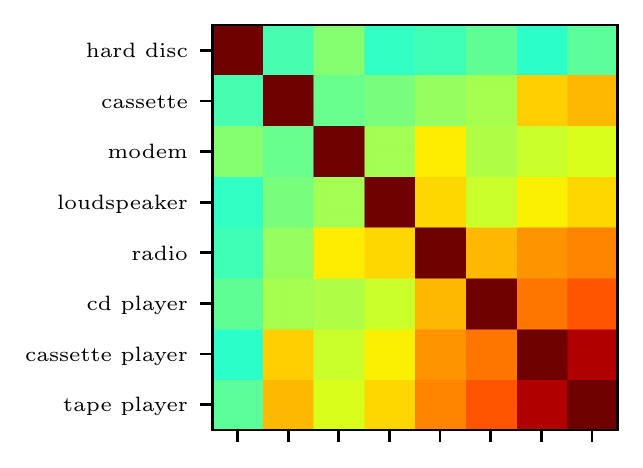}
        \end{flushright}
        }
        \parbox{6.2cm}{
        \begin{flushright}
        \includegraphics[]{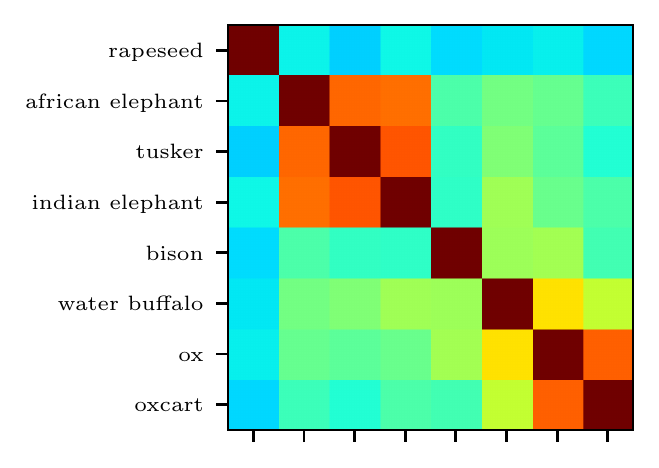}
        \end{flushright}
        } \hspace{8mm}
        \parbox{2.0cm}{
        \vspace{-1mm}
        \includegraphics[height=4.25cm]{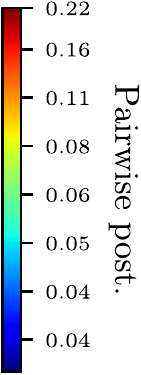}
        }
    }%
    \vspace{-4mm}
    \caption{
    Latent similarity between different classes.
    The colormap indicates the pairwise distance of the $\mu_y$ as well as the expected pairwise posterior,
    meaning e.g. the binary decision between ``tabby cat" and ``tiger cat" is associated with $20\%$ expected uncertainty, by construction (see text).
    The distance on the diagonal is 0 (outside colormap range).
    \vspace{-4mm}
    }
    \label{fig:similarity_matrix_regions}
\end{figure*}
\begin{figure}
    \centering
    \includegraphics[width=\linewidth]{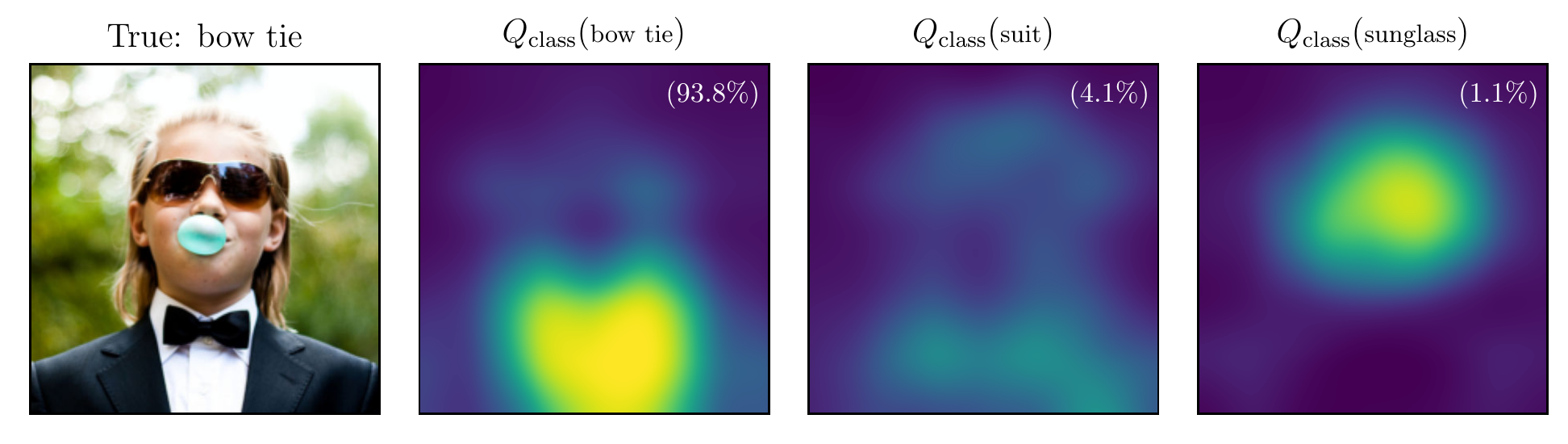}
    \includegraphics[width=\linewidth]{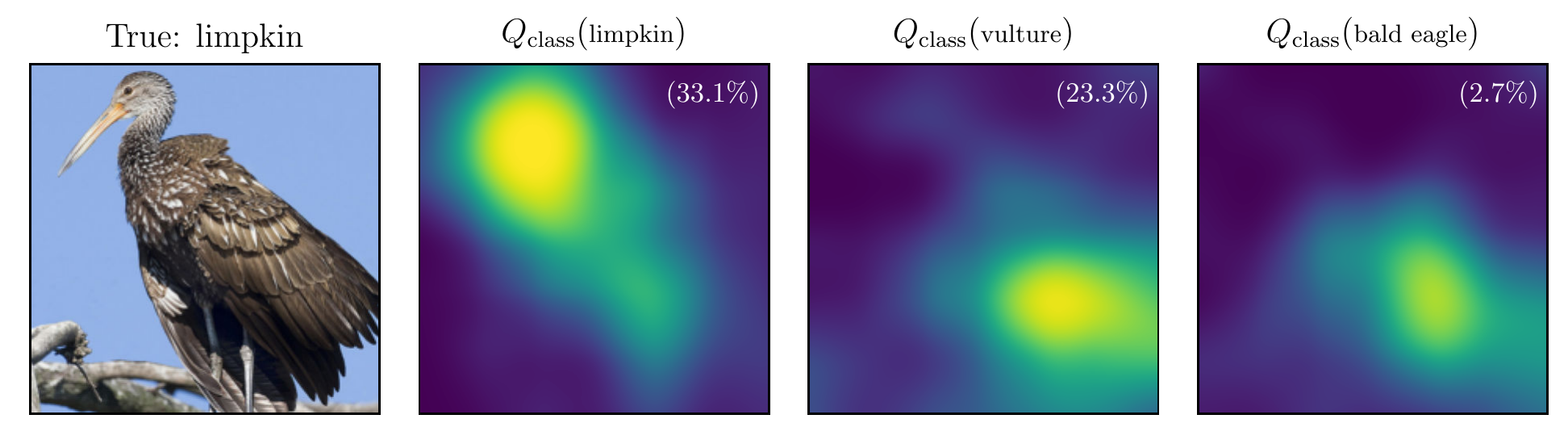}
    \includegraphics[width=\linewidth]{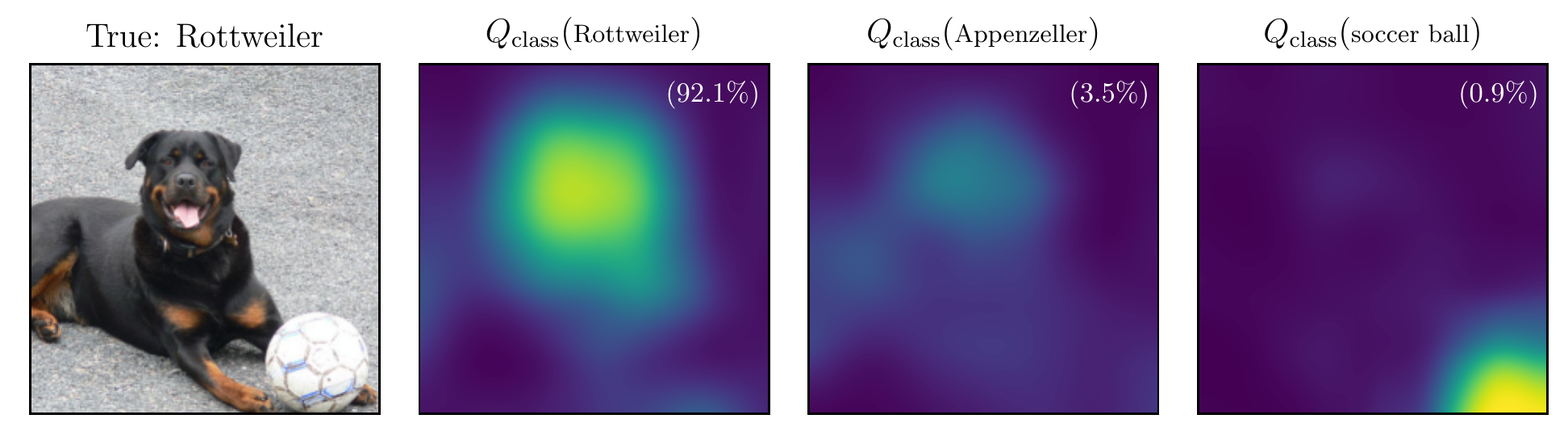}
    \caption{
    Examples of the prediction heatmaps. Summing the bright areas directly gives the final class prediction.
    Top: bowtie and sunglasses are located, suit is distributed over a large area.
    Middle: The head of the bird causes it to be classified as a limpkin, whereas the feathers are more indicative of an eagle or vulture.
    Bottom: The heatmaps of both Rottweiler and Appenzeller classes are located in the same area (ambiguous classes), while the soccer ball is separate.
    }
    \vspace{-4mm}
    \label{fig:class_heat_maps}
\end{figure}
\\[1.5mm]
\noindent \textbf{Class similarities:}
\label{subsec:class_similarities}
Building on Fig.~\ref{fig:decision_process_latent_space}, we see that different classes have various amounts of overlap,
which represents the relationship between them.
This is not possible for a feed-forward model, as there is no latent space where the input data is embedded in such a way.
We observe that the locations $\mu_y$ of the Gaussian mixture components are close together for classes that are semantically similar,
and far apart for classes that are dissimilar.

Importantly, this also has implications for predictions the model makes.
For instance, in Fig.~\ref{fig:decision_process_latent_space}, top right, the classes overlap a lot.
This means more points will lie in the overlap zone, and consequently more of these decisions will be uncertain, 
compared to e.g. bottom left, where most inputs will be in only one of two classes.
More precisely, the closer two class centers are, the larger is the overlap, 
and the larger the proportion of split decisions between these classes.
In fact, if a class A is the top prediction, 
the expected confidence for any other class B can be worked out explicitly from the distance between $\mu_A$ and $\mu_B$ in latent space, see Appendix \ref{app:explainable_similarity}.
Some examples are shown in Fig.~\ref{fig:similarity_matrix_regions}, with the full similarity matrix in Appendix Fig. \ref{app:fig:similarity_matrix}.

These considerations highlight an important fact:
the latent mixture model contains a built-in uncertainty between classes.
A decision between similar classes will always be uncertain, by the structure of the latent space alone.
This may be one of the reasons explaining why the predictive uncertainties are better calibrated in such GCs.
\\[1.5mm]
\noindent \textbf{Posterior Heatmaps:}
To increase the trust in a decision, it is often helpful to show which regions of the image were relevant.
Examples are widespread where models e.g.~base the decision on the background of the image, not the object in question, 
or focus only on a specific detail that identifies an object.
Approaches such as CAM or GradCAM \cite{zhou2016learning, selvaraju2017grad} are used to generate coarse heatmaps showing regions that are influential for a particular decision.
With the IB-INN, we can provide such heatmaps as a direct decomposition of the prediction output,
meaning they can be understood simply as a different way of representing the model output,
rather than a post-hoc explanation technique.

To produce a spatially structured output, we consider the following:
Due to the invertibility of every part of the model,
we can start from the output $z$, and transform it back through the DCT operation.
Unlike standard mean-pooling, the DCT pooling does not lose any information in either direction.
We define the following for short:
\vspace{-1mm}
\begin{align}
 w^{(y)} 
         & = \mathrm{DCT}^{-1}\big(z - \mu_y\big).
\end{align}\\[-6mm]
Importantly, $w^{(y)}$ has the spatial structure of the final convolutional outputs, $w^{(y)}_{kl}$, with height- and width indexes $k$ and $l$.
Because the DCT is linear and orthogonal, it conserves distances, i.e. $\| z - \mu_y \| = \| w^{(y)} \|$,
which allows us to write
\vspace{-3mm}
\begin{equation}
 q(z|y) \!\propto\! \exp\left( - \frac{\| w^{(y)} \| ^2}{2}  \right) \!=\! \exp\left( - \sum_{kl}  \frac{\big(w_{kl}^{(y)}\big)^2}{2} \right)
 \label{eq:sum_decomposition}
\end{equation} \\[-4mm]
This means the latent density is can be written as a sum over spatial coordinates inside the exponential.
We can do the same kind of decomposition to the posterior with a few extra steps, noting $q(y|x) = q(z|y)p(y) / q(z)$.
This leads to our heatmap $Q_\mathrm{Class}(k,l,y)$, that sums to the class posterior over space in the same way as in Eq.~\ref{eq:sum_decomposition}:
\vspace{-2mm}
\begin{equation}
\qt(y|x) = \exp\left( \sum_{kl} Q_\mathrm{Class}(k,l,y) \right).
\end{equation}\\[-4mm]
$Q_\mathrm{Class}$ has a single hyperparameter that adjusts the contrast of the heatmaps. The derivation is given in Appendix \ref{app:explainable_class_heatmaps}.
Examples are shown in Fig.~\ref{fig:class_heat_maps}.
Similarly, we can compute a salience map $Q_\mathrm{Salience}(k,l,y)$,
that decomposes $\qt(x)$ spatially, showing which parts of the image contain the most information according to the model,
explained and shown in Appendix \ref{app:explainable_salience}.

\newcommand{\xatt}{x_\mathrm{adv}}

\begin{table*}
\resizebox{\textwidth}{!}{
\newcommand{\fmA}[1]{\textcolor{Green!60!black}{\textbf{#1}}}
\newcommand{\fmB}[1]{\textcolor{YellowOrange!90!black}{#1}}
\newcommand{\fmC}[1]{\textcolor{Red!70!black}{\textit{#1}}}

\begin{tabular}{| @{}r |r | r | r | r | r | r r r | r r r r | r r r r | r r r r@{}|}
\multicolumn{6}{c}{} & \multicolumn{3}{c}{Noise} & \multicolumn{4}{c}{Blur} &
\multicolumn{4}{c}{Weather} & \multicolumn{4}{c}{Digital} \\
\hline
$\beta$ & Clean Error & mCE & \multicolumn{1}{c|}{\,rel. mCE\,} & $\Delta$ entrop. & OoD &
\scriptsize{Gauss.} & \scriptsize{Shot} & \scriptsize{Impulse} &
\scriptsize{Defocus} & \scriptsize{Glass} & \scriptsize{Motion} & \scriptsize{Zoom} &
\scriptsize{Snow} & \scriptsize{Frost} & \scriptsize{Fog} & \scriptsize{Bright} &
\scriptsize{Contrast} & \scriptsize{Elastic} & \scriptsize{Pixel} & \scriptsize{JPEG}\\
\hline 
\hline
0               & --              & --              & --              & --              & 77.51 &
\fmA{94.9}      & \fmA{94.3}      & \fmA{98.0}      & 
\fmA{95.7}      & \fmA{89.8}      & \fmA{88.3}      & \fmA{89.5}      & 
\fmC{38.1}      & \fmC{43.1}      & \fmA{94.8}      & \fmC{44.7}      & 
\fmA{96.7}      & \fmB{65.5}      & \fmB{63.0}      & \fmB{66.2}       \\
1               & 32.7            & 98.5            & 116             & 1.62             & 67.9  &
\fmA{95.3}      & \fmA{95.2}      & \fmA{98.6}      & 
\fmA{92.9}      & \fmA{87.1}      & \fmB{84.9}      & \fmA{87.4}      & 
\fmC{33.0}      & \fmC{45.4}      & \fmA{96.5}      & \fmC{43.5}      & 
\fmA{97.0}      & \fmB{60.4}      & \fmB{61.9}      & \fmB{55.6}       \\
2               & 28.27           & 92.5            & 119             & 1.75           & 73.6  &
\fmA{94.8}      & \fmA{95.2}      & \fmA{98.5}      & 
\fmA{87.8}      & \fmB{82.6}      & \fmB{81.3}      & \fmB{84.9}      & 
\fmC{30.9}      & \fmC{43.2}      & \fmA{96.5}      & \fmC{44.1}      & 
\fmA{95.2}      & \fmB{56.6}      & \fmB{61.0}      & \fmC{51.2}       \\
4               & 26.31           & 88.2            & 117             & 1.72          & 70.84 &
\fmA{92.7}      & \fmA{93.8}      & \fmA{97.4}      & 
\fmB{77.6}      & \fmB{76.7}      & \fmB{75.6}      & \fmB{81.7}      & 
\fmC{31.0}      & \fmC{43.2}      & \fmA{95.5}      & \fmC{44.5}      & 
\fmA{89.2}      & \fmC{54.1}      & \fmB{61.7}      & \fmC{48.0}       \\
8               & 25.41           & 86.8            & 117             & 1.81           & 65.85 &
\fmA{89.3}      & \fmA{91.2}      & \fmA{94.6}      & 
\fmB{56.9}      & \fmB{63.5}      & \fmB{63.1}      & \fmB{73.7}      & 
\fmC{37.6}      & \fmC{46.6}      & \fmA{87.8}      & \fmC{45.1}      & 
\fmB{71.2}      & \fmC{53.1}      & \fmB{65.1}      & \fmC{49.1}       \\
16              & 24.46           & 84.9            & 115             & 1.79           & 62.43 &
\fmB{83.7}      & \fmB{84.6}      & \fmA{88.0}      & 
\fmC{46.7}      & \fmB{56.7}      & \fmB{63.5}      & \fmB{67.9}      & 
\fmC{43.2}      & \fmC{52.0}      & \fmB{80.2}      & \fmC{45.6}      & 
\fmB{66.3}      & \fmC{53.3}      & \fmB{62.0}      & \fmC{42.7}       \\
32              & 23.82           & 83.1            & 113             & 1.71            & 55.83 &
\fmB{81.6}      & \fmB{81.5}      & \fmB{84.0}      & 
\fmC{39.8}      & \fmC{51.6}      & \fmC{50.1}      & \fmC{54.8}      & 
\fmC{43.9}      & \fmC{44.3}      & \fmB{61.6}      & \fmC{44.6}      & 
\fmC{53.9}      & \fmC{52.4}      & \fmC{52.5}      & \fmC{41.1}       \\
$\infty$        & 23.73           & 83.4            & 114             & 1.58          & 44.24 &
\fmC{39.5}      & \fmC{44.5}      & \fmC{40.6}      & 
\fmC{42.8}      & \fmC{48.1}      & \fmC{46.3}      & \fmC{46.0}      & 
\fmC{40.9}      & \fmC{38.9}      & \fmC{36.1}      & \fmC{44.3}      & 
\fmC{48.5}      & \fmC{52.2}      & \fmC{47.9}      & \fmC{47.0}       \\
\hline
ResNet          & 22.6            & 78.2            & 109             & 1.51            & --    &
\multicolumn{15}{c|}{--} \\ \hline \end{tabular}
}
\caption{We report the error on the unperturbed images (\emph{clean error}), 
the mean corruption error (\emph{mCE}) and the relative mCE, describing the relative performance degradation caused the corruptions (\emph{$\Delta$ entrop.}).
Furthermore, we report the OoD ROC-AUC detection score (\emph{OoD}) averaged over all corruptions as well as for the individual corruptions.
Meaning of colors:
\textcolor{Green!60!black}{\textbf{good detection $\geq 85\%$}};
\textcolor{YellowOrange!90!black}{some detection $>55\%$};
\textcolor{Red!70!black}{\textit{random or worse detection $\leq 55\%$}}.
}
\vspace{-4mm}
\label{tbl:robustness:corruptions}
\end{table*}

\subsection{Robustness}
\label{subsec:measuring_robustness}
\noindent
\textbf{Different Measures of Robustness:}
In current literature, there is no agreement upon a single measurement that clearly defines robustness in deep learning.
In general, the question is how a model reacts to out-of-distribution (OoD) inputs, 
meaning inputs that do not come from the same distribution as the training data.
We identify four different concepts of robustness, which are commonly used:

\noindent
(1) Especially for dataset shifts that preserve the semantic information,
a robust model is one that \textbf{retains good performance} for the OoD inputs.

\noindent
(2)
There are other cases where definition (1) is not applicable:
There is no `correct' prediction if the OoD input does not contain any of the classes which were trained for. 
The second idea of robustness is therefore
that the model should at least make \textbf{uncertain predictions} for OoD inputs, measured by discrete entropy of the predictive outputs \cite{snoek2019can}.
In reality, standard (non-robust) models make highly confident predictions on OoD data \cite{snoek2019can}.

\noindent
(3)
A robust model can be one that is able to \textbf{explicitly detect} OoD inputs.
In this case, along with the usual task output, the model has some auxiliary output that indicates whether an input is OoD. 
The model is robust by explicitly indicating that it's prediction may not be trusted in these cases.
GCs are uniquely suited for this, as the estimated likelihood of the inputs can serve as a built-in OoD detection mechanism,
but other approaches also exist~\cite{lee2018ooddetection,hsu2020ooddetection,chen2020ooddetection}.
To measure this, metrics such as the area under the receiver-operator curve can be used (AUC-ROC).

\noindent
(4)
In the context of adversarial attacks, robustness is commonly understood to be the \textbf{amplitude of adversarial perturbation}
necessary to trick the model \cite{xu2019adversarial}.
\\[2.0mm]
\textbf{Handling Corrupted Images:}
We first consider the robustness test in the sense of (1) established by \cite{hendrycks2019benchmarking}.
Here, the existing ImageNet validation images are corrupted with $5$ severity levels in $15$ different ways, examples are shown in Appendix \ref{app:robustness_corrupted_images}.
The authors propose the mean corruption error (mCE) and the relative mean corruption error (rel.~mCE) score to measure the robustness of a classifier.
We also measure the increase in predictive entropy as in \cite{snoek2019can} for robustness in the sense of (2), and perform OoD detection (3).

As can be seen in Table \ref{tbl:robustness:corruptions} the GC does not show an improvement compared to the ResNet in terms of (rel.) mCE, regardless of $\beta$.
However, it infers more uncertain predictions on corrupted data.
For OoD detection, we observe overall better scores for smaller values for $\beta$. 
We find the GC trained with $\beta=2$ to be the most robust classification model: 
It is able to detect a wide range of corruption types while being a reasonably good classifier 
(4.54 percentage point classification accuracy gap compared to the $\beta=\infty$ model and 5.67 gap compared to the ResNet). 
\begin{figure}
    \centering
    \includegraphics[width=\linewidth]{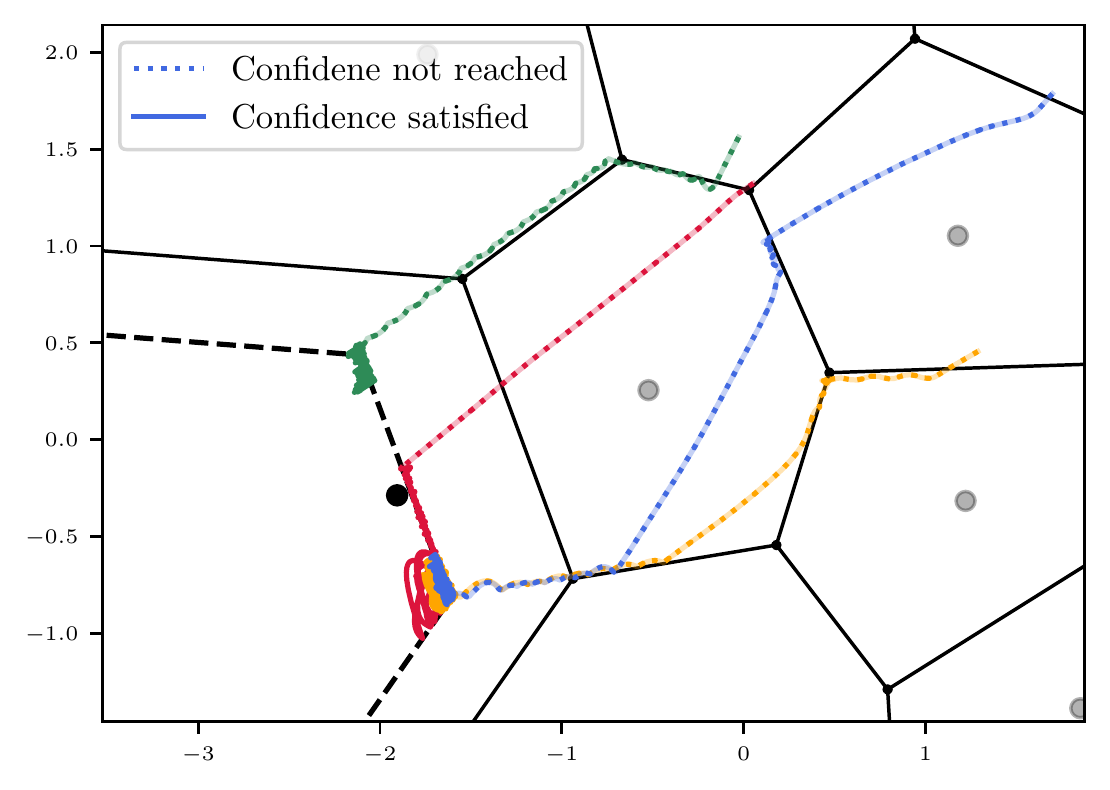}
    \includegraphics[width=\linewidth, clip, trim=0 3.5mm 0 1.5mm]{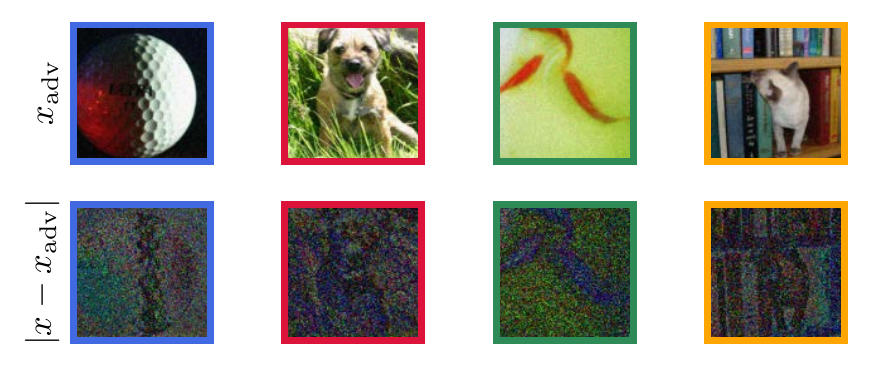}
    \vspace{-5mm}
    \captionof{figure}{
    Trajectory of four adversarial attacks 
    shown in latent space (\emph{colored curves}), with $\kappa=1$, $d=0$ (standard CW).
    The large black dot indicates the position of $\mu_\mathrm{target}$, the target class being `Harvestman (spider)'.
    The solid black lines are the decision boundaries to the  surrounding classes.
    The dashed black lines are the boundaries of the region where the classifier is fooled with sufficiently high confidence corresponding to $\kappa$.
    In the dotted section of the colored trajectories, the classifier is not yet fooled with sufficiently high confidence.
    In the solid section, the classifier has been fooled, and the attack only tries to reduce the perturbation.
    Below, the four perturbed images are shown, along with the absolute perturbation.
    More examples and detailed explanation in Appendix \ref{app:robustness_trajectories}.
    \vspace{-4mm}
    }
    \label{fig:adversarial_trajectory}
\end{figure}
\\[2.0mm]
\textbf{Handling Adversarial Attacks:}
We are interested in finding out if generative classifiers are more robust to adversarial attacks in the sense of (4). 
We are not proposing a new, competitive method of adversarial attack defense,  the goal is simply to examine whether GCs are naturally more robust to adversarial attacks on ImageNet, 
in the same way it was observed for e.g. MNIST previously \cite{li2018generative, schott2019towards}.
For this, we perform the well established `Carlini-Wagner' white-box targeted attack introduced in \cite{carlini2017towards},
which optimizes the following objective:
\vspace{-2mm}
\begin{equation}
    \mathcal{L}_\mathrm{CW} = \| x - \xatt \|^2 + c \cdot \mathcal{L}_\mathrm{class}^{(\kappa)}(y_\mathrm{target}),
\end{equation}\\[-5mm]
i.e. the attacked image $\xatt$ should be close to the original image $x$,
while being classified as a target class $y_\mathrm{target}$.
$\kappa$ is a hyperparameter that specifies how large the difference in logits should be between $y_\mathrm{target}$
and the next highest class, controlling how confident the classifier will be forced to be in its (wrong) decision.
When facing a model such as a GC, which can detect attacks, it is also possible to add an extra loss term $\mathcal{L}_\mathrm{detect}$ in order to fool the detection mechanism as well,
as proposed in \cite{carlini2017adversarial}:
\vspace{-1.5mm}
\begin{equation}
    \mathcal{L}_\mathrm{CWD} = \| x - \xatt \|^2 + c \cdot \mathcal{L}_\mathrm{class}^{(\kappa)}(y_\mathrm{target}) + d \cdot \mathcal{L}_\mathrm{detect}
\end{equation}\\[-5mm]
The full formulation of the attack objectives is given in Appendix \ref{app:robustness_adversarial_objectives}.

For evaluation, we examine standard CW attacks and two detection-fooling attacks with $d=66$ and $d=1000$, each for three values of $\kappa \in \{ 0.01, 1, \infty\}$.
For these 9 attack settings, we measure the L2 perturbation of the images after the attack and the ROC-AUC of the attack detection.

\begin{figure}
     \centering
     \includegraphics[width=\linewidth]{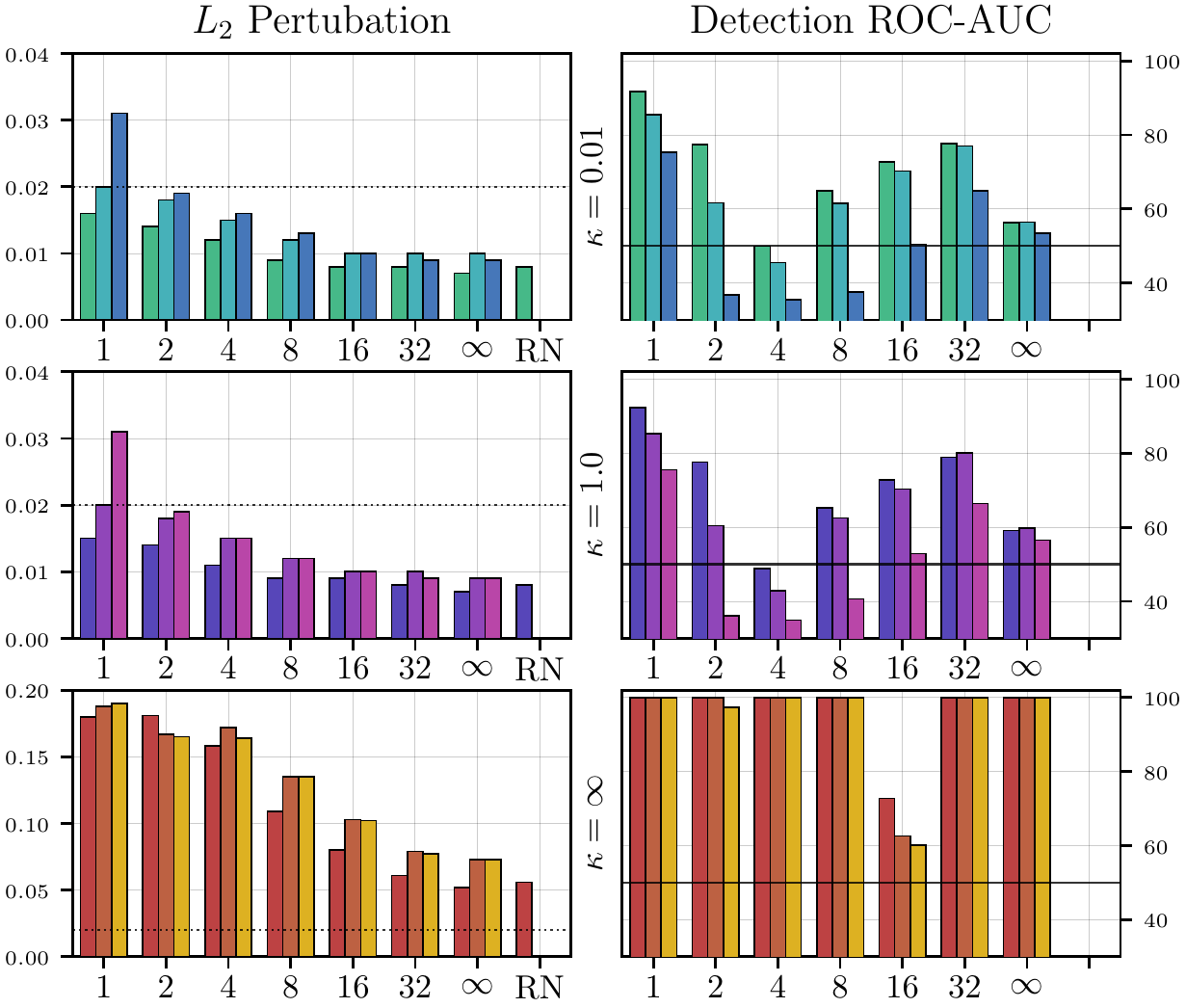}
     \caption{Behaviour of GCs under adversarial attacks.
     The first column of plots shows the mean perturbation, the second shows the detection ROC-AUC.
     The three rows of plots correspond to adversarial attacks with $\kappa = 0.01$ (any confidence for the target prediction is enough),
     $\kappa = 1$ (should have high confidence), and $\kappa = \infty$ (should be as confident as possible).
     The labels on the x-axis give the values of $\beta$, `RN' is a ResNet-50.
     The three bars for each $\beta$ correspond to: standard adversarial attack ($d=0$), $d=66$, and $d=1000$, 
     i.e. the detection mechanism is fooled at the same time as the prediction.
     The dotted line in the perturbation plots roughly indicates the level at which attacks are visible by eye.
     Note that this is subjective and only a rough indication.
     The line in the detection plots indicates random performance, i.e. the OoD detection does nothing useful.
     \vspace{-4mm}
     }
     \label{fig:adversarial_bar_graph}
\end{figure}

The results are presented in Fig \ref{fig:adversarial_bar_graph}, from which we make several key observations.
We conclude that the GC requires roughly $2\times$ larger perturbations for a standard adversarial compared to the ResNet, in line with \cite{li2018generative}.
We also observe the attack detection mechanism to be partially robust against attacks; even with $d=1000$ it still works reasonably well for some cases.
Furthermore, within this setting, the size of the perturbation is even more extensive compared to the standard attack setting.
Fooling the classifier to predict the wrong class with greater confidence also increases the necessary perturbation as the detectability.
The full results and additional measurements are provided in Appendix \ref{app:robustness_adversarial_results}.
An intuitive visualization of the adversarial attack in latent space is shown in  Fig.~\ref{fig:adversarial_trajectory}.

\vspace{-1.5mm}
\section{Conclusion}
\vspace{-1mm}
\noindent
In this work we have addressed the question of trustworthiness for image classification.
In the past, many properties linked with trustworthiness have been ascribed to generative classifier (GCs), such as increased robustness and explainability.
To the best of our knowledge, we are the first to design, successfully apply, and examine a GC for an application with real-world complexity, 
here the classification of the ImageNet dataset.
Our GC performs nearly on-par with a standard discriminative classifier (DC), here ResNet, when tuned for discriminative performance.
We observe that our GC offers significant improvements over standard DCs in terms of explainability and native out-of-distribution detection capability,
but does not automatically solve all aspects of trustworthiness:
Contrary to common belief,
it does not generalize better under image corruptions than a DC, and it does not fully prevent adversarial attacks.
In the future, we expect that robustness can be increased with further modifications or additional post-processing algorithms, as already exist for DCs.
Finally, we contribute downloadable GC models for ImageNet to the community.
These can serve as pre-trained GCs, much in the same way as pre-trained ResNet architectures do for discriminative classification.

\vspace{-1.5mm}
\section{Acknowledgements}
\vspace{-1mm}
\noindent
LA received funding by the Federal Ministry of Education and Research of Germany project High Performance Deep Learning Framework (No 01IH17002). 
CR and UK received financial support from the European Research Council (ERC) under the European Unions Horizon 2020 research and innovation program (grant agreement No 647769). 
We thank the Center for Information Services and High Performance Computing (ZIH) at Dresden University of Technology for generous allocations of computation time. 
Furthermore we thank our colleagues (alphabetically), Tim Adler, Felix Draxler, Jakob Kruse, Titus Leistner, Jens Mueller, and Peter Sorrenson for their help, support and fruitful discussions.

\FloatBarrier
{\small
\bibliographystyle{ieee_fullname}
\bibliography{ms}
}
\clearpage

\addtocontents{toc}{\protect\setcounter{tocdepth}{2}}
\twocolumn[
  \begin{@twocolumnfalse}
\begin{center}
{\Large \bf Generative Classifiers as a Basis for Trustworthy Image Classification \\
\Large \bf -- Appendix --}
\end{center}
\vspace*{10mm}
\FloatBarrier
 \end{@twocolumnfalse}
  ]

\tableofcontents
\appendix


\section{Methods -- Additional Materials}
\subsection{Out-of-Distribution Detection}
\label{app:ood_detectionn_method}
\begin{figure}
    \centering
    \includegraphics[width=\linewidth]{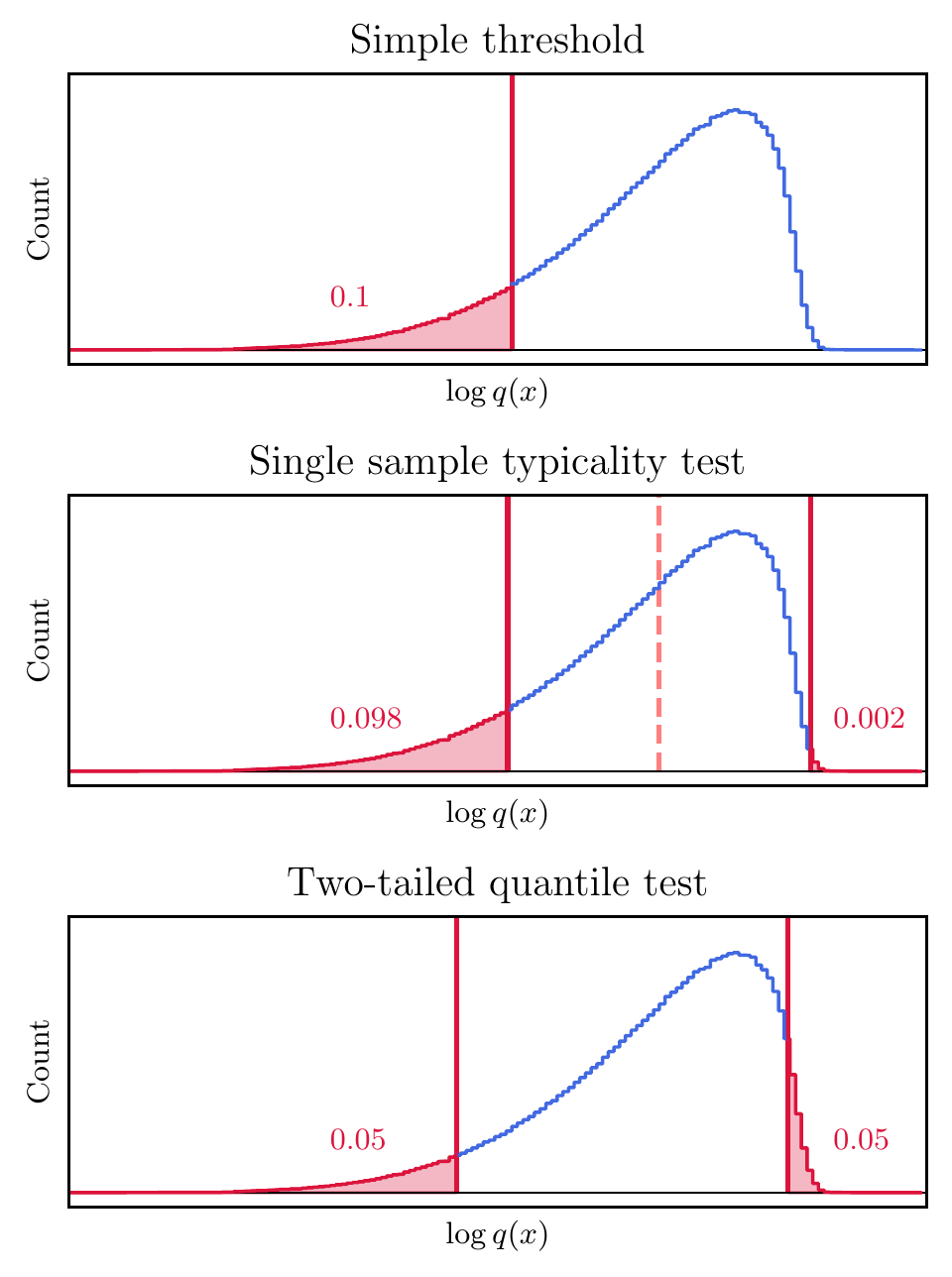}
    \caption{Illustration of three different OoD tests based on the estimated likelihood.
    The curve shows the distribution of likelihood scores in the training set.
    The blue part counts as in-distribution, and the red part as OoD.
    The threshold is chosen such that the red area (false positive rate, $p$-value), 
    is 0.1 in all three cases, for illustration.
    In practice, this would be chosen much lower, e.g. $0.001$.
    The small red numbers indicate the fraction of training samples above and below each threshold.
    }
    \label{fig:illst_ood_tests}
\end{figure}

Originally, ony a single threshhold on the learned likelihood was used to detect OoD inputs,
which fails in several cases where the likelihoods of the inputs are unnaturally high \cite{nalisnick2018deep}.
As a way to correct for this, the typicality test \cite{nalisnick2019detecting} uses both an upper and a lower threshold, 
centered symmetrically around the mean log-likelihood of the training data (Fig.~\ref{fig:illst_ood_tests}, middle).
For our ImageNet models, we observe that the distribution of log-likelihood values in the training set is highly asymmetrical (see Fig.~\ref{fig:illst_ood_tests}).
Therefore, we introduce our third possibility, a two-tailed quantile test.
Instead of the thresholds being symmetric around the mean, they are chosen so that an equal mass of the log-likelihood histogram 
lies above the upper and below the lower threshold (Fig.~\ref{fig:illst_ood_tests}, middle).
In practice, we only measure minor differences in performance between the single-sample typicality test and the two-tailed quantile test.

All three tests can also be seen as hypothesis tests, with the null hypothesis being that the input is in-distribution. 
The $p$-value for the hypothesis test is the fraction of training samples with scores in the OoD-zone, which also equals the false positive rate.
To evaluate the OoD detection capabilities, we do not use a single threshold value, but want a measure that is independent of it.
This is because the acceptable false negative/false positive trade-off depends on the context/application that the model is used in.
By varying the $p$-value of the test, we produce a receiver operating characteristic (ROC) curve.
The area under this curve (ROC-AUC), in percent, serves as a scalar measurement of the OoD detection capabilities.
An ROC-AUC of 100\% means that the OoD samples are perfectly separated from the in-distribution samples and can always be identified correctly.
A value of 50\% indicates that the test performs exactly as well as randomly deciding.
Below 50\%, worse than random performance, the OoD data appears to be more in-distribution as a significant fraction of the training data itself.

\section{Experiments -- Additional Materials}

\subsection{Network Architecture}
\label{app:architecture_and_training}

In the following, we outline the design choices and training procedure used for training the INN model as a GC on the ImageNet dataset.
It has been noted in the past that there are strong parallels between ResNet residual blocks \cite{he2016deep} and INN affine coupling blocks \cite{dinh2016density}, described further below.
In fact, under some additional constraints, standard ResNet residual blocks can also be numerically inverted \cite{behrmann2018invertible}.
Therefore, a standard ResNet is not only the most fitting comparison to our GC, but also informs many of our design choices.
The argument is, that ResNets contain many carefully tested design choices, leading to their excellent discriminative performance.
Adopting these choices where possible saves us from performing an infeasible number of ablations and comparisons ourselves,
and still achieve relatively good performance empirically.

\begin{figure}
    \resizebox{\linewidth}{!}{\newcommand{\ColorMerge}{PineGreen!40}
\newcommand{\ColorSubnet}{Black!5}
\newcommand{\ColorPermute}{Dandelion!40}
\newcommand{\ColorAffine}{OrangeRed!40}
\newcommand{\ColorCh}{RubineRed!80!Black}

\newcommand{\ColorConvolution}{Plum!40!White}
\newcommand{\ColorRelu}{VioletRed!40!White}

\newcommand\SideOffs{2.5cm}
\newcommand\DY{}

\tikzset{%
  >={Latex[width=2mm,length=2mm]},
            layer/.style = {rectangle, rounded corners=0.5mm, draw=black,
                           minimum width=7cm, minimum height=1cm,
                           text centered, font=\large\sffamily},
            op/.style = {circle, draw=black, font=\large\sffamily,
                           minimum width=1.7cm, minimum height=1.7cm},
            subnet/.style = {rectangle, rounded corners, draw=black, font=\large\sffamily,
                           minimum width=2.0cm, minimum height=2.0cm},
            connec/.style = {->, very thick},
            annot/.style = {font=\sffamily, text=\ColorCh},
            connecZoomed/.style = {->, line width=1.5mm, >={Latex[width=3mm,length=3mm]}}
}

\begin{tikzpicture}[node distance=1.7cm,
    every node/.style={font=\sffamily}, align=center]

    \begin{scope}
        \node (split) [layer, fill=\ColorMerge] {Split channels};
        \node (concat) [layer, fill=\ColorMerge, above of=split, yshift=8cm] {Concatentate channels};
        \node (global) [layer, fill=\ColorAffine, above of=concat] {Global affine};
        \node (permute) [layer, fill=\ColorPermute, above of=global] {Soft permute};

        \node (input) [below of=split]{\large Block input~~\normalsize \color{\ColorCh} 48};
        \node (output) [above of=permute]{\large Block output~~\normalsize \color{\ColorCh} 48};

        \node (affine) [op, fill=\ColorAffine, above of=split, yshift=5cm, xshift=\SideOffs] {Affine};
        \node (sub) [subnet, fill=\ColorSubnet, above of=split, yshift=2cm] {Coupling \\ subnet};

        \node[annot, align=right, left of=affine, yshift=-6mm, xshift=3mm] 
            {\color{black}$s,t$ \\ \color{\ColorCh} 24+24}; 
        \node[annot, below of=sub, yshift=2mm, xshift=3mm] {24}; 
        \node[annot, below of=concat, yshift=3mm, xshift=\SideOffs-3mm] {24}; 
        \node[annot, below of=concat, yshift=-50mm, xshift=-\SideOffs+3mm] {24}; 
        \node[annot, below of=concat, yshift=-50mm, xshift=\SideOffs-3mm] {24}; 
        \node[annot, below of=concat, yshift=-50mm, xshift=\SideOffs-3mm] {24}; 

        \node[annot, above of=concat, yshift=-9mm, xshift=3mm] {48}; 
        \node[annot, above of=global, yshift=-9mm, xshift=3mm] {48}; 

        \draw[connec] (input) -- (split);
        \draw[connec] ($(split.north) + (-\SideOffs, 0)$) 
                      -- ($(concat.south) + (-\SideOffs, 0)$);
        \draw[connec] ($(split.north) + (-\SideOffs, 0)$) 
                      to[out=90, in=270] (sub.south);
        \draw[connec] (sub.north) to[out=90, in=180] (affine.west);
        \draw[connec] ($(split.north) + (\SideOffs, 0)$) -- (affine);
        \draw[connec] (affine) -- ($(concat.south) + (\SideOffs, 0)$);
        \draw[connec] (concat) -- (global);
        \draw[connec] (global) -- (permute);
        \draw[connec] (permute) -- (output);
    \end{scope}

    \begin{scope}[xshift=10cm, yshift=0.9cm]
        \newcommand{\NodeGrouping}{-3mm}

        \draw[rounded corners=2mm, draw=black, fill=\ColorSubnet] (-4.2cm,-1.4cm) rectangle (4.2cm, 12.5cm);

        \node (bn1) [layer, fill=\ColorRelu] {Batch norm.~+ ReLU};
        \node (conv1) [layer, fill=\ColorConvolution, above of=bn1, yshift=\NodeGrouping] 
            {$1 \times 1$ Conv.~\textcolor{\ColorCh}{24} $\rightarrow$ \textcolor{\ColorCh}{64}};

        \node (bn2) [layer, fill=\ColorRelu, above of=conv1] {Batch norm.~+ ReLU};
        \node (conv2) [layer, fill=\ColorConvolution, above of=bn2, yshift=\NodeGrouping] 
            {$3 \times 3$ Conv.~\textcolor{\ColorCh}{64} $\rightarrow$ \textcolor{\ColorCh}{64}};

        \node (bn3) [layer, fill=\ColorRelu, above of=conv2] {Batch norm.~+ ReLU};
        \node (conv3) [layer, fill=\ColorConvolution, above of=bn3, yshift=\NodeGrouping] 
            {$1 \times 1$ Conv.~\textcolor{\ColorCh}{64} $\rightarrow$ \textcolor{\ColorCh}{256}};

        \node (bn4) [layer, fill=\ColorRelu, above of=conv3] {Batch norm.~+ ReLU};
        \node (conv4) [layer, fill=\ColorConvolution, above of=bn4, yshift=\NodeGrouping] 
            {$1 \times 1$ Conv.~\textcolor{\ColorCh}{256} $\rightarrow$ \textcolor{\ColorCh}{24 + 24}};

        \node[below of=bn1, yshift=-3mm] (input) {};
        \node[above of=conv4, yshift=10mm] (output) {};
        \node[above of=conv4, xshift=22mm, yshift=-4mm, text=\ColorSubnet!Black] {\Large Coupling subnet};

        \draw[connecZoomed] (input) -- (bn1);
        \draw[connecZoomed] (bn1) -- (conv1);
        \draw[connecZoomed] (conv1) -- (bn2);
        \draw[connecZoomed] (bn2) -- (conv2);
        \draw[connecZoomed] (conv2) -- (bn3);
        \draw[connecZoomed] (bn3) -- (conv3);
        \draw[connecZoomed] (conv3) -- (bn4);
        \draw[connecZoomed] (bn4) -- (conv4);
        \draw[connecZoomed] (conv4) -- (output);
    \end{scope}

\end{tikzpicture}}
    
    \caption{
    Illustration of the coupling blocks used, as well as the structure of the subnetworks used to predict the affine components.
    The purple numbers indicate the number of feature channels, given as an example for the fist resolution level (see Table \ref{tab:dims_and_features}, \emph{Conv\_2\_x}).
    }
    \label{fig:coupling}
\end{figure}
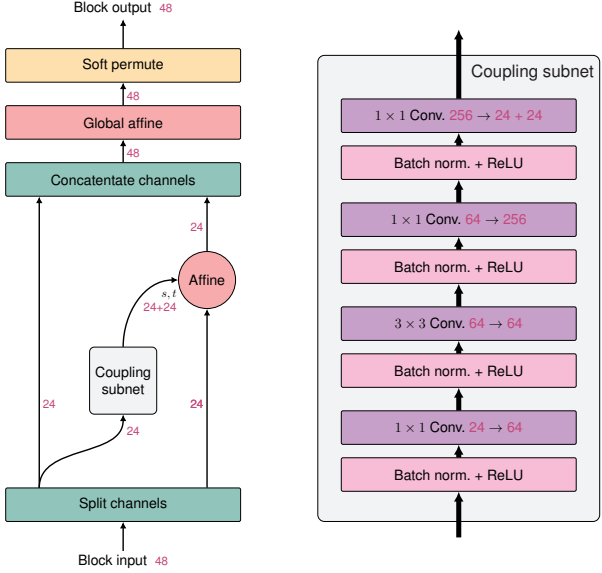

\paragraph{Affine coupling operation.}
As a basic building block of our network, we use the affine coupling block
shown in Fig.~\ref{fig:coupling}.
Such blocks were fist introduced in \cite{dinh2016density}, and are exactly and cheaply invertible,
as well as having a tractable Jacobian determinant.
The incoming features are first split in two halves, say $u_1$ and $u_2$, along the channel dimension.
The first half $u_1$ is not changed, and passed straight through. 
A subnetwork, similar to the residual subnetwork of a ResNet then predicts affine coefficients $s,t$ from $u_1$,
which are used to perform an affine transformation on the other half of the features $u_2$.
This gives us outputs $v_1$, $v_2$:
\begin{equation}
   v_2 = s(u_1) \odot u_2 + t(u_1) \quad \text{and} \quad v_1 = u_1
\end{equation}
To invert this operation given only $v_1, v_2$, note that $u_1 = v_1$ is trivially available, so the same coefficients $s, t$ can be re-computed for the inverse.
With these, the affine transformation itself can be analytically inverted, to get back $u_2 = (v_2 - t(v_1)) \oslash s(v_1)$.
To guarantee invertibility, we restrict $s(\cdot) > 0$. 
In theory, $s(\cdot) \neq 0$ suffices, but this complicates the situation and does not improve expressive power:
mirroring an output dimension is irrelevant for the network and the loss.
We ensure $s(\cdot) > 0$ by using $\exp(\alpha \, \mathrm{tanh}(\cdot))$ activation on the $s$-outputs of the subnetwork, as previously in \cite{dinh2016density},
where $\alpha$ is a fixed hyperparameter. In principle, $\exp$ alone would be enough, but this leads to instabilities during training, as it can become infinitely large.
Importantly to note, the subnetwork itself never has to be inverted, and is always computed forward. 
Therefore, it can contain the usual operations such as convolutions or batch normalization.

To compare, in a standard ResNet block, a copy operation is used instead of the split, and a simple addition is performed in place of the affine transformation. 
Apart from this, the structure is very similar.

\paragraph{Complete coupling blocks.}
The expressive power of the affine coupling above is insufficient: 
half the data is not touched at all, and the remaining varibles can only be scaled up/down by a factor of at most $\exp(\pm \alpha)$.
We add two more invertible operations to solve these problems:
We first perform a global channel-wise affine transformation to all variables with scaling $s_\text{global}$ and bias $t_\text{global}$.
This technique was already proposed in \cite{dinh2016density} and refined in \cite{kingma2018glow} as `ActNorm'.
Note that in feed-forward networks, this is also often done as part of the batch normalization layers.
Again, $s_\text{global}$ must be positive, and we achieve the best results choosing $s_\text{global} = s_0 \mathrm{softplus}(\gamma) = s_0 \log( 1 + e^\gamma )$.
Here, $\gamma$ and $t_\text{global}$ are learned directly as free parameters, and $s_0$ is a scalar hyperparameter which we fix to $0.1$,
while $\gamma$ is initialized to $10$.

Secondly, we want to use a different split in the next block, and therefore have to apply some invertible operation that mixes the channels.
So far, there is no `default' approach to this in the INN literature. 
Various methods exist, such as simply swapping the two halves \cite{dinh2014nice}, learned householder reflections \cite{tomczak2016improving}, 
fixed permutations \cite{ardizzone2018analyzing}, and learning unconstrained mixing matrices \cite{kingma2018glow}, among others.
While it is desirable to use a learned mixing operations, we do not find any benefits in practice.
The method used for \cite{kingma2018glow} has no guaranteed invertibility, and the training can simply crash when the matrix becomes singular.
The householder matrices from \cite{tomczak2016improving} quickly become computationally expensive with many reflections,
and in our case bring no empirical benefit over fixed (not learned) mixing.
Instead, we use a random orthogonal matrix from the $O(N)$ Haar distribution after each coupling block, that stays fixed during training.
This encourages more mixing than a simple hard permutation, and empirically gives the best results with our architecture.

With an orthogonal mixing matrix, the overall log-Jacobian-determinant of one coupling block can be shown to be
\begin{equation}
    \log \big| \det(J) \big| = \sum \log s(u_1) + \sum \log s_\text{global}.
\end{equation}
Due to the chain rule, and product decomposition of the determinant, the sum of the log-Jac-det of each coupling block
will give the log-Jac-det of the entire network. An illustration of a coupling block is given in Fig.~\ref{fig:coupling}, left.

\paragraph{Subnetworks.}

We adopt the ResNet design choices for building the affine subnetworks, with one modification:
we add an additional 1x1 projection layer as the final output. 
This is motivated by the fact that the INN has less feature maps than the ResNet for all but the last resolution level.
Therefore, the expressive power would be limited by only having this few output channels for the final convolution. 
The subnetwork design is shown in Fig.~\ref{fig:coupling}, right.

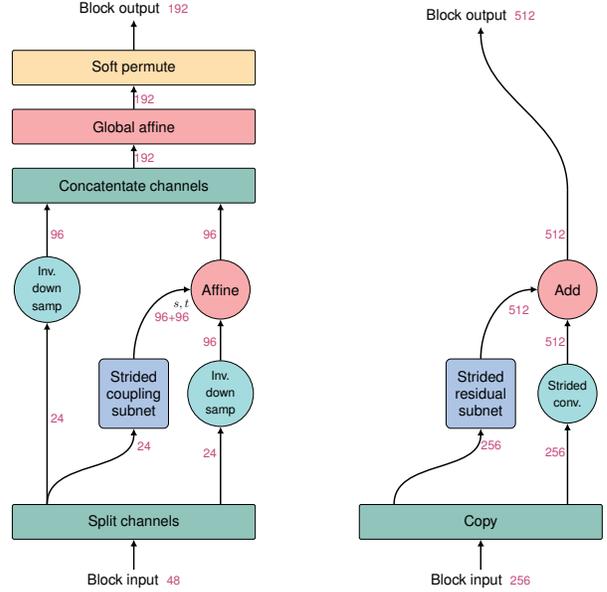
\begin{figure}
    \centering
    \resizebox{\linewidth}{!}{\newcommand{\ColorMerge}{PineGreen!40}
\newcommand{\ColorSubnet}{RoyalBlue!30}
\newcommand{\ColorPermute}{Dandelion!40}
\newcommand{\ColorAffine}{OrangeRed!40}
\newcommand{\ColorDownsamp}{BlueGreen!40}
\newcommand{\ColorCh}{RubineRed!80!Black}

\newcommand{\ColorConvolution}{Plum!40!White}
\newcommand{\ColorRelu}{VioletRed!40!White}

\newcommand\SideOffs{2.5cm}
\newcommand\DY{}

\tikzset{%
  >={Latex[width=2mm,length=2mm]},
            layer/.style = {rectangle, rounded corners=0.5mm, draw=black,
                           minimum width=7cm, minimum height=1cm,
                           text centered, font=\large\sffamily},
            op/.style = {circle, draw=black, font=\large\sffamily,
                           minimum width=1.7cm, minimum height=1.7cm},
            subnet/.style = {rectangle, rounded corners, draw=black, font=\large\sffamily,
                           minimum width=2.0cm, minimum height=2.0cm},
            connec/.style = {->, very thick},
            annot/.style = {font=\sffamily, text=\ColorCh},
            connecZoomed/.style = {->, line width=1.5mm, >={Latex[width=3mm,length=3mm]}}
}

\begin{tikzpicture}[node distance=1.7cm,
    every node/.style={font=\sffamily}, align=center]

    \begin{scope}
        \node (split) [layer, fill=\ColorMerge] {Split channels};
        \node (concat) [layer, fill=\ColorMerge, above of=split, yshift=8cm] {Concatentate channels};
        \node (global) [layer, fill=\ColorAffine, above of=concat] {Global affine};
        \node (permute) [layer, fill=\ColorPermute, above of=global] {Soft permute};

        \node (input) [below of=split]{\large Block input~~\normalsize \color{\ColorCh} 48};
        \node (output) [above of=permute]{\large Block output~~\normalsize \color{\ColorCh} 192};

        \node (affine) [op, fill=\ColorAffine, above of=split, yshift=5cm, xshift=\SideOffs] {Affine};
        \node (downR) [op, fill=\ColorDownsamp, above of=split, yshift=2cm, xshift=\SideOffs]{\normalsize Inv. \\\normalsize down\\\normalsize samp};
        \node (downL) [op, fill=\ColorDownsamp, above of=split, yshift=5cm, xshift=-\SideOffs]{\normalsize Inv. \\\normalsize down\\\normalsize samp};

        \node (sub) [subnet, fill=\ColorSubnet, above of=split, yshift=2cm] {Strided \\ coupling \\ subnet};

        \node[annot, align=right, left of=affine, yshift=-6mm, xshift=3mm] 
            {\color{black}$s,t$ \\ \color{\ColorCh} 96+96}; 
        \node[annot, below of=sub, yshift=2mm, xshift=3mm] {24}; 
        \node[annot, below of=concat, yshift=3mm, xshift=\SideOffs-3mm] {96}; 
        \node[annot, below of=concat, yshift=3mm, xshift=-\SideOffs+3mm] {96}; 

        \node[annot, below of=concat, yshift=-50mm, xshift=-\SideOffs+3mm] {24}; 
        \node[annot, below of=concat, yshift=-60mm, xshift=\SideOffs-3mm] {24}; 
        \node[annot, below of=concat, yshift=-28mm, xshift=\SideOffs-3mm] {96}; 

        \node[annot, above of=concat, yshift=-9mm, xshift=3mm] {192}; 
        \node[annot, above of=global, yshift=-9mm, xshift=3mm] {192}; 

        \draw[connec] (input) -- (split);
        \draw[connec] ($(split.north) + (-\SideOffs, 0)$) -- (downL);
        \draw[connec] (downL) -- ($(concat.south) + (-\SideOffs, 0)$);
        \draw[connec] ($(split.north) + (-\SideOffs, 0)$) 
                      to[out=90, in=270] (sub.south);
        \draw[connec] (sub.north) to[out=90, in=180] (affine.west);
        \draw[connec] ($(split.north) + (\SideOffs, 0)$) -- (downR);
        \draw[connec] (downR) -- (affine);
        \draw[connec] (affine) -- ($(concat.south) + (\SideOffs, 0)$);
        \draw[connec] (concat) -- (global);
        \draw[connec] (global) -- (permute);
        \draw[connec] (permute) -- (output);
    \end{scope}

    \begin{scope}[xshift=10cm]
        \node (split) [layer, fill=\ColorMerge] {Copy};

        \node (concat) [above of=split, yshift=8cm] {};

        \node (input) [below of=split]{\large Block input~~\normalsize \color{\ColorCh} 256};
        \node (output) [above of=concat, yshift=3.2cm]{\large Block output~~\normalsize \color{\ColorCh} 512};

        \node (downR) [op, fill=\ColorDownsamp, above of=split, yshift=2cm, xshift=\SideOffs]{\normalsize Strided \\ \normalsize conv.};
        \node (affine) [op, fill=\ColorAffine, above of=split, yshift=5cm, xshift=\SideOffs] {Add};
        \node (sub) [subnet, fill=\ColorSubnet, above of=split, yshift=2cm] {Strided \\ residual \\ subnet};

        \node[annot, align=right, left of=affine, yshift=-6mm, xshift=3mm] {512}; 
        \node[annot, below of=sub, yshift=2mm, xshift=3mm] {256}; 
        \node[annot, below of=concat, yshift=3mm, xshift=\SideOffs-3.5mm] {512}; 
        \node[annot, below of=concat, yshift=-60mm, xshift=\SideOffs-3.5mm] {256}; 
        \node[annot, below of=concat, yshift=-28mm, xshift=\SideOffs-3.5mm] {512}; 

        \draw[connec] (input) -- (split);
        \draw[connec] ($(split.north) + (-\SideOffs, 0)$) 
                      to[out=90, in=270] (sub.south);
        \draw[connec] (sub.north) to[out=90, in=180] (affine.west);
        \draw[connec] ($(split.north) + (\SideOffs, 0)$) -- (downR);
        \draw[connec] (downR) -- (affine);
        \draw[very thick] (affine) -- ($(concat.south) + (\SideOffs, 0)$);
        \draw[connec] ($(concat.south) + (\SideOffs, 0)$) to[out=90, in=270] (output);
    \end{scope}

\end{tikzpicture}}
    \caption{Illustration of our downsampling coupling blocks (\emph{left}), 
    compared to the standard ResNet downsampling blocks (\emph{right}).
    The invertible downsampling operation (blue circles) reorders inputs in a checkerboard pattern as in \cite{jacobsen2018irevnet}.}
    \label{fig:down_coupling}
\end{figure}

\paragraph{Downsampling blocks.}
In the past, various invertible downsampling operations have been used, e.g.~\cite{dinh2016density, jacobsen2018irevnet, ardizzone2019guided}.
Notably, none of these have a learnable component, such as strided convolutions.
Instead, we introduce a \emph{downsampling coupling block}, as a natural extension of the downsampling residual blocks present at the end of each ResNet section.
Shown in more detail in Fig.~\ref{fig:down_coupling}, we use two of the invertible re-ordering and re-shaping operations from \cite{jacobsen2018irevnet}, but nested within a single coupling block. 
This way, the subnetwork can make use of a strided $3\times 3$ convolution as a learned component to the downsampling.
Note that we did not perform rigorous ablations of this introduction, and chose it mainly for better conformity to standard ResNets.

\paragraph{Network layout.}

The overall network layout is the same as for the standard ResNet-50, which offers a good trade-off between performance and model complexity.
The input images are immediately downsampled twice, once using a downsampling coupling block with a $7\times 7$ convolution,
then with a Haar wavelet transform as in \cite{ardizzone2019guided}.
The ResNet analogue is the so-called \emph{entry flow}, which also uses a strided $7\times 7$ convolution and a max-pooling operation.
A series of coupling blocks follow this, with downsampling blocks distributed throughout, chosen in the same way as for the ResNet-50,
detailed in Tab.~\ref{tab:dims_and_features}.
The output of the INN consists of 3072 two dimensional feature maps at a resolution of $7\times 7$ (compared to 2048 feature maps for the ResNet).

\begin{table}
    \centering
    \resizebox{\linewidth}{!}{

\begin{tabular}{l|r|r|r|r|r|r}
\textbf{Layer} & \textbf{Blocks} & \textbf{Im. size} & \multicolumn{2}{c|}{\textbf{Channels}} & \multicolumn{2}{c}{\textbf{R.F.}} \\
      &        &          & INN  & ResNet                & INN & ResNet   \\
    \hline

Input & & 224 & 3 & 3 & & \\
Entry flow & 1 & 112 & 12 & 64 & 8 & 6\\
Pool (Haar/max) & & 56 & 48 & 64 & 10 & 10 \\
    \hline 
Conv\_2\_x & 3 & 56 & 48 & 256 &  34 &  34 \\
Conv\_3\_x & 4 & 28 & 192 & 512 & 106 &  90 \\
Conv\_4\_x & 6 & 14 & 768 & 1024 & 314 &  266 \\
Conv\_5\_x & 3 & 7 & 3072 & 2048 & 538 &  426 \\
    \hline 
Pool (DCT/avg.) & & 1 & 150\,528 & 2048 & $\infty$  & $\infty$ \\
\end{tabular}}
    \vspace{2mm}
    \caption{For each of the resolution levels in the INN and ResNet-50,
    the number of coupling/residual blocks and spatial size is given,
    along with the number of feature channels and the maximum possible receptive field (R.F.).}
    \label{tab:app_dims_and_features}
\end{table}

\begin{figure}
\includegraphics[width=\linewidth]{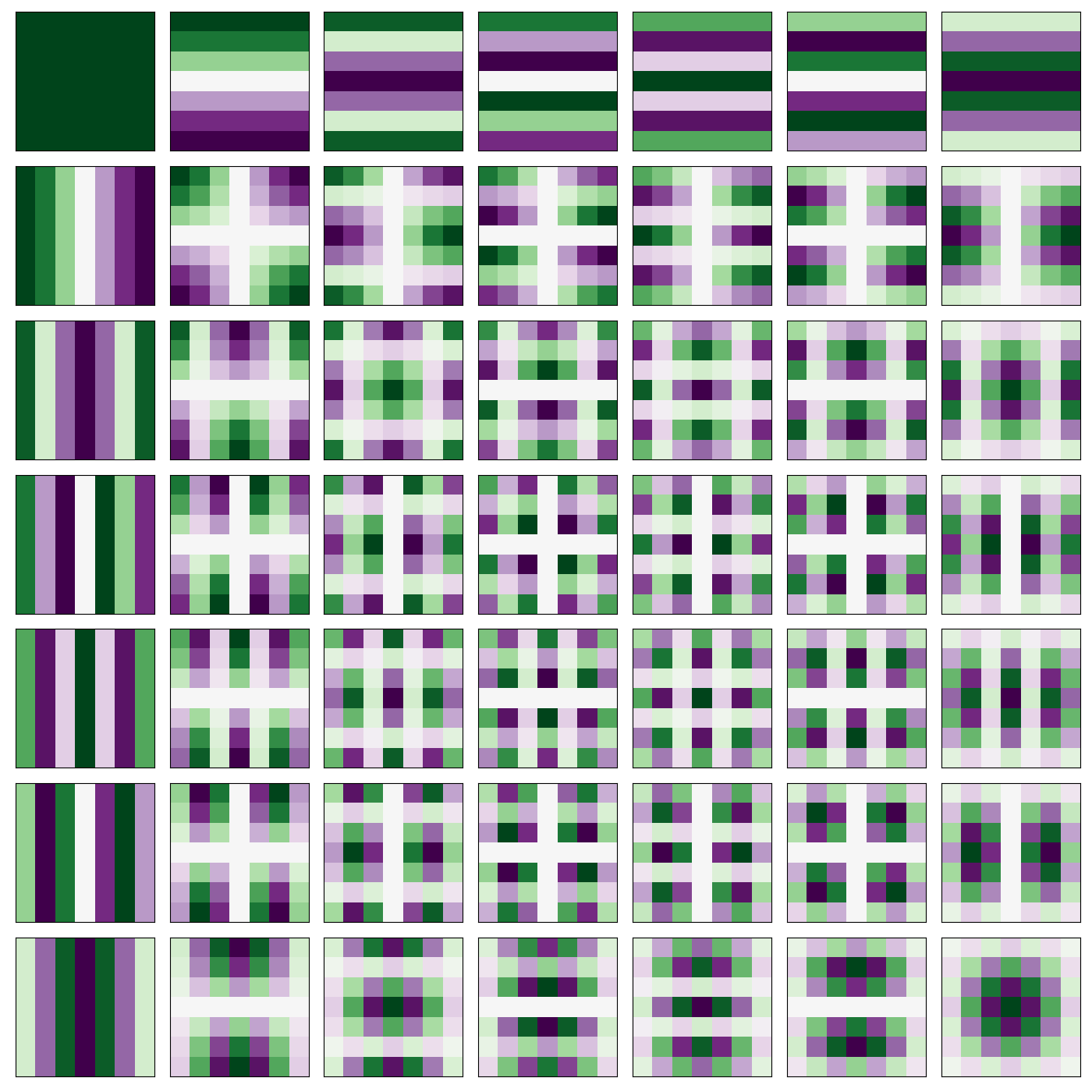}
\caption{Each $7 \times 7$ feature map is transformed to $49$ orthogonal output features, using the DCT coefficients shown above.
\textcolor{Green!80!black}{Green $> 0$},
\textcolor{Purple!80!black}{Purple $< 0$},
\textcolor{black!40!white}{white $= 0$}.
The top left most output feature is equal to the mean pooling operation.
}
\label{fig:dct_coeffs}
\end{figure}

In the ResNet, the output feature maps are passed through a global mean pooling operation.
As explained in \cite{jacobsen2018excessive}, a discrete cosine transform (DCT) presents the best invertible alternative to this:
From our 3072 feature maps, the DCT also produces mean pooled outputs, along with $48$ other outputs per feature  map,
that encode the remaining information.
The DCT coefficients are visualized in Fig.~\ref{fig:dct_coeffs}.
As a final step, the ResNet performs a linear projection to the 1000 logits.
The analogous operation for the INN is taking the distance of the output $z$ to each of the 1000 cluster centers.

\begin{figure}
\centering
\includegraphics[width=0.7\linewidth]{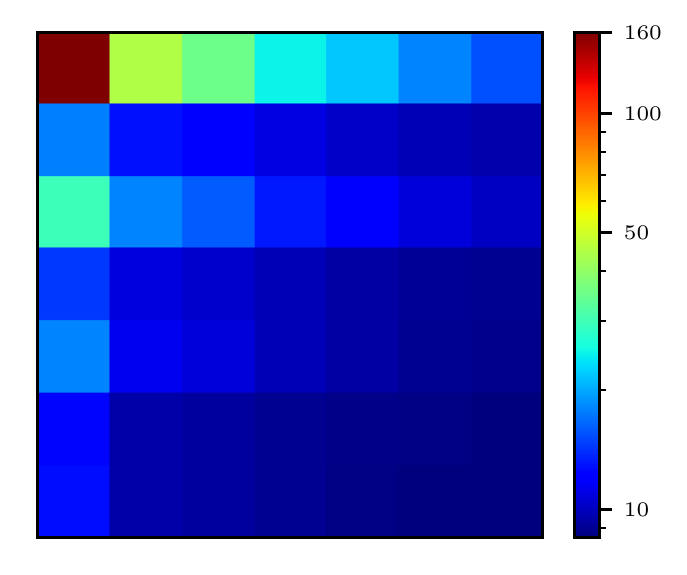}
\caption{For the 49 DCT components images shown in Fig.~\ref{fig:dct_coeffs},
the mean spread of the corresponding entries of $\mu_y$ across classes is shown.
Intuitively, this is how much each DCT component contributes to classification.
A value of 0 means that these dimensions do not affect the classification at all.
The mean pooled component has by far the largest influence,
and the contribution of the high order components (bottom right) is negligible.
Due to the random horizontal flip augmentation, the horizontally anti-symmetric components hardly contribute (alternating rows).
}
\vspace{2mm}
\label{fig:dct_influence}
\end{figure}

\paragraph{Low-rank $\boldsymbol{\mu_y}$.}
If each entry of $\mu_y$ is learned independently, the total number of parameters for $D$-dimensional latent space and $M$ classes will be 
$DM\approx 150\,\text{Mil}$ for ImageNet. 
This is completely impractical, as $\mu_y$ alone would make up the majority of network parameters, which will only lead to overfitting.
We solve this by dividing up $\mu_y$ into two parts, corresponding to the mean-pooled and the higher-order DCT variables:
$\mu_y = [\mu_{\text{mean},y}, \; \mu_{\text{rest},y}]$.
We freely learn all approx.~$3\,\mathrm{Mil}$ parameters of $\mu_{\text{mean},y}$,
and choose a low-rank representation for the remaining $\mu_{\text{rest},y}$, using $K$ prototype vectors $\mu_k$:
\begin{equation}
    \mu_{\text{rest},y} = \sum_{k=1}^K \alpha_{yk} \mu_k
\end{equation}
Both $\mu_k$ and $\alpha_{yk}$ are learned.
This reduces the number of parameters to $D_\mathrm{mean}M + K (D_\mathrm{rest} + M)$.
Choosing $K=128$ empirically gives the best validation performance, 
and results in approx.~$19\,\mathrm{Mil}$ parameters, almost a factor of 10 less than the full $DM$.
However, it is important to note that this is still much more than the fully connected layer of a standard ResNet, with approx.~$2\,\mathrm{Mil}$ parameters.
This indicates it might be possible to find an even more efficient representation of $\mu_y$ without sacrificing performance.
The influence of each component of the low-rank $\mu_y$ is shown in Fig.~\ref{fig:dct_influence}. 
While $\mu_{\text{mean},y}$ contributes by far the most, training without $\mu_{\text{rest},y}$ entirely (setting it to zero),
degrades the validation top-1 prediction performance by several percentage points.

\paragraph{Data augmentation and training.}
As data augmentation, we perform the usual random crops and horizontal flips, with two additions:
Firstly, as is standard practice with normalizing flows specifically, 
we add uniform noise with amplitude $1/255$ to the images, to remove the quantization.
This is necessary when training with the Jacobian, as the quantization otherwise leads to problems.
Secondly, we use label smoothing \cite{szegedy2016rethinking} with $\alpha = 0.05$.
This is necessary to prevent the mixture centroids from drifting further and further apart:
training with perfectly hard labels makes the implicit assumption that all class components are infinitely separated.

The training scheme is the same as for the standard ResNet \cite{he2016deep}:
we use the SGD optimizer with a momentum of $0.9$ and the weight decay set to $0.0001$. We set the initial learning rate slightly lower to $0.07$ compared to $0.1$ for the original ResNet.
We also perform two subsequent cooling steps whenever the loss plateaus, decreasing the learning rate by a factor of $10$ each time.
The batch size is 64 per GPU, training on 6 GPUs.

The constraint of invertibility is associated with an extra cost of parameters and computation cost compared to a purely feed-forward network.
Table \ref{app:tab:computational_cost} summarizes this in comparison to a standard ResNet-50.
Both in terms of network parameters, as well FLOPs needed for one forward pass of the network,
the cost of the INN is about twice as high as the ResNet.
We are optimistic this overhead can be reduced in the future with more efficient INN architectures.

\begin{table}
    \begin{center}
    \begin{tabular}{l | r | r  }
     & ResNet & INN \\
     \hline
Network parameters (M) & 23.5 & 55.4  \\
All parameters (M)     & 25.6 & 77.5  \\
FLOPs (G)          & 4.07 & 9.08
\end{tabular}
    \end{center}
    \caption{Number of parameters and computational cost for each model.
    \emph{`Network parameters'} only counts the coupling/residual blocks.
    \emph{`All parameters'} additionally includes the fully connected output layer of the ResNet,
    and the parametrization of $\mu_y$ for the INN.
    The (M) and (G) indicates Mega and Giga respectively.
    For FLOPs, the fused multiply-add instruction (FMA) is counted as a single FLOP, as it is commonly a single instruction in modern computing architectures.
    }
    \label{app:tab:computational_cost}
\end{table}

\subsection{Receptive Field}
\label{app:receptive_field}
While the maximum possible receptive field (RF) of the INN and a standard trained ResNet are roughly comparable (see Table \ref{tab:dims_and_features}), 
we see large differences in the effective RF.
For the effective RF, we pick a feature space column $u$, before the DCT pooling operation.
Meaning, from the $H\times W\times 3072$ feature space, $u$ will be the $1\times1\times3072$ column.
We choose a column from the center to avoid interactions with the edges.
We call the individual features $u_l$ ($l=1\dots3072$).
We now measure the gradient w.r.t. each channel of each image pixel $x_{ijk}$, for real input images. 
The pixel position is $ij$, and the color channel is $k$.
We define the `sensitivity' of the model at each position as the $L_1$ norm of the gradient of the features w.r.t.~that input position, averaged over images from the test set:
\begin{equation}
\mathrm{Sensitivity}(i,j) = \mathbb{E}_{x\in \mathrm{test}} \left[ \; \sum_{k = 1}^3 \sum_{l = 1}^{3072} \left| \frac{\partial u_{l}}{x_{ijk}} \right| \;\right]
\end{equation}
There are other definitions that would be equally sensible (squared gradients, frobenius norm, etc.), 
but the results always show the same behaviour.

\begin{figure}
    \centering
    \includegraphics[width=\linewidth]{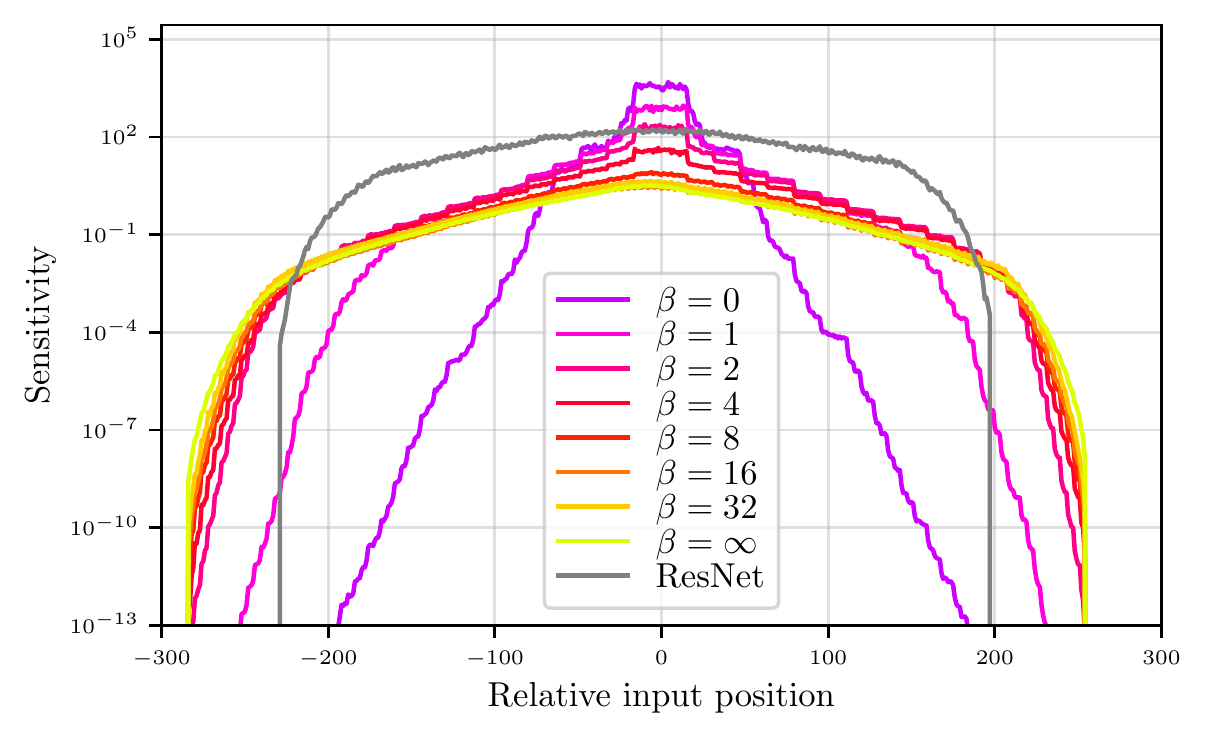}
    \caption{Effective receptive field for each value of $\beta$, just before the final pooling operation.
    Note the logarithmic sensitivity axis.}
    \label{app:fig:eff_receptive_field}
\end{figure}

The cross-sectional shape of this represents the effective RF, and is shown in Fig.~\ref{app:fig:eff_receptive_field}.
We observe that for low $\beta$, the effective RF is very narrow. 
In fact it is almost as narrow as it could possibly be: 
for $\beta \leq 4$, the FWHM of the sensitivity is only 64 pixels.
This is the same we would get from only the downsampling steps, without any spatial convolutions (with 6 downsamplings, $2^6 = 64$).
This could indicate that for the likelihood estimation, local details and structures are more important than any long-range features.
For higher values of $\beta$, the response more closely matches that of a standard trained ResNet 
(1.25 times wider in line with the 1.25 times larger maximum possible RF).

\subsection{Calibration Error}
\label{app:calibration_error_results}
The calibration of a model measures the truthfulness of the predictive posteriors.
In short, if we consider predictions where the model is e.g.~80\% confident in a class,
we would expect the prediction to be correct 80\% of the time.
If it were correct more often, it would be underconfident, and vice versa, more commonly,
if it were correct in much fewer than 80\% of cases, it would be overconfident.
Plotting the fraction of correct predictions $R$ over the binned confidence $C$ of predictions gives the so-called calibration curve $R(C)$.
For a perfectly calibrated model, the curve will follow the diagonal, but usually the behaviour deviates.

To quantitatively measure the deviations, 
we compute the expected- (ECE), the max- (MCE) and the overconfidence calibration error (OCE). 
More details on the computation of these measures can be found e.g. in Appendix D of \cite{ardizzone2020exact}.
The ECE measures the expected distance from the diagonal, weighted by the bin count $n(C)$ at any confidence:
\begin{equation}
    \mathrm{ECE} = \frac{1}{n_\mathrm{tot}} \sum_C n(C) | C - R(C) |
\end{equation}
But for tasks with more than $\sim 10$ classes, the ECE is almost completely dominated by the `negative' predictions: 
for any ImageNet prediction, typically only a few classes have a meaningful confidence,
while e.g. 990 of the 1000 classes will have confidences $<0.1\%$.
So the lower end of the curve is weighted $\sim 100$ times stronger than the rest of the curve,
severely shifting the ECE statistic towards the very low confidence regime.
The MCE measures the maximum distance from the diagonal:
\begin{equation}
    \mathrm{MCE} = \underset{C}{\max} \; | C - R(C) |
\end{equation}
The MCE is not affected by the same phenomenon as the ECE, but in return
is subject to random fluctuations of sparsely populated regions on the curve; it only takes a single bin into account.
Finally, the OCE measures the normalized fraction of wrong predictions that are highly confident with $C \geq C_\mathrm{crit}$, 
where we use $C_\mathrm{crit}=99.7\%$.
\begin{equation}
    \mathrm{OCE} = \frac{1}{1 - C_\mathrm{crit}} \sum_{C \geq C_\mathrm{crit}} | 1 - R(C) |
\end{equation}
For instance, an OCE of 3.5 would mean that in these high-confidence cases, the model is wrong 3.5 times more often than allowed,
the error rate should be $\leq 1-C_\mathrm{crit} = 0.3\%$ in these cases.
This measures more directly the cases we may be interested in: we want to be able to trust the decisions if they are very confident.
The OCE is less noisy than MCE in our case, as it takes more samples into account.

\begin{figure}
  \parbox{\linewidth}{
    \parbox{0.48\linewidth}{
    \centering {$\beta = 1$} 
      \includegraphics[trim=20 0 20 0,clip,width=\linewidth]{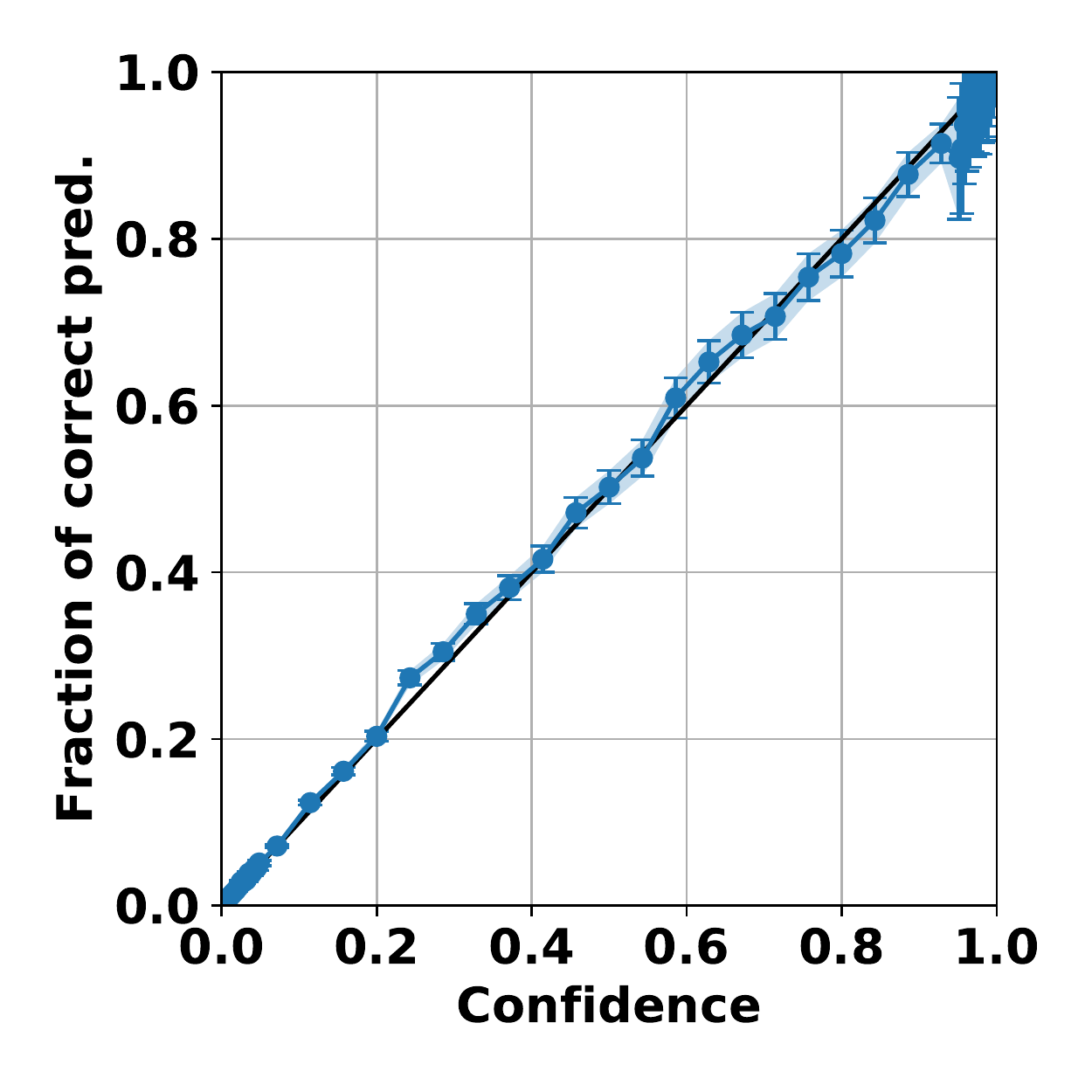} 
      \vspace{-4mm}
       }
    \hfill
    \parbox{0.48\linewidth}{
      \centering {ResNet} 
      \includegraphics[trim=20 0 20 0,clip,width=\linewidth]{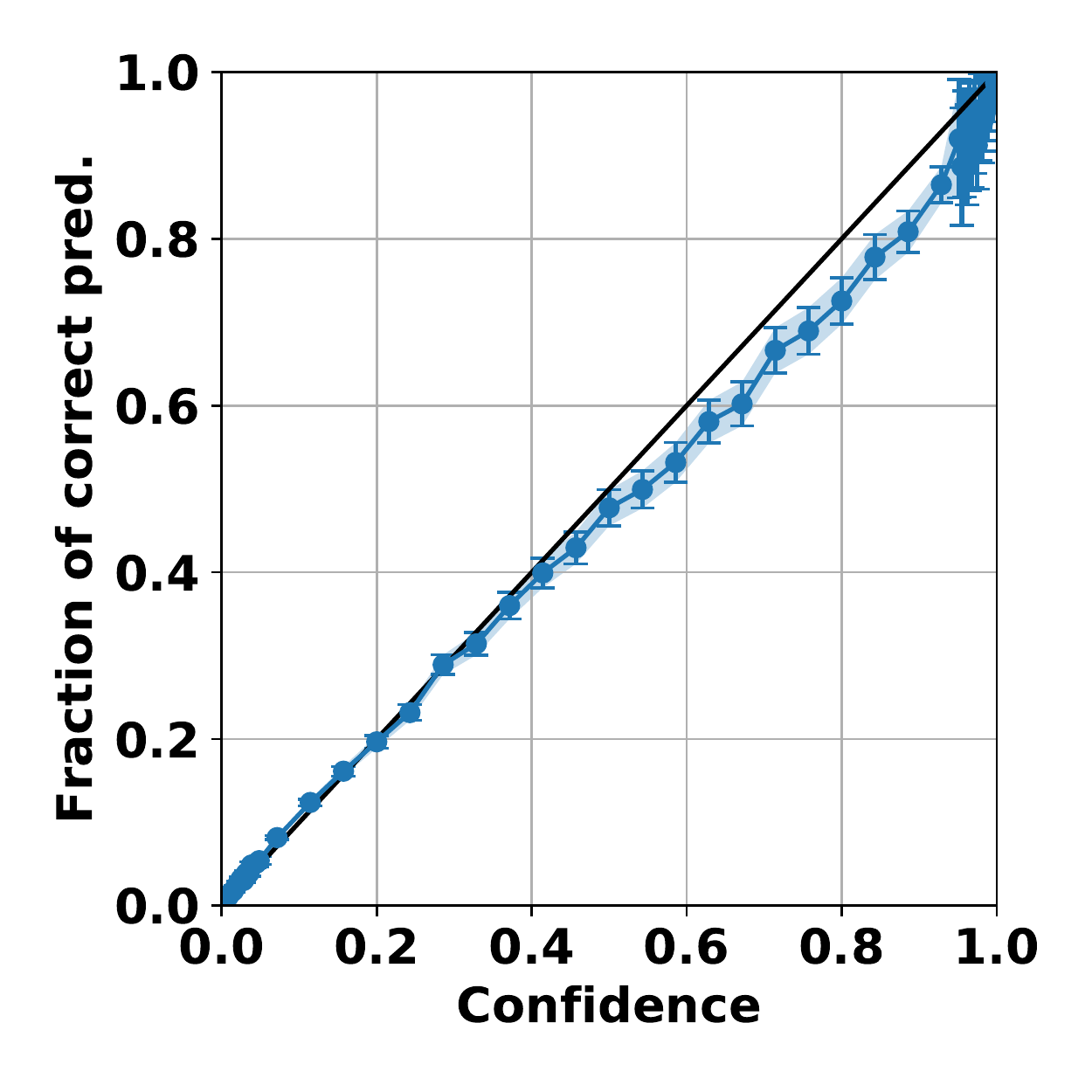} 
      \vspace{-4mm}
       }
   }
  \caption{Calibration curves for the model with $\beta=1$, and a standard ResNet-50, for reference.
  Deviations below diagonal = overconfidence, above = underconfidence. 
  The error bars are the Poisson errors computed from the bin count.
  }
  \label{fig:app_calibration_curves}
\end{figure}

We report the result in Table \ref{app:tab:calibration_error}, and show curves in Fig.~\ref{fig:app_calibration_curves}.
In short, we confirm previous observations e.g. in \cite{ardizzone2020exact}: the GC models are better calibrated than DCs.
The OCE shows the clearest trend of increasing overconfidence with $\beta$.
Even from the $\beta=\infty$ model to the standard ResNet, there is a significant jump in the calibration error, 
also seen clearly in the full calibration curves.
As the loss function for training at $\beta = \infty$ is essentially the same as a standard ResNet, this must be due to the construction of the model.
Explained further in  \ref{subsec:class_similarities} (`Class similarities'), our conjecture is that it is due to the latent space structure specifically.
 
\begin{table}
\begin{center}
\resizebox{\linewidth}{!}{

\begin{tabular}{l|c|c|c|c|c|c|c|c}
 & \multicolumn{7}{c|}{IB-INN, $\beta=$} & RN \\
 & $1.0$ & $2.0$ & $4.0$ & $8.0$ & $16.0$ & $32.0$ & $\infty$ &  \\
 \hline
ECE (\%) & $0.16$ & $0.16$ & $0.16$ & $0.17$ & $0.16$ & $0.17$ & $0.17$ & $0.17$ \\
MCE (\%) & $5.54$ & $3.13$ & $5.47$ & $4.57$ & $5.50$ & $5.28$ & $5.10$ & $7.72$ \\
OCE & $3.87$ & $4.13$ & $4.31$ & $4.73$ & $4.15$ & $4.94$ & $5.12$ & $6.75$
\end{tabular}

}
\end{center}
\caption{
Calibration Errors for different values for $\beta$ and for the ResNet. Expected Calibration Error (ECE), Max Calibration Error (MCE), Overconfidence Calibration Error (OCE) (see text for definitions).}
\label{app:tab:calibration_error}
\end{table}

\section{Explainability -- Additional Materials}
\subsection{2D Decision Space}
\label{app:explainable_2d_space}
In the following, we show another possibility to visualize the decision space for a smaller set of classes.
In our case, we select 10 labels from all ImageNet classes.
Starting from the full model,
the $\mu_y$ of the selected classes are constrained to a plane
and fine tuned, reaching 90\% accuracy for this simplified 10-class case.
This allows us to show the entire decision space in a single 2D plot.
The decision boundaries between all classes form a Voronoi tessellation of the decision space.
All latent vectors inside the Voronoi cell of a certain class 
will have the highest probability under that class.
In the case where the $\mu_y$ are not constrained to a plane and all 1000 classes are used,
the behaviour is the same, with high-dimensional polygons for each class,
but this can not be readily visualized.

\begin{figure}
    \centering
    \includegraphics[width=\linewidth]{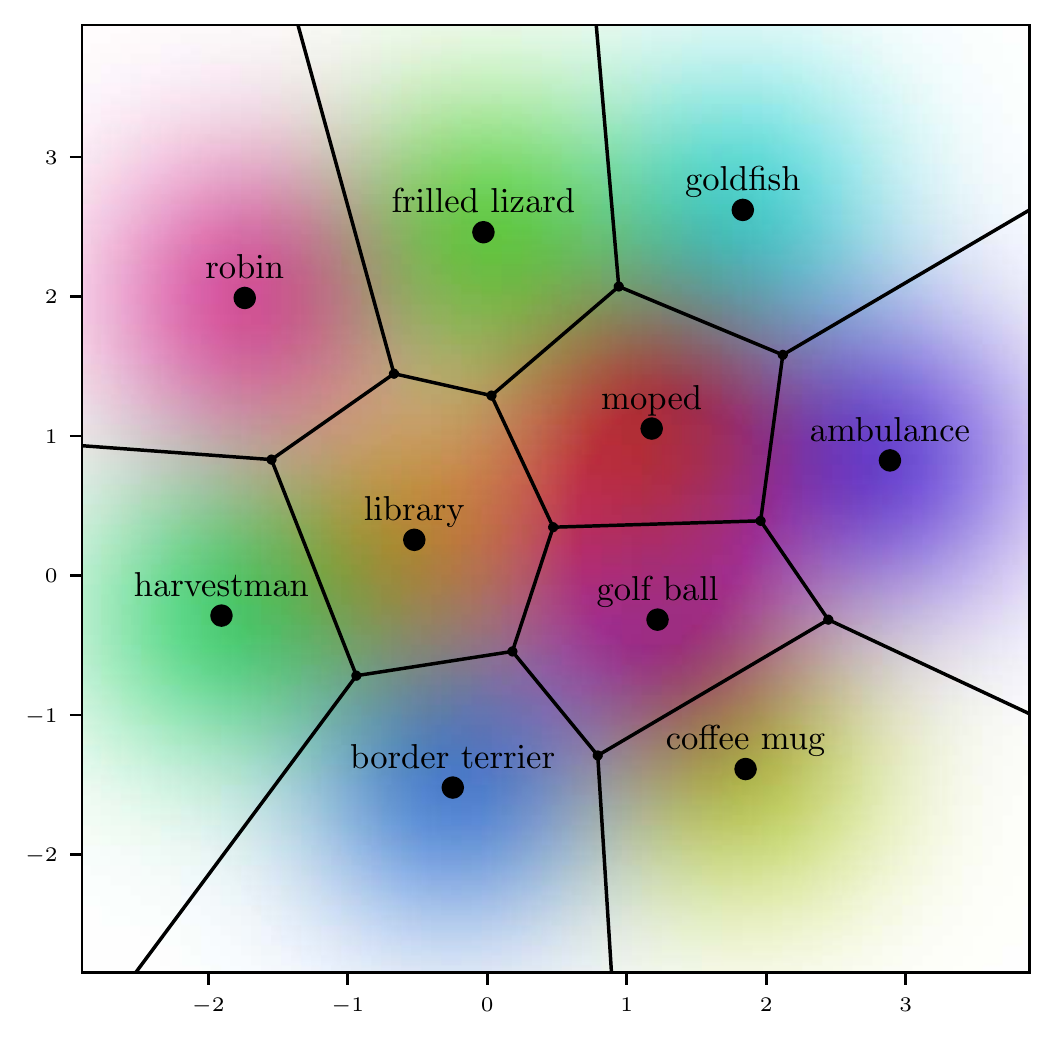}
    \caption{
    Latent space of a model with only ten classes, where the $\mu_y$ (black points) are constrained to a plane.
    The black lines are the decision boundaries, e.g. all points inside the `moped'-polygon 
    will be classified as a moped.
    The background is colored according to the probability density of each mixture component.
    }
    \label{fig:2d_latent_plane}
\end{figure}

\subsection{Class Similarity Matrix}
\label{app:explainable_similarity}
\newcommand{\dmu}{\Delta \mu}
\newcommand{\conf}{c}
\newcommand{\cconf}{C}
\newcommand{\diff}{\mathrm{d}}
For the pairwise predictive uncertainty, we only consider two classes, $y\in \{ 1, 2\}$.
We denote the distance of the class centers as $\dmu = \| \mu_1 - \mu_2 \|$.
We assume $y=1$ is the top prediction.
This is just for simplification, as 1 and 2 can be swapped in the derivation if $y=2$ is the top prediction.
The prediction confidence $c$ for any latent vector $z$ is then between $0.5$ and $1.0$, computed as
\begin{equation}
\conf(z) = \frac{ q(z \mid y\!=\!1) }{ q(z \mid y\!=\!1) + q(z \mid y\!=\!2) }
\end{equation}
The model's latent density is
\begin{equation}
q(Z) = \frac{1}{2} \mathcal{N}(\mu_1; 1) + \frac{1}{2} \mathcal{N}(\mu_2; 1)
\end{equation}
This allows us to explicitly work out how the confidences will be distributed
through the change-of-variables formula.
Note that $z$ can be expressed in cylindrical coordinates
oriented along the line connecting $\mu_1$ and $\mu_2$.
All the radial parts integrate out, only the position along this line is relevant.
After some substitutions and simplifications, we obtain
\begin{equation}
    p(\conf) = \frac{1}{A} \underbrace{
            \left( \conf - \conf^2 \right)^{- 3/2} 
            \exp \left( - \frac{1}{2 \dmu^2} \log^2 \left( \frac{1}{\conf} - 1 \right) \right)
        }_{\coloneqq \rho(\conf)}
\end{equation}
$A$ is the normalization constant and has no closed form:
\begin{equation}
    A = \int_{1/2}^{1} \rho(\conf) \diff \conf
\end{equation}
And we simply call the unnormalized density $\rho(\conf)$.
Finally, the expected confidence $\overline{\cconf}$ can be readily computed as
\begin{equation}
    \overline{\cconf} = \frac{\int \conf \rho(\conf) \diff \conf}{\int \rho(\conf) \diff \conf}
\end{equation}
The expected \emph{uncertainty} as opposed to the confidence is simply $1 - \overline{\cconf}$.

\begin{figure}
    \includegraphics[width=\linewidth]{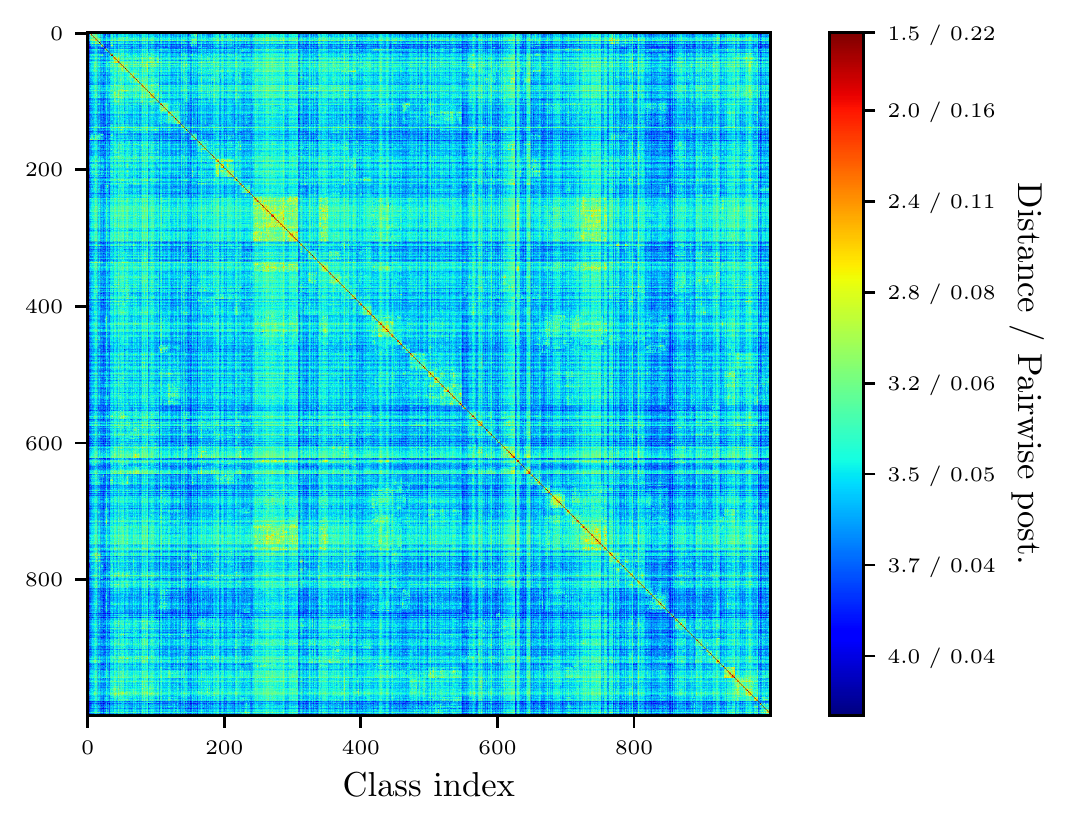}
    
    \caption{Similarity matrix between all 1000 classes. The two large clusters around class index 250 and 750 are dogs.
    The colormap indicates the pairwise distance of the $\mu_y$ as well as the expected pairwise posterior,
    meaning e.g. the binary decision between a tabby cat and a tiger cat is associated with $20\%$ expected uncertainty, by construction (see text).}
    \label{app:fig:similarity_matrix}
\end{figure}

\subsection{Saliency Heatmaps}
\label{app:explainable_salience}
As outlined in the paper, to derive the saliency and posterior heatmaps, we start with the following definition:
\begin{align}
 w^{(y)} & = \mathrm{DCT}^{-1}(z) - \mathrm{DCT}^{-1}(\mu_y) \nonumber \\
         & = \mathrm{DCT}^{-1}\big(z - \mu_y\big).
\end{align}
Because the DCT operation is linear and orthogonal, it conserves distances,
and we can write
\begin{equation}
 q(z|y) \propto \exp\left( - \frac{1}{2} \| z - \mu_y \|^2 \right) = \exp\left( - \frac{1}{2} \| w^{(y)} \| ^2 \right)
\end{equation}
We can consider the spatial structure present in $w^{(y)}$: 
It will have three indices, $k,l$ for the spatial position and $m$ for the feature channels: $w^{(y)}_{klm}$.
We can simply factorize over the spatial dimensions. For the $\log$-probability, we get
\begin{align}
   \log q(z|y) &= \sum_{k,l} -\frac{1}{2} \| w^{(y)}_{kl,:} \|^2  + const.\\
   & \coloneqq \sum \log q(w_{kl}|y)
\end{align}
We will ignore the $const = - {\mathrm{dim}(z) \log(2\pi)}/{2}$ for convenience.

This spatial decomposition of $\log p(z|y)$ allows us to make various heatmap visualziations in a principled way.

First, we consider $-\log p(w_{kl}|y)$. Considering the sum over pixels,
this looks like a point-wise entropy. 
The common interpretation from information theory is, that this is a measure how much information is contained 
in each part of the image.
The values in each pixel sum to $-\log p(z|y)$, which is then the overall entropy of the latent vector for this image.
To remove the class dependence, we plot the `saliency heatmap':
\begin{equation}
Q_\mathrm{Saliency}(k,l) = - \log \left( \sum_y q(w_{kl}|y) p(y) \right)
\end{equation}
Some examples for this are shown in Fig.~\ref{fig:app_class_heat_maps}.

\subsection{Posterior Heatmaps}
\label{app:explainable_class_heatmaps}

\begin{figure*}

\parbox{0.48\textwidth}{
    \includegraphics[width=\linewidth]{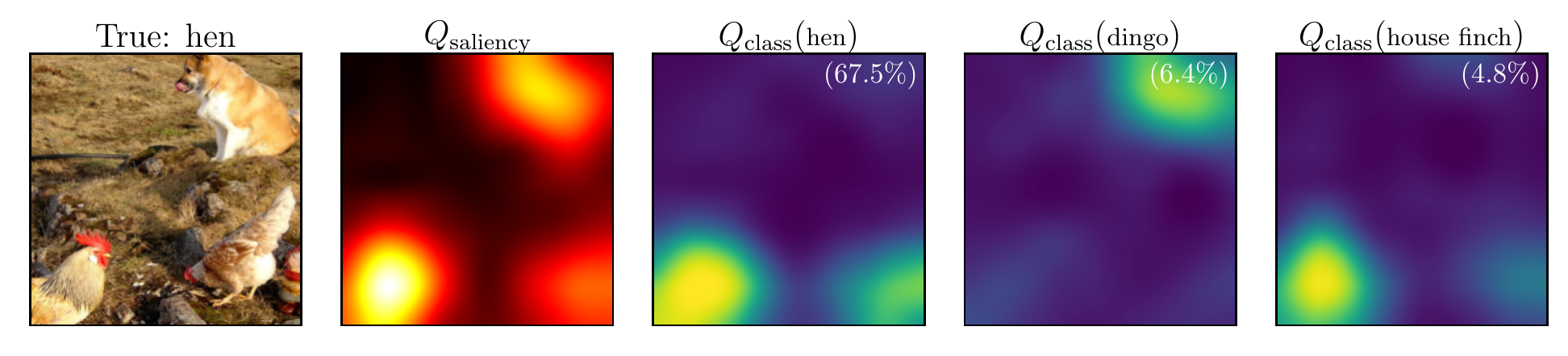}
    \includegraphics[width=\linewidth]{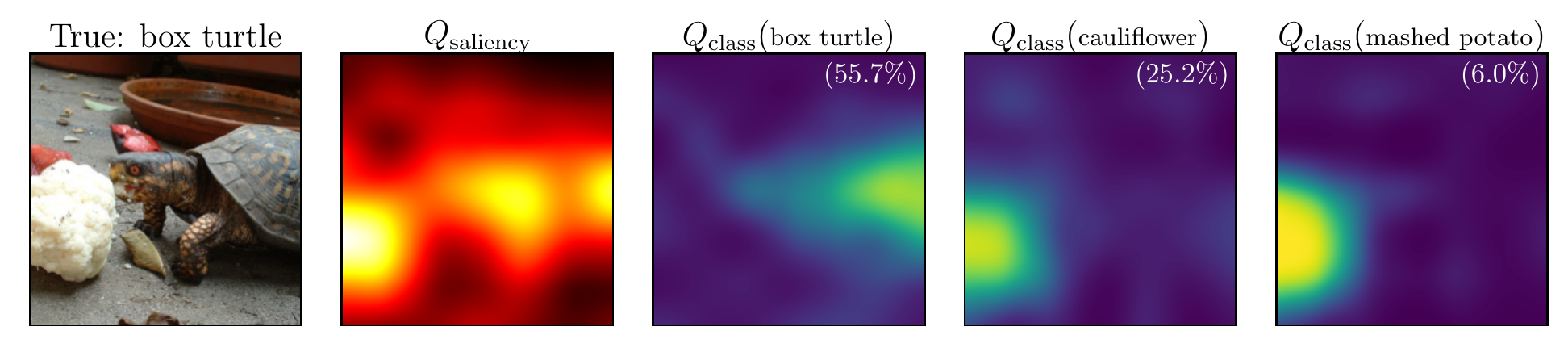}
    \includegraphics[width=\linewidth]{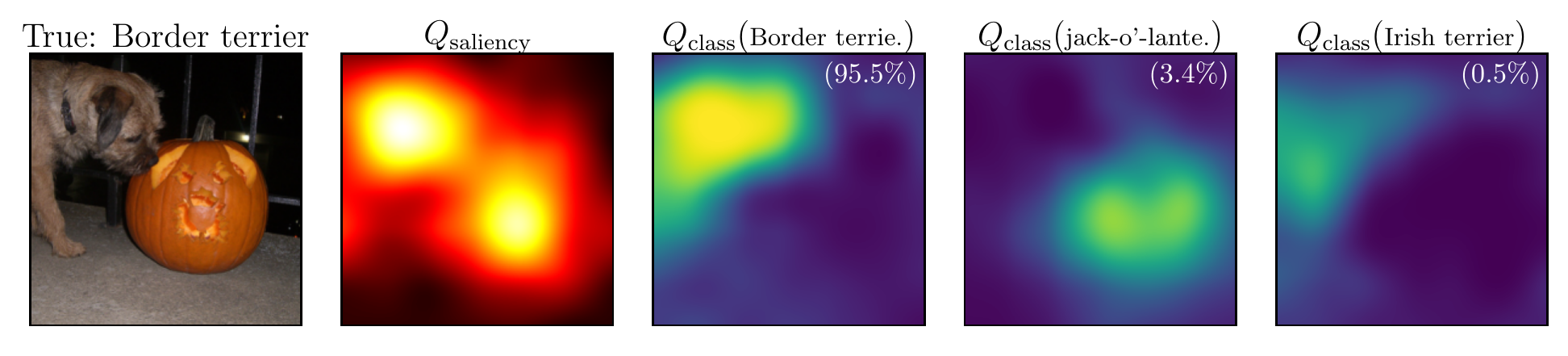}
    \includegraphics[width=\linewidth]{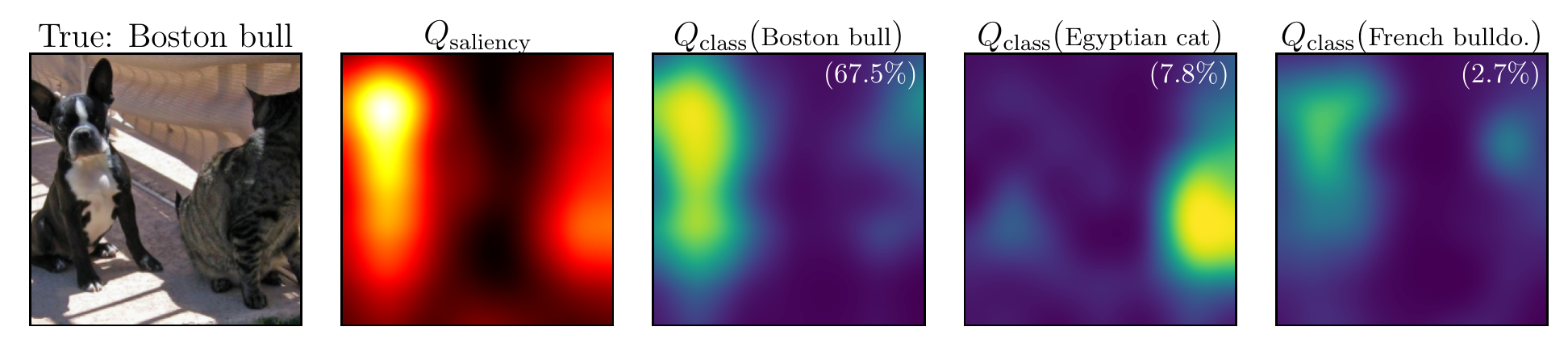}
    \includegraphics[width=\linewidth]{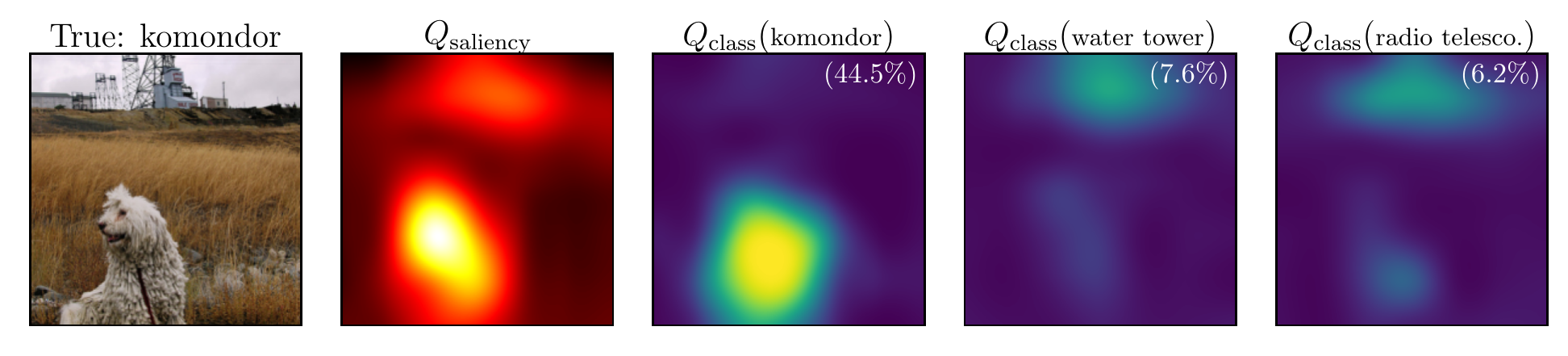}
}
\hfill
\parbox{0.48\textwidth}{
    \includegraphics[width=\linewidth]{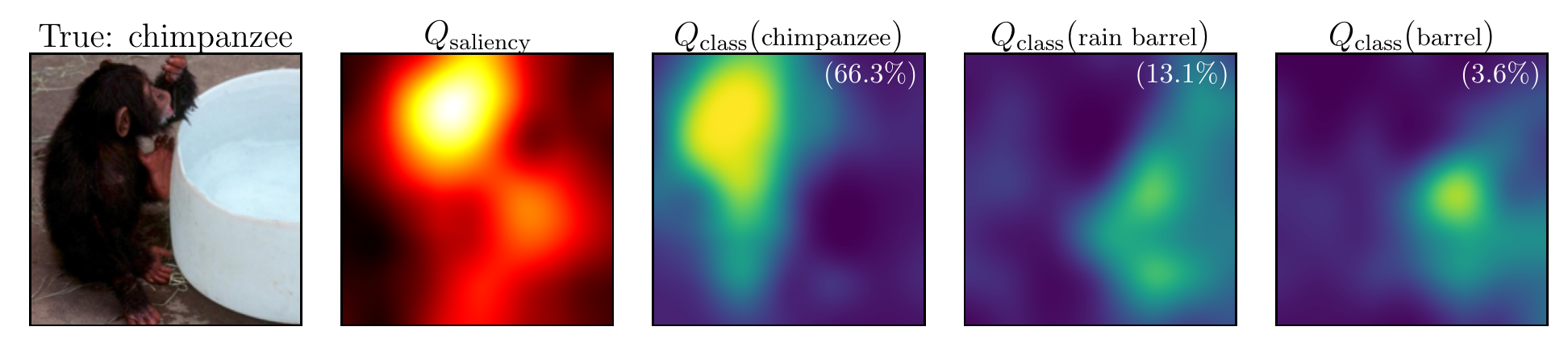}
    \includegraphics[width=\linewidth]{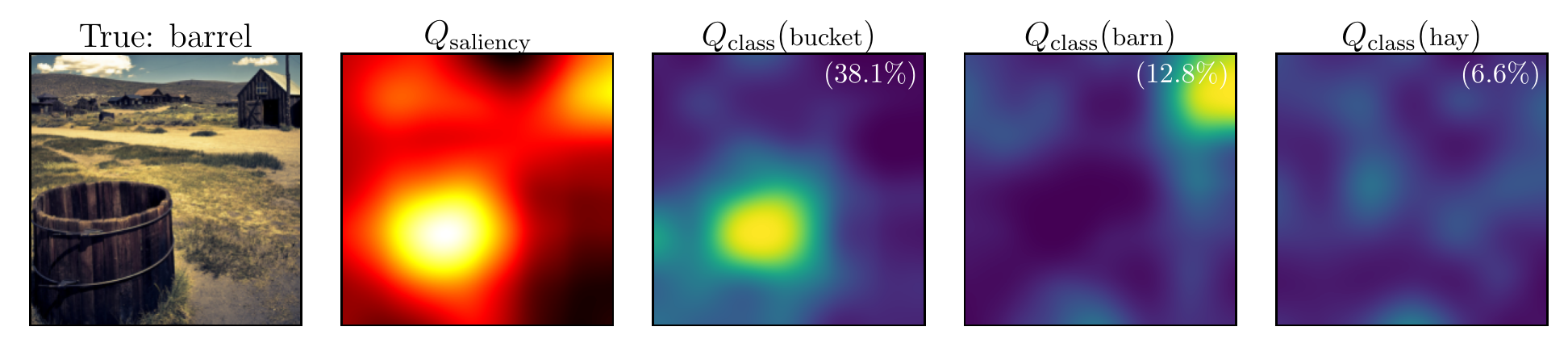}
    \includegraphics[width=\linewidth]{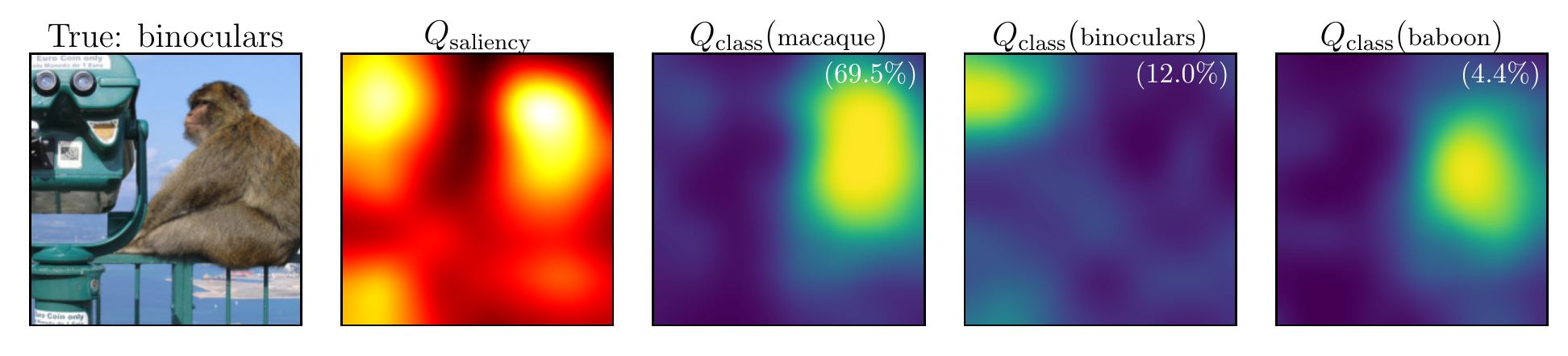}
    \includegraphics[width=\linewidth]{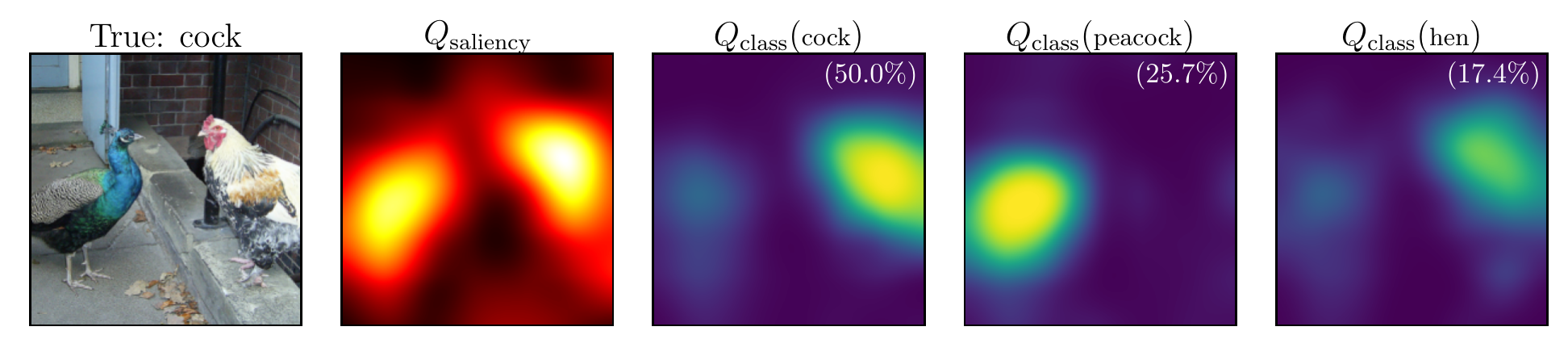}
    \includegraphics[width=\linewidth]{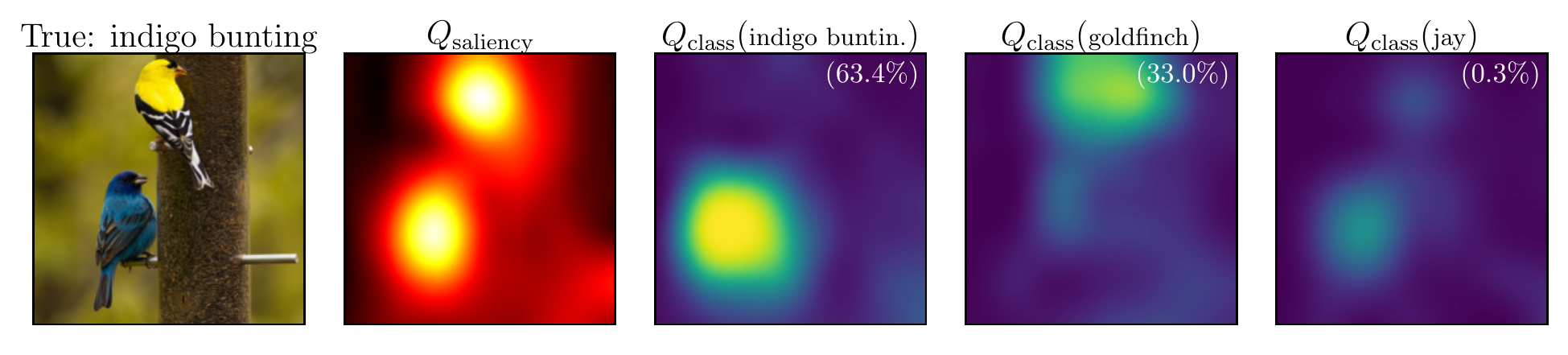}
}
\caption{Additional examples for saliency maps and posterior heatmaps for the top three classes.
The white inset numbers indicate the confidence in that class, which is equal to the exponential of the sum over the posterior heatmap (see text).
}
\label{fig:app_class_heat_maps}
\end{figure*}

We can now consider the class prediction:
\begin{equation}
q(y|x) = \frac{q(z|y)}{\sum_{y'} q(z|y')} \eqqcolon \frac{q(z|z)}{S(z)},
\label{eq:heatmap_true_post}
\end{equation}
where $p(y) = 1/M$ and the Jacobian $|\det J|$ both cancel out.
We therefore plot for any class the following `class posterior heatmap':
\begin{equation}
    Q_\mathrm{Class}(k,l,y) = \log p(w_{kl} | y) - S_{kl} \quad \text{s.t.} \quad \sum_{kl} S_{kl} = S
\end{equation}
The $-S_{kl}$ term means a fixed `image' is subtracted from each heatmap, representing the denominator, which is constant for all classes.
There is some freedom to choose $S_{kl}$, as long as it sums to $S$.
When distributing it evenly over space, the differences in the heatmaps between classes are hard to see by eye, 
compared to the common differences within the heatmaps shared across classes, which are larger by magnitude.
Heuristically, we instead find the best contrast when we choose the relative weight of each $S_{kl}$ in the following way:
\begin{align}
S_{kl} &= S \;\frac{r_{kl} + 0.03}{\sum_{kl}(r_{kl} + 0.03)} 
\end{align}
where $r_{kl}$ is the same as $\log p(w_{kl})$ but normalized to the $[0,1]$-range over each image.
Additional examples are shown in Fig.~\ref{fig:app_class_heat_maps}.

Comparing to Eq.~\ref{eq:heatmap_true_post}, we see that summing $Q_\mathrm{Class}$ over feature-space pixels gives exactly
the log-prediction $\log \qt(y|x)$. 
So $Q_\mathrm{Class}$  represents a spatial decomposition of the actual predictive output:
\begin{equation}
q(y|x) = \exp\left( \sum_{kl} Q_\mathrm{Class}(k,l,y) \right)
\end{equation}


\section{Robustness -- Additional Materials}

\subsection{Corrupted Images Examples}
\label{app:robustness_corrupted_images}
\begin{figure*}
\centering
\parbox{0.495\textwidth}{
\centering
\includegraphics[width=0.85\linewidth]{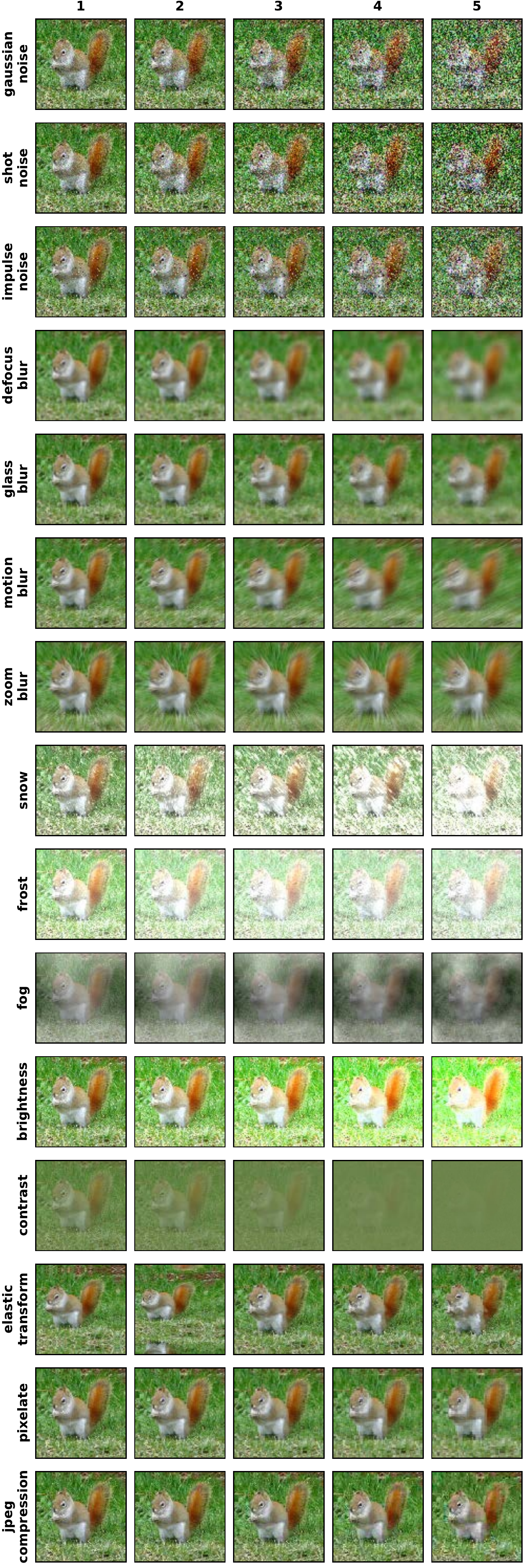}
}
\hfill
\parbox{0.495\textwidth}{
\centering
\includegraphics[width=0.85\linewidth]{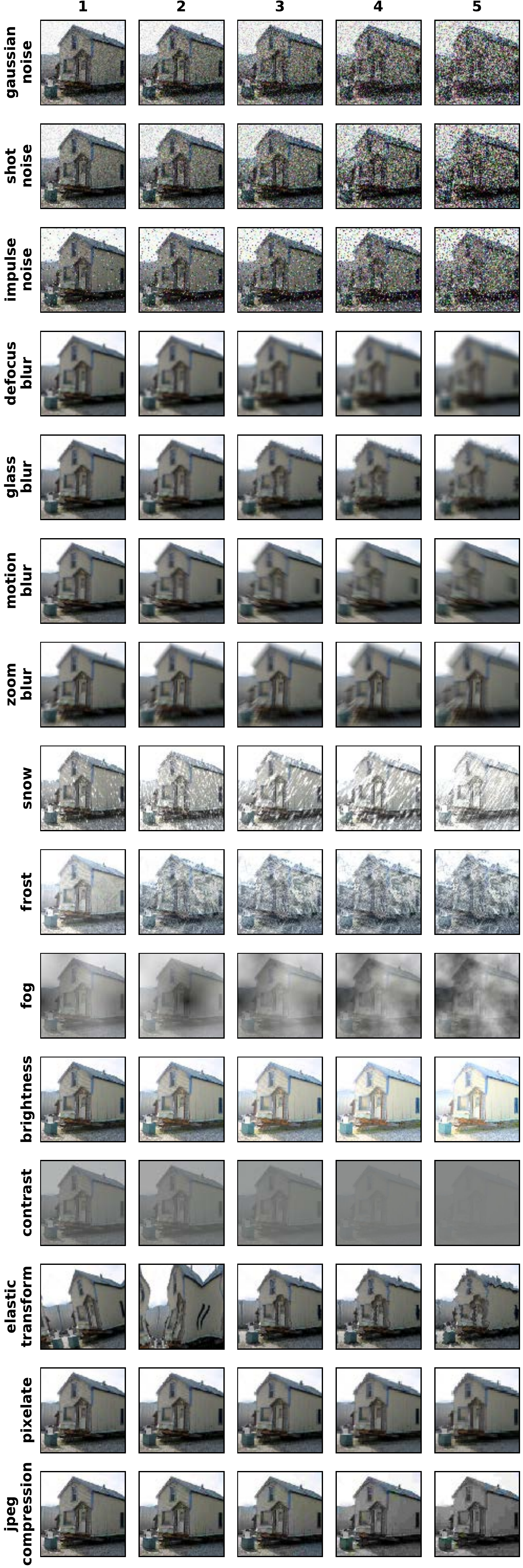}
}
\caption{We show the different corruption types (rows) with their severity levels from 1-5 (columns) applied to two sample images (Samples belong to the ImageNet classes 'fox squirrel' and 'mobile home').
}
\label{fig:app_corruptions_examples}
\end{figure*}

Examples of the different corruptions and the severity levels are shown in Fig.~\ref{fig:app_corruptions_examples}.

\subsection{Adversarial Attack Objectives}
\label{app:robustness_adversarial_objectives}
As explained in the paper, we performed the well established `Carlini-Wagner' white-box targeted attack method introduced in \cite{carlini2017towards}. 
Here, the attacked image is parametrized as $\xatt = \frac{1}{2}(\mathrm{tanh}(w) + 1)$, to ensure the image values are between 0 and 1.
The attack then consists of optimizing $w$ directly to minimize the following objective:
\begin{align}
     \mathcal{L}_{\mathrm{CW}}(w, x) = & \left\| \xatt(w) - x \right\|^2 \nonumber \\
     & + c  \max \Big( \max (\{ l_y : y \neq t \}) - l_t, -\kappa \Big)
    \label{eq:robustness:carlini}
\end{align} 

The original image is $x$, and the logits output by the model for each class $y$ are $l_y$.
The target class, that the attacked image is supposed to be classified as, is $t \coloneqq y_\mathrm{target}$.
The logits are recomputed by the model on each iteration using the updated $\xatt(w)$,
which they depend on: $l_y = l_y(\xatt(w))$.
The gradients are propagated through the model.
We call the max-term $\mathcal{L}^{(\kappa)}_\mathrm{class}(y_\mathrm{target})$ in the paper.

In other words, the attack objective simultaneously attempts to make $\xatt$ and $x$ the same, 
and to maximize the difference between the logit of the target class, and the currently next highest predicted class.
Once the distance is larger than the hyperparameter $\kappa$ in favour of the target class, this loss term does not contribute anymore.
Adjusting $\kappa$ therefore has a direct influence on the confidence of the (wrong) predictive posterior. 
From an attackers point of view it is optimal to fool a classifier to make certain but wrong predictions by setting a high value for $\kappa$, while finding a $w$ so that $\xatt$ is as close as possible to the original image $x$. 
Ideally the differences between $\xatt$ and $x$ remain imperceptible to the human eye. 
From the victim networks point of view, the targeted wrong prediction should be as uncertain as possible, and the difference between $\xatt$ and $x$ as large as possible.

For GCs, there are not logits per se. Instead, we use the conditional log-likelihoods $l_y = \log p(x|y)$, to get the same behaviour.
We performed all adversarial attacks on the same randomly chosen 200 test images, paired with the fixed random target class each.
To perform the attack, we use the Adam optimizer with its initial learning rate set to $0.01$, as in \cite{carlini2017towards}. 
We performed the attacks with three different values for $\kappa$: $0.01$, $1.0$, $\infty$. 
The parameter $c$ was fixed and set to $10$,
which is the lowest possible value for achieving a 100\% attack success rate on all our tested models.
We assume the attack converged  
whenever $\mathcal{L}_{CW}$ stops improving for 20 consecutive gradient steps.

As illustrated previously in \cite{carlini2017adversarial}, any adversarial attack defense- or detection mechanism can itself become target of a modified attack,
fooling the classification and the detection at the same time.

In line with this work, we construct a modified attack loss to achieve fooling the two-tailed quantile test we utilized for detecting attacks.
As stated in the main part of this work we denote it as $\mathcal{L}_{CWD}(w, x)$:
\begin{align}
\mathcal{L}_\mathrm{CWD}(w, x) & = \mathcal{L}_\mathrm{CW}(w, x) \nonumber \\
& + \! d \! \cdot \! \underbrace{\Big(\underset{x' \sim X_\mathrm{train}}{\mathrm{median}}\big(\log q(x')\big) - \log q(\xatt(w))\Big)^2}_{\mathcal{L}_\mathrm{detect}} 
\label{eq:robustness:carlini-detection}
\end{align}
In the added $\mathcal{L}_\mathrm{detect}$ term, $\underset{x' \sim X_\mathrm{train}}{\mathrm{median}}\big(\log q(x')\big)$ stands for the median estimated probability density (PD) of the training set and $\log q(\xatt(w))$ for the estimated PD of the perturbed image. 
Intuitively, we are now forcing $\xatt$ to move to the center of the distribution of PD values of the training data.
If it reaches the median exactly, the ROC-AUC detection score will be $0\%$ (also see Sec.~\ref{app:ood_detectionn_method}).

\subsection{Adversarial Trajectories}
\label{app:robustness_trajectories}

\begin{figure*}
    \parbox{0.32\linewidth}{
        \centering
        \includegraphics[width=\linewidth]{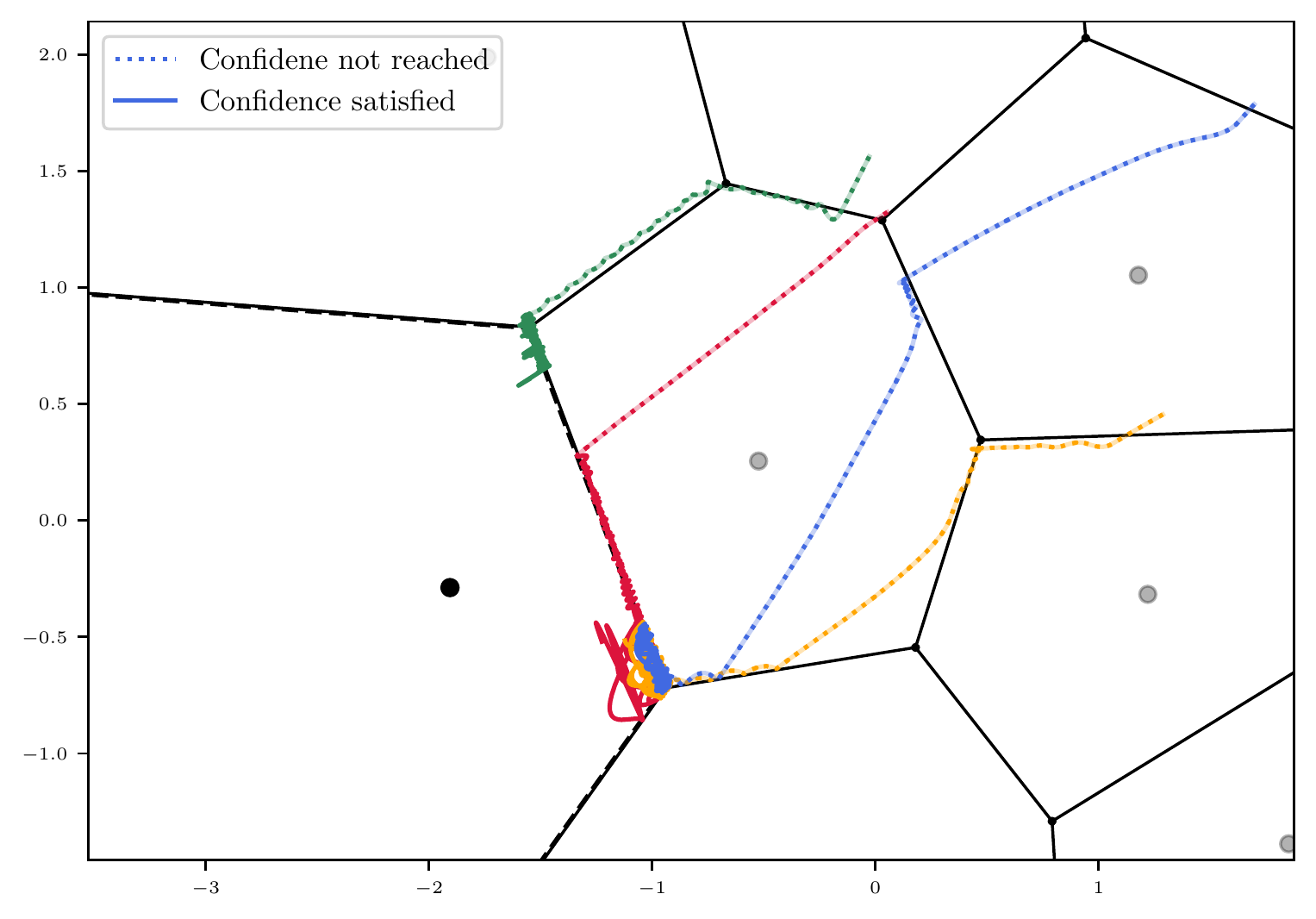}
        $\kappa = 0.01$
    } %
    \parbox{0.32\linewidth}{
        \centering
        \includegraphics[width=\linewidth]{figures/adversarial_trajectories/adversarial_trajectory_class_3_2d_kappa_1.0.pdf}
        $\kappa = 1.0$
    } %
    \parbox{0.32\linewidth}{
        \centering
        \includegraphics[width=\linewidth]{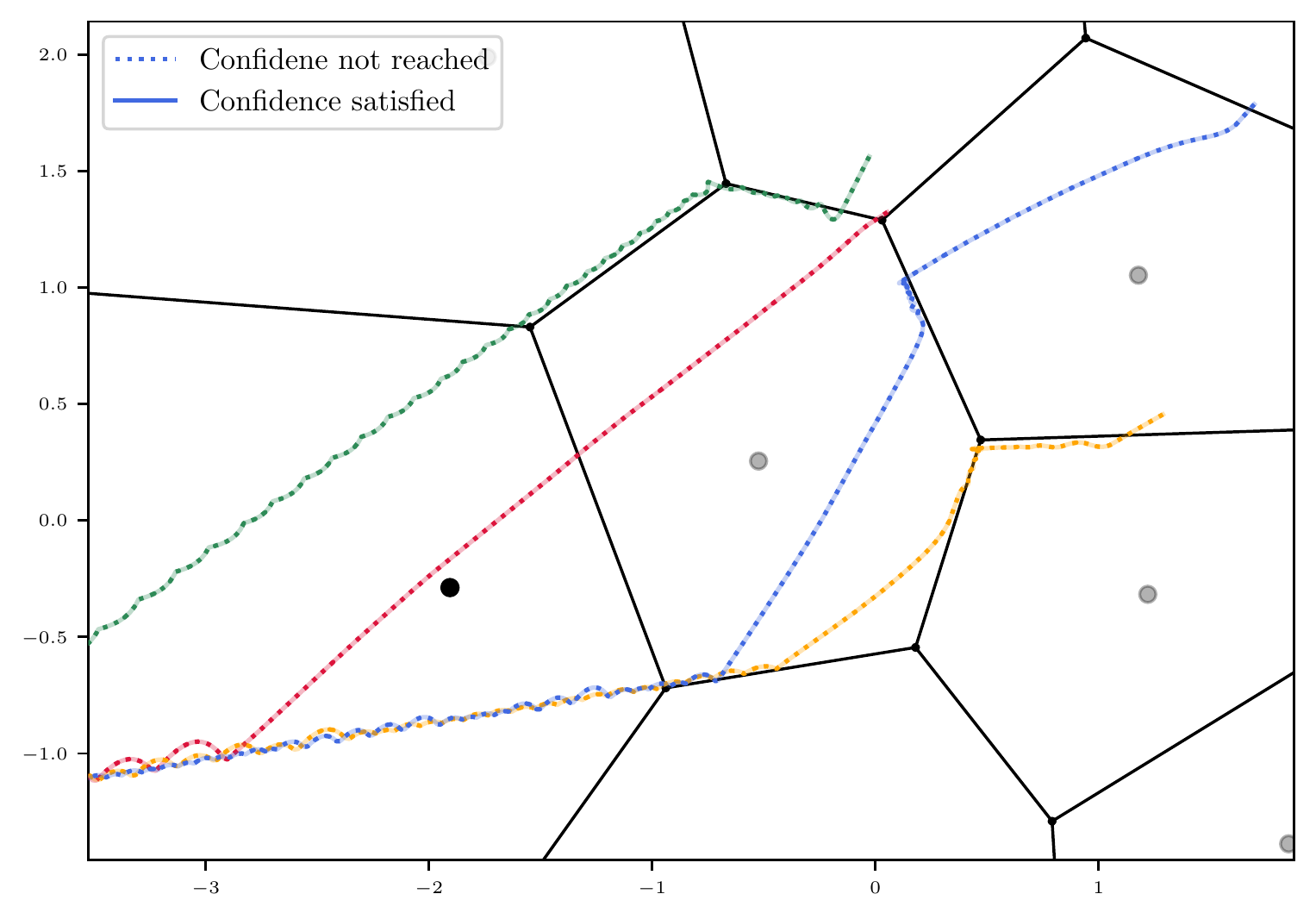}
        $\kappa = \infty$
    }
    \caption{Additional examples for Fig.~\ref{fig:adversarial_trajectory} with different settings for $\kappa$.}
    \label{fig:app_extra_trajectories}
\end{figure*}

\begin{figure}
    \centering
    \includegraphics[width=\linewidth]{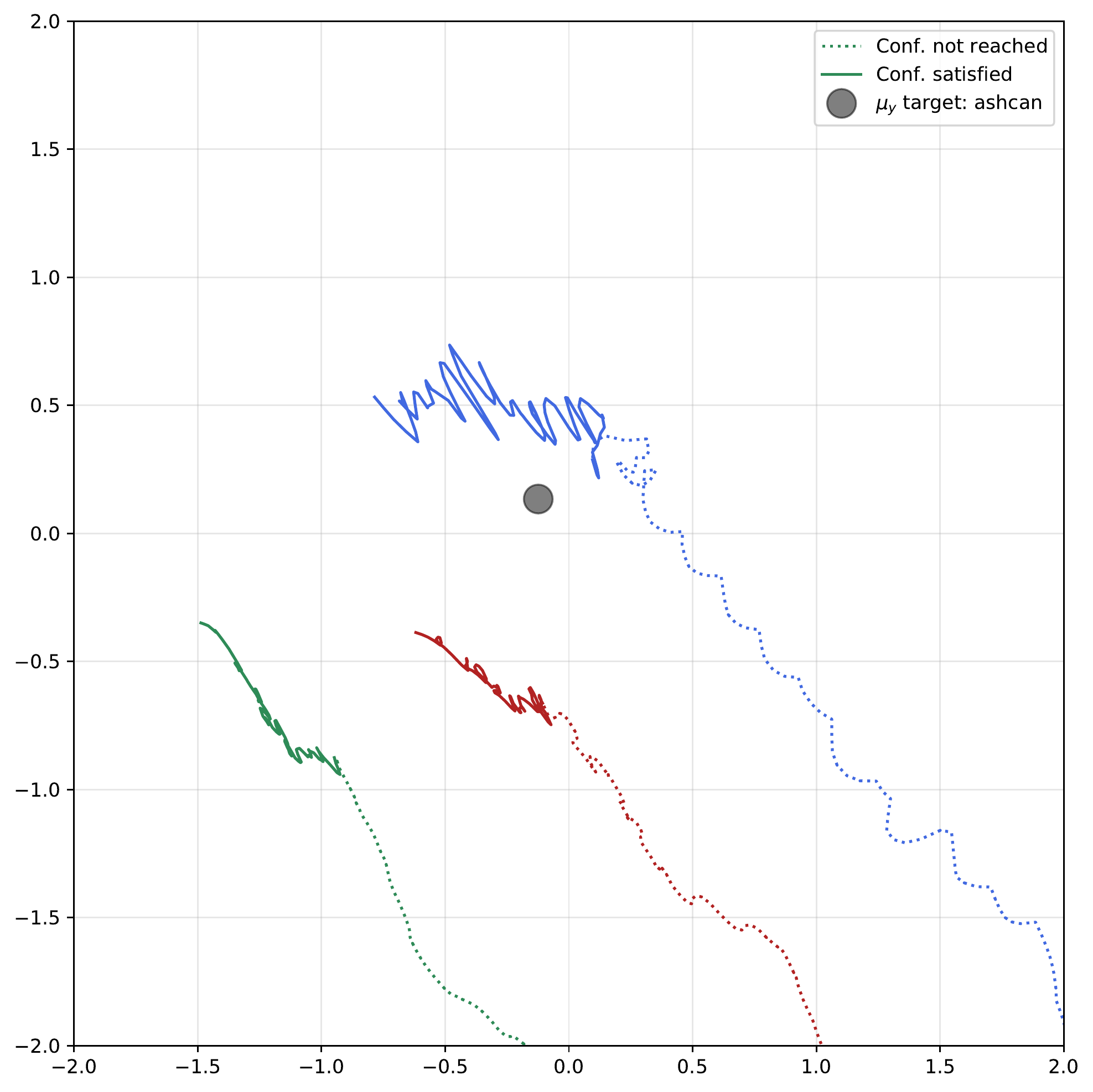}
    \caption{Visualization of the adversarial trajectories for the full model, $\kappa=1$. 
    The trajectory is projected to 2D by fitting a plane through the five classes that the trajectory passes closest to.}
    \label{fig:app_1000_class_trajectory}
\end{figure}

We find that the attack consists of two distinct stages. 
First, the attack attempts to cross into the area belonging to the target class, leaving a certain margin specified implicitly by $\kappa$.
Second, the attack minimizes the magnitude of the adversarial perturbation, while staying inside this region (sometimes stepping outside the region for a single iteration). 
We can visualize the attack's trajectory and its effect on the decicison explicitly, using the 2D decision model from Sec.~\ref{app:explainable_2d_space},
see  Fig.~\ref{fig:app_extra_trajectories}.
We also perform the same visualization for the full model with 1000 classes, shown in Fig.~\ref{fig:app_1000_class_trajectory}.
We observe the same behaviour, although the decision boundaries can no longer be visualized.
For the 2D figure, we consciously chose a target class located at the `edge' of the latent space, not circled by other classes on all sides.
This is because for the 1000 class case in higher dimensions, all classes are essentially guaranteed to be such `edge' classes.

An important lesson to take from this is that the area of maximal confidence of the attack is not necessarily closest to $\mu_\mathrm{target}$.
Instead, the confidence depends
on the \emph{difference} of the squared distance to the other classes (see Eq.~\ref{eq:robustness:carlini}). 
Especially for high $\kappa$, sufficient confidence is only achieved far outside of the original distribution,
which is what leads to the almost perfect detection score for $\kappa=\infty$ reported in Sec.~\ref{app:robustness_adversarial_results}, Table~\ref{tbl:robustness:kappa-inf} under the OoD column. This result is also also visually illustrated in Fig.~\ref{fig:adversarial_bar_graph} (second column, third row) in the main paper.

\subsection{Adversarial Attacks -- Full Results}
\label{app:robustness_adversarial_results}
\begin{figure*}
     \centering
     \includegraphics[width=\linewidth]{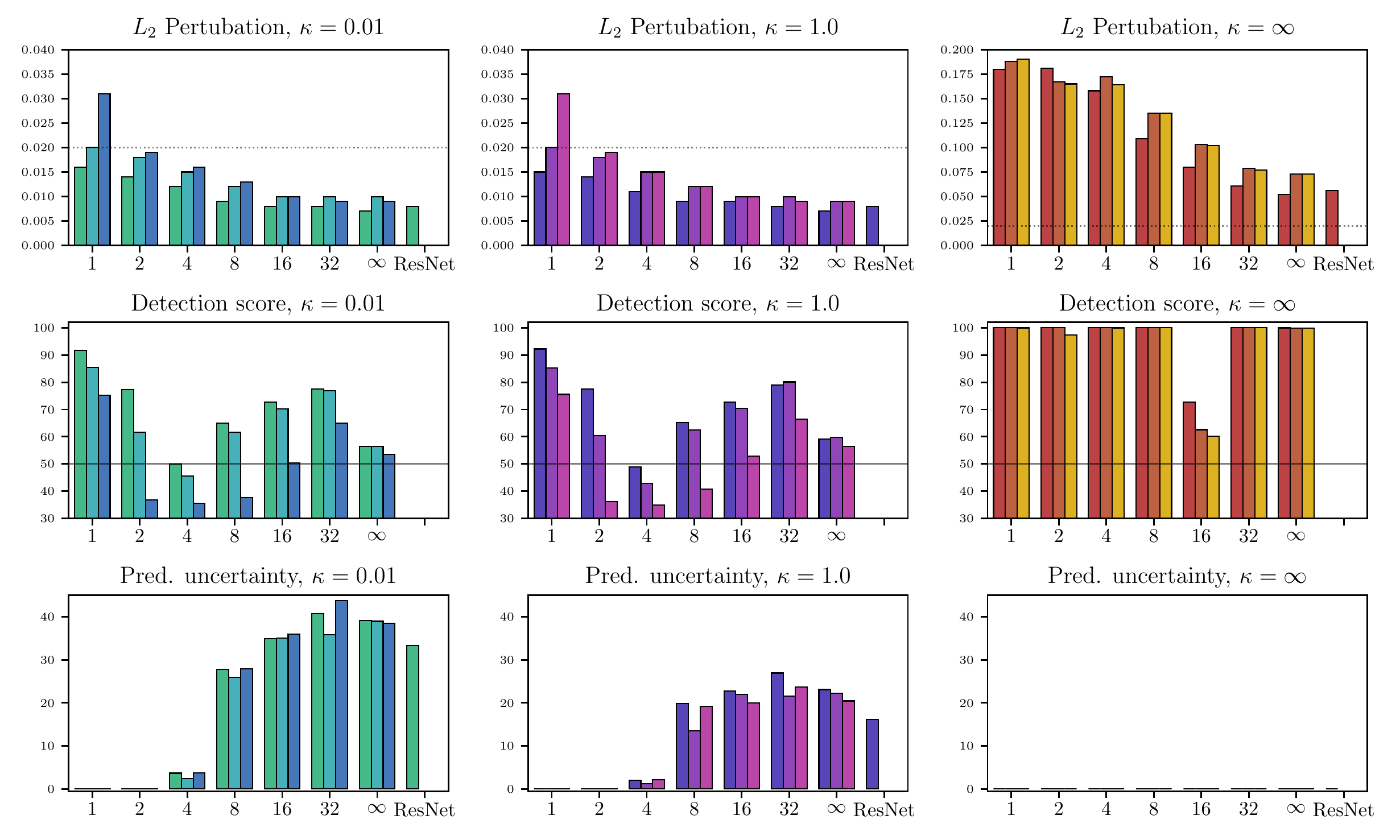}
     \caption{
         Behaviour of GCs under adversarial attacks.
     The three rows of plots give the mean perturbation, detection score, and uncertainty of the wrong prediction ($1 - \mathrm{confidence}$).
     The three columns of plots correspond to adversarial attacks with $\kappa = 0.01$ (targeted prediction with any confidence is enough),
     $\kappa = 1$ (targeted prediction should have high confidence), and $\kappa = \infty$ (targeted prediction should be as confident as possible).
     The three bars for each $\beta$ correspond to: standard adversarial attack ($d=0$), as well as $d=66$ and $d=1000$, 
     i.e. the detection mechanism is fooled at the same time as the prediction.
     The dotted line in the top row roughly indicates the level at which attacks are clearly visible by eye.
     Note that this is subjective and only a rough indication.
     The line in the second row indicates random performance, i.e. the OoD detection does nothing useful. 
     }
     \label{fig:app_adversarial_bar_graph}
\end{figure*}

All results concerning adversarial attacks are summarized in Fig.~\ref{fig:app_adversarial_bar_graph}, and Table \ref{tbl:robustness:kappa-inf}, corresponding to Fig.~\ref{fig:adversarial_bar_graph} in the main paper.

\begin{figure*}[t!]
    \parbox{0.48\textwidth}{
        \centering
        \includegraphics[trim=40 20 0 50,clip, width=0.9\linewidth]{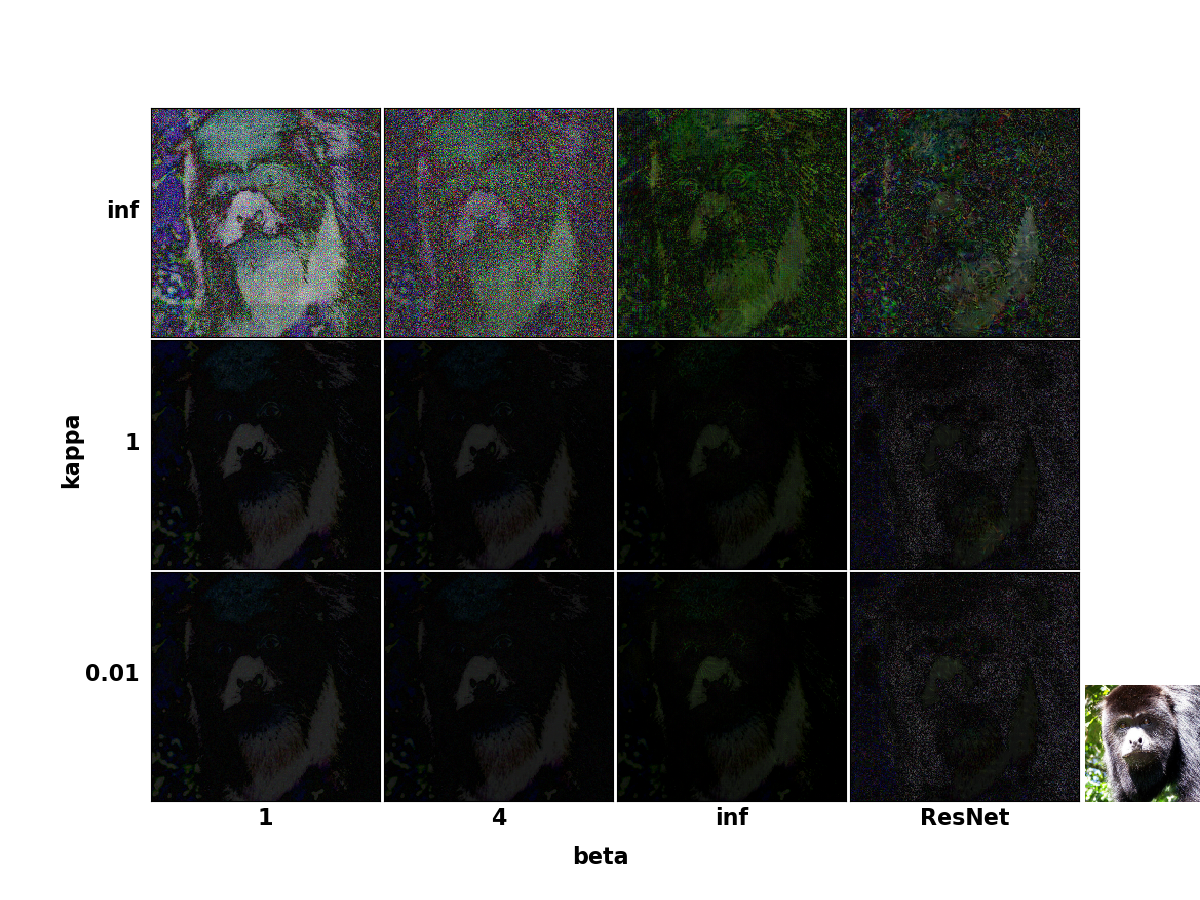}
        \captionof{figure}{Qualitative results demonstrating the influence of $\kappa$ (controlling the classifiers final confidence on targeted classes) and $\beta$ (controlling the generative modeling capability of the classifier) on adversarial attack robustness. The discriminative classifier ResNet is added for reference. 
        The figure, showing the per-pixel errors in RGB space, gives the absolute difference between the original (bottom right corner) and the adversarially perturbed image, amplified by a scaling factor for visibility. 
        For adversarial attacks to achieve highly confident posteriors (high value for $\kappa$) the noise has to be amplified. In order to successfully trick a classifier with better generative modeling capabilities (low value for beta) the noise added by the attack has to be even larger. }
        \label{fig:robustness:attack_kappa}
    }
    \hfill
    \parbox{0.48\textwidth}{
        \centering
        \includegraphics[trim=0 20 0 50,clip, width=0.9\linewidth]{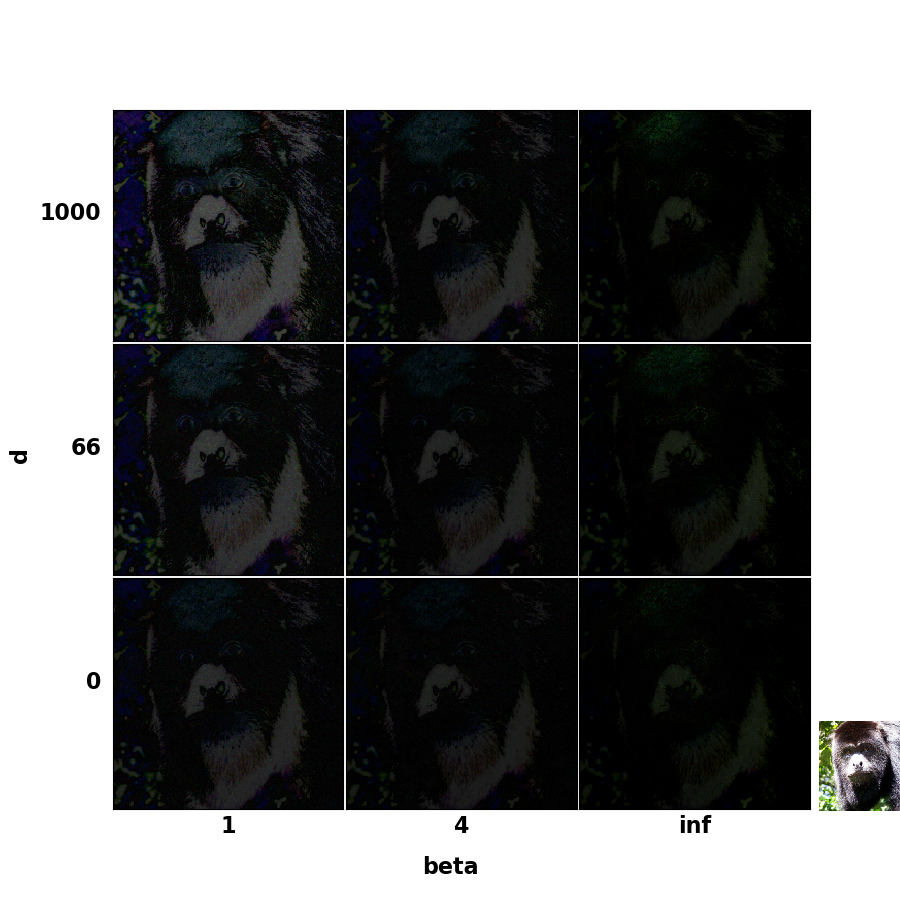}
        \captionof{figure}{Qualitative results demonstrating the influence of $d$ (controlling the strength put on fooling the attack detection mechanism) and $\beta$ on adversarial attack robustness for $\kappa$ fixed to $1$. The more weight is put on fooling the attack detection mechanism (higher values for $d$), the more noise must be added to the input image by the adversarial attack. In order to fool the generally stronger detection mechanism of classifiers with higher generative modeling capabilities, the noise must be even higher.}
        \label{fig:robustness:attack_d}
    }
\end{figure*}

\begin{figure*}[!htb]
\begin{tikzpicture}[overlay]
     \node at (18.3,-12) {\includegraphics[width=2cm]{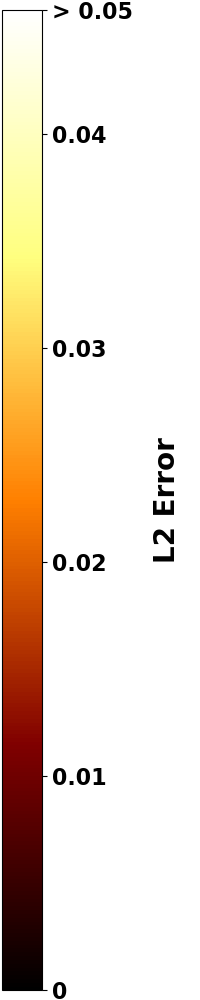}};
\end{tikzpicture}
\input{figures/validation/attacks_kappa-0.01}
\vspace{-2mm}
\caption{$\kappa=0.01$, $d=0$}
\vspace{1mm}
\label{fig:robustness:attacks_kappa-0.01}
\input{figures/validation/attacks_kappa-0.01_d-66}
\vspace{-2mm}
\caption{$\kappa=0.01$, $d=66$}
\vspace{1mm}
\label{fig:robustness:attacks_kappa-0.01_d-66}
\input{figures/validation/attacks_kappa-0.01_d-1000}
\vspace{-2mm}
\caption{$\kappa=0.01$, $d=1000$}
\label{fig:robustness:attacks_kappa-0.01_d-1000}
\end{figure*}

\begin{figure*}
\begin{tikzpicture}[overlay]
     \node at (18.3,-12) {\includegraphics[width=2cm]{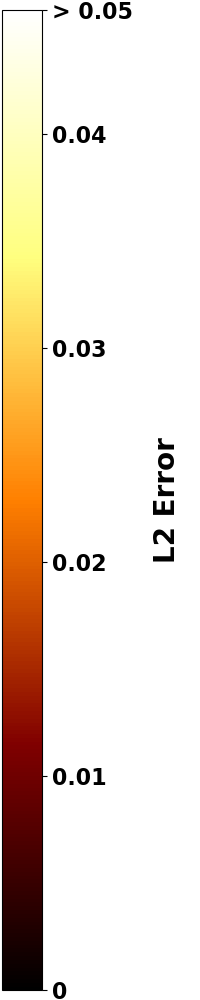}};
\end{tikzpicture}
\input{figures/validation/attacks_kappa-1}
\vspace{-2mm}
\caption{$\kappa=1$, $d=0$}
\vspace{1mm}
\label{fig:robustness:attacks_kappa-1}
\input{figures/validation/attacks_kappa-1_d-66}
\vspace{-2mm}
\caption{$\kappa=1$, $d=66$}
\vspace{1mm}
\label{fig:robustness:attacks_kappa-1_d-66}
\input{figures/validation/attacks_kappa-1_d-1000}
\vspace{-2mm}
\caption{$\kappa=1$, $d=1000$}
\label{fig:robustness:attacks_kappa-1_d-1000}
\end{figure*}

\begin{figure*}
\begin{tikzpicture}[overlay]
     \node at (18.3,-12) {\includegraphics[width=2cm]{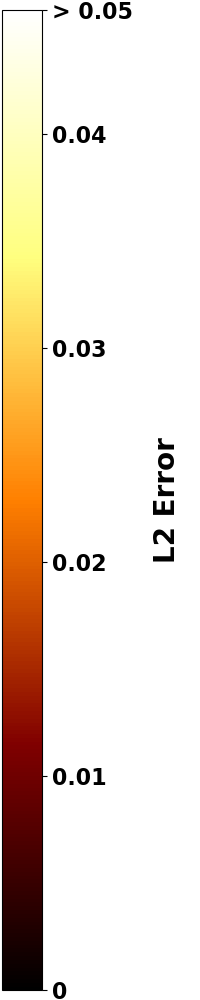}};
\end{tikzpicture}
\input{figures/validation/attacks_kappa-inf}
\vspace{-2mm}
\caption{$\kappa=\infty$, $d=0$}
\vspace{1mm}
\label{fig:robustness:attacks_kappa-inf}
\input{figures/validation/attacks_kappa-inf_d-66}
\vspace{-2mm}
\caption{$\kappa=\infty$, $d=66$}
\vspace{1mm}
\label{fig:robustness:attacks_kappa-inf_d-66}
\input{figures/validation/attacks_kappa-inf_d-1000}
\vspace{-2mm}
\caption{$\kappa=\infty$, $d=1000$}
\label{fig:robustness:attacks_kappa-inf_d-1000}
\end{figure*}

In our evaluation, we observe the GCs to be measurably more robust compared to the DC in terms of necessary adversarial perturbation in order to successfully fool the model.
For achieving a successful attack, the adversarial noise generally needs to be amplified for models with better generative modeling capabilities (smaller values for $\beta$), as is the case for higher values for $\kappa$ (forcing highly confident but wrong predictions).
We would expect the trend to continue for $\beta < 1$, for the adversarial perturbations to be even larger, 
but at that point the task performance may not be satisfactory anymore. We show this qualitatively in Fig.~\ref{fig:robustness:attack_kappa}. 
The gap to the ResNet (roughly factor 2) is consistent to what was observed for a simplified version of CIFAR10 in \cite{li2018generative}.
In terms of $\kappa$, the adversarial perturbations increase a lot for $\kappa = \infty$ (forcing confident fooled predictions), but the increase is homogeneous across models including the ResNet.
Furthermore, as can be seen in Figure \ref{fig:robustness:attacks_kappa-inf}, optimizing for highest possible confidence results in adversarial noise that is clearly visible to the human eye. 
For $\kappa=0.01$ and $\kappa=1$ on the other hand, the applied noise is a lot harder to perceive by humans (See Fig. \ref{fig:robustness:attacks_kappa-0.01} and Fig. \ref{fig:robustness:attacks_kappa-1}). 
We make a second important observation:
For most models, the predictive confidence is similar to the ResNet.
However, $\beta=1$ and $\beta=2$ are 100\% confident in their (wrong) prediction, even for low values of $\kappa$.
During the attack, this occurs while the fooling part of the loss is already satisfied and has no effect.
The phenomenon is purely due to the attack reducing the amplitude of the perturbation.
Evidently, by reducing the attack amplitude, the image moves into an even more confident region of latent space.
So in the sense of predictive uncertainty on adversarial examples, GCs actually seem to be more vulnerable to adversarial attacks.

Regarding the adversarial perturbation needed in order to successfully fool the model while simultaneously trying to fool the attack detection mechanism we make three main observations:
the adversarial noise is increased when putting a higher focus on fooling the attack detection mechanism. 
This can also be clearly seen in Fig.~\ref{fig:robustness:attack_d} and in the quantitative comparison in Fig. \ref{fig:app_adversarial_bar_graph}, first column when comparing the three bars per $\beta$.
Second, as shown in the second column of Fig.~\ref{fig:app_adversarial_bar_graph}, we observe the attack detection capabilities generally to decrease. 
For the good detection models such as $\beta =1$, the score stays reasonably high, while the weaker models have a detection score significantly worse than random.
Lastly, the predictive uncertainty is not affected by the detection attack at all (Fig.~\ref{fig:app_adversarial_bar_graph}, third column).

Inspecting the perturbed images also provides some clues as to how the attack fools the detection mechanism:
They show uniformly decreased contrast.
As shown in \cite{nalisnick2018deep}, such low-contrast images have unnaturally high estimated likelihoods.
In our case, this seems to compensate for the lower estimated likelihood caused by the noise-like adversarial perturbations, to make the image appear `typical' overall.

\begin{table*}
\resizebox{\textwidth}{!}{
\begin{tabular}{lccccccccccccccccccl}
\cline{1-19}
\multicolumn{1}{|c|}{\cellcolor[HTML]{8B8B8B}{\color[HTML]{FFFFFF} }}                                   & \multicolumn{3}{c|}{\cellcolor[HTML]{8B8B8B}{\color[HTML]{FFFFFF} }}                                                                                                                                                                                       & \multicolumn{3}{c|}{\cellcolor[HTML]{8B8B8B}{\color[HTML]{EFEFEF} }}                                                                                                                                                                                       & \multicolumn{3}{c|}{\cellcolor[HTML]{8B8B8B}{\color[HTML]{EFEFEF} }}                                                                                                                                                                                       & \multicolumn{4}{c|}{\cellcolor[HTML]{8B8B8B}{\color[HTML]{EFEFEF} \textbf{OoD}}}                                                                                                                                                                                                                                                                     & \multicolumn{4}{c|}{\cellcolor[HTML]{8B8B8B}{\color[HTML]{EFEFEF} \textbf{Entropy}}}                                                                                                                                                                                                                                                                            & \multicolumn{1}{c|}{\cellcolor[HTML]{8B8B8B}{\color[HTML]{EFEFEF} }}                                    &  \\
\end{tabular}}
\caption{$\kappa=0.01$}
\label{tbl:robustness:kappa-0.01}

\resizebox{\textwidth}{!}{\begin{tabular}{lccccccccccccccccccl}
\cline{1-19}
\multicolumn{1}{|c|}{\cellcolor[HTML]{8B8B8B}{\color[HTML]{FFFFFF} }}                                   & \multicolumn{3}{c|}{\cellcolor[HTML]{8B8B8B}{\color[HTML]{FFFFFF} }}                                                                                                                                                                                       & \multicolumn{3}{c|}{\cellcolor[HTML]{8B8B8B}{\color[HTML]{EFEFEF} }}                                                                                                                                                                                       & \multicolumn{3}{c|}{\cellcolor[HTML]{8B8B8B}{\color[HTML]{EFEFEF} }}                                                                                                                                                                                       & \multicolumn{4}{c|}{\cellcolor[HTML]{8B8B8B}{\color[HTML]{EFEFEF} \textbf{OoD}}}                                                                                                                                                                                                                                                                     & \multicolumn{4}{c|}{\cellcolor[HTML]{8B8B8B}{\color[HTML]{EFEFEF} \textbf{Entropy}}}                                                                                                                                                                                                                                                                            & \multicolumn{1}{c|}{\cellcolor[HTML]{8B8B8B}{\color[HTML]{EFEFEF} }}                                    &  \\
\end{tabular}}
\caption{$\kappa=1$}
\label{tbl:robustness:kappa-1}

\resizebox{\textwidth}{!}{\begin{tabular}{lccccccccccccccccccl}
\cline{1-19}
\multicolumn{1}{|c|}{\cellcolor[HTML]{8B8B8B}{\color[HTML]{FFFFFF} }}                                   & \multicolumn{3}{c|}{\cellcolor[HTML]{8B8B8B}{\color[HTML]{FFFFFF} }}                                                                                                                                                                                       & \multicolumn{3}{c|}{\cellcolor[HTML]{8B8B8B}{\color[HTML]{EFEFEF} }}                                                                                                                                                                                       & \multicolumn{3}{c|}{\cellcolor[HTML]{8B8B8B}{\color[HTML]{EFEFEF} }}                                                                                                                                                                                       & \multicolumn{4}{c|}{\cellcolor[HTML]{8B8B8B}{\color[HTML]{EFEFEF} \textbf{OoD}}}                                                                                                                                                                                                                                                                     & \multicolumn{4}{c|}{\cellcolor[HTML]{8B8B8B}{\color[HTML]{EFEFEF} \textbf{Entropy}}}                                                                                                                                                                                                                                                                            & \multicolumn{1}{c|}{\cellcolor[HTML]{8B8B8B}{\color[HTML]{EFEFEF} }}                                    &  \\
\end{tabular}}
\caption{$\kappa=\infty$}
\label{tbl:robustness:kappa-inf}

\caption{Table \ref{tbl:robustness:kappa-0.01},\ref{tbl:robustness:kappa-1} and \ref{tbl:robustness:kappa-inf} show the quantitative results of our adversarial attack experiments. Each table was obtained by performing the attack with a different value for $\kappa \in {0.01,1,\inf}$. A high value for $\kappa$ aims a more certain posterior for targeted classes. The cell background colors green, orange and red stand for different values for $d$ to ease the comparison across models and tables for a human reader. The variable $d$ quantifies the strength on fooling the intrinsic attack detection mechanism of our learned classifiers. Note, that the ResNet does not model the data likelihood and therefore has not this capability. We report the maximum class probability (Confidence), the pixel-wise $l^2$-distance between the original and the adversarially perturbed image averaged over all pixels (Corruption), the success rate of the attack (Success), the one (1t-tt) and two-tailed ($p(x)$) typicality test OoD detection scores, as well as the posterior predictive uncertainty for the original ($X$) and the corrupted validation data $x_{corr.}$. Furthermore, we report the likelihood of the original validation data ($p(X)$ Val). }
\end{table*}

\clearpage

\clearpage

\clearpage

\end{document}